%% file: main.tex
\documentclass{article}

\usepackage{iclr2027_conference,times}
\input{math_commands.tex}

\usepackage{amsmath,amssymb}
\usepackage{booktabs} 
\usepackage{graphicx}
\usepackage{float}
\usepackage{capt-of}
\usepackage{placeins}
\usepackage{microtype}
\usepackage{multirow}
\usepackage{tabularx}
\usepackage{xcolor}
\usepackage{algorithm}
\usepackage{listings}
\usepackage[pagebackref=true,breaklinks=true,colorlinks=true,urlcolor=magenta,bookmarks=false,citecolor=blue]{hyperref}
\usepackage{url}

\definecolor{draftred}{RGB}{170,35,35}
\definecolor{placeholdergray}{RGB}{245,245,245}

\newcommand{\crm}{\textsc{CRM}}
\newcommand{\scopename}{SCOPE-4D}
\newcommand{\ourmodel}{\scopename{} (Ours)}
\newcommand{\scopesftname}{\scopename\ensuremath{_{\mathrm{SFT}}}}
\newcommand{\scopesft}{\scopesftname{} (Ours)}
\newcommand{\scopedata}{SCOPE-5K}
\newcommand{\scopephantom}{\textit{SCOPE-colon-phantom}}
\newcommand{\scopereal}{\textit{SCOPE-colon-real}}

\title{SCOPE-4D: Endoscopic 4D Geometry Foundation Models}

\input{authors}
\iclrfinalcopy
\begin{document}

\maketitle
\lhead{}

\input{sections/00_teaser}
\input{sections/00_abstract}
\input{sections/01_introduction}
\input{sections/02_related_work}
\input{sections/03_data}
\input{sections/04_method}
\input{sections/05_experiments}
\input{sections/06_conclusion}

\clearpage
\subsection*{AI use statement}
Generative AI assisted manuscript drafting, editing, and literature and code inspection.
A vision-language model also supports video screening in the data pipeline
(Section~\ref{sec:data}).

\subsection*{Reproducibility statement}
We document data splits and sampling, annotation checks, motion-target
construction, training settings, and dataset-specific evaluation protocols. Complete
per-domain geometry and tracking-threshold results accompany the main comparisons.

\subsection*{Ethics statement}
This work combines existing endoscopic datasets with in-house phantom recordings
and clinical video. The clinical clips in \scopereal{} are used for retrospective
reconstruction assessment without altering clinical procedures or patient care.
Six clinical and biomedical experts participated in a blinded reconstruction-ranking study (Appendix~\ref{app:user-study}). The reported experiments
evaluate reconstruction quality and do not establish safety for clinical deployment.

\bibliography{references}
\bibliographystyle{iclr2027_conference}

\clearpage
\appendix
\input{sections/appendix}

\end{document}

%% file: math_commands.tex
\usepackage{amsmath,amsfonts,bm}

\def\eqref#1{equation~\ref{#1}}

\def\1{\bm{1}}

\DeclareMathAlphabet{\mathsfit}{\encodingdefault}{\sfdefault}{m}{sl}
\SetMathAlphabet{\mathsfit}{bold}{\encodingdefault}{\sfdefault}{bx}{n}



%% file: authors.tex
\author{\parbox[t]{\dimexpr\textwidth-2\tabcolsep\relax}{\normalfont\bfseries
  Chaoyi Zhou\textsuperscript{1,2}\thanks{This work was done during Chaoyi Zhou's internship at United Imaging Intelligence, Boston, MA, USA.},
  Zhongpai Gao\textsuperscript{1}\thanks{Corresponding authors.},
  Anwesa Choudhuri\textsuperscript{1},
  Meng Zheng\textsuperscript{1},\\
  Benjamin Planche\textsuperscript{1},
  Run Wang\textsuperscript{2},
  Terrence Chen\textsuperscript{1},
  Siyu Huang\textsuperscript{2}\footnotemark[2],
  Ziyan Wu\textsuperscript{1}\\[4pt]
  {\normalfont
  \textsuperscript{1}United Imaging Intelligence, Boston, MA, USA\\
  \textsuperscript{2}Clemson University, Clemson, SC, USA}
}}

%% file: sections/00_teaser.tex
\begin{figure*}[h]
  \centering
  \includegraphics[width=\linewidth]{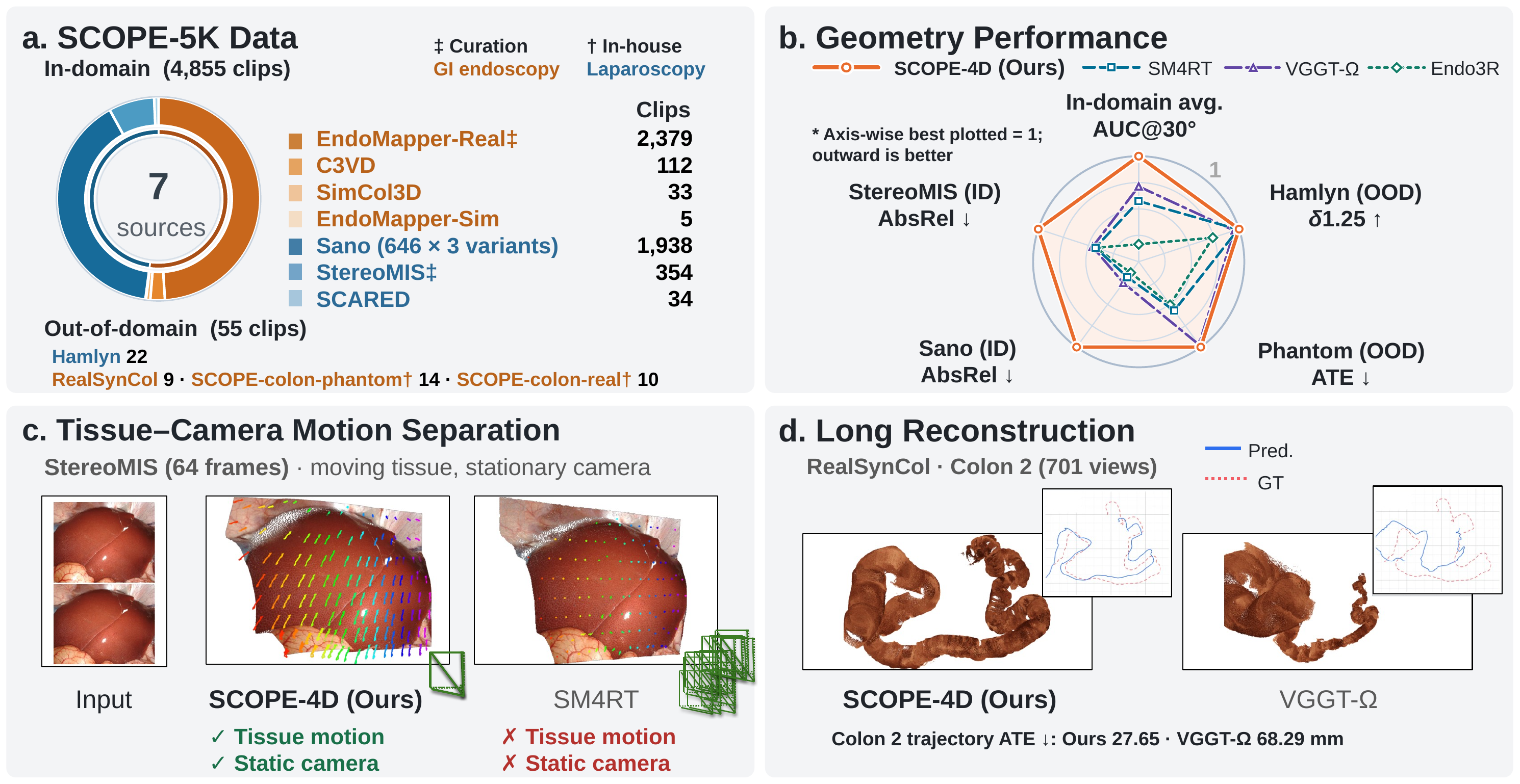}
  \caption{\textbf{SCOPE-4D overview.} (a) Our curation and annotation pipeline
  builds \scopedata{} for geometric supervision and cross-domain evaluation.
  \scopename{} (b) achieves strong cross-domain geometry,
  (c) recovers tissue motion while preserving camera stability in a
  stationary-camera sequence, unlike SM4RT,
  and (d) enables stable long-sequence reconstruction.}
  \label{fig:teaser}
\end{figure*}

%% file: sections/00_abstract.tex
\begin{abstract}
Geometric understanding supports endoscopic navigation and robotic assistance,
but learning reliable endoscopic geometry faces two challenges: scarce geometric
annotations and ambiguity between camera motion and tissue deformation.
We present \scopename{}, an endoscopic 4D geometry foundation model that jointly
predicts camera parameters, dense geometry, and 3D tissue trajectories from
monocular RGB video in a single forward pass. Our curation and annotation
pipeline constructs \scopedata{}, a collection of approximately 5,000 clips
spanning real and synthetic gastrointestinal endoscopy and laparoscopy.
The collection provides rich geometric supervision and includes newly collected
phantom and real-colonoscopy evaluation sets. Geometric supervised fine-tuning
on \scopedata{} learns endoscopic priors that improve camera and depth estimation.
Common--Residual Motion (\crm) further constrains local deformation relative to
common tissue movement. Together with geometric supervision, CRM and trajectory
supervision further improve camera and depth estimation over geometric fine-tuning
alone while enabling dense 3D tissue tracking.
Evaluations on public and newly collected benchmarks demonstrate strong
in-domain and out-of-domain geometry, superior 3D tracking, and more stable
long-sequence colon reconstruction. A blinded user study further
supports the perceived reconstruction quality on real clinical video. Together, these results demonstrate
the value of large-scale endoscopic supervision and motion constraints for
joint geometry estimation and tissue tracking.
\textbf{Project page:} \url{https://chaoyizh.github.io/SCOPE-4D-page/}.
\end{abstract}

%% file: sections/01_introduction.tex
\section{Introduction}
\label{sec:intro}

Endoscopic navigation, surgical planning, and robotic assistance require spatial
understanding within a confined workspace and limited field of view. Recovering
camera poses, depth, and dense point clouds from video can help clinicians assess
tissue shape and position instruments.

Traditionally, Structure-from-Motion (SfM) and Multi-View Stereo (MVS) recover scene
geometry from multiple images~\citep{schonberger2016sfm,schonberger2016mvs}.
Geometry foundation models instead learn transferable priors through large-scale
pretraining on natural scenes. DUSt3R predicts pointmaps from image pairs~\citep{wang2024dust3r};
VGGT and VGGT-$\Omega$ jointly predict cameras and dense geometry in a single
forward pass~\citep{wang2025vggt,wang2026vggtomega}, and D4RT extends feed-forward
prediction to dynamic 4D reconstruction~\citep{zhang2026d4rt}.

Endoscopy adds tissue deformation to this reconstruction problem. EndoNeRF and
EndoGaussian model deforming tissue with camera and depth guidance, but require
per-sequence optimization~\citep{wang2022endonerf,liu2024endogaussian}.
Feed-forward methods avoid this fitting: AF-SfMLearner uses self-supervised image
reconstruction~\citep{shao2022afsfm}, while Endo3R and Endo-FASt3R adapt pretrained
general-domain models~\citep{guo2025endo3r,zeinoddin2025endofast3r}.
However, photometric supervision remains ambiguous under weak texture and
deformation, and pretrained priors have limited endoscopic coverage.

Two challenges remain in building endoscopic geometry foundation models.
\textbf{First, endoscopic geometric annotations are scarce and limited in coverage.}
Datasets such as SCARED and C3VD provide valuable reference geometry, yet cover a
limited range of subjects and acquisition conditions
\citep{allan2021scared,bobrow2023c3vd}. Obtaining reliable geometric supervision
from raw clinical video at scale is difficult: weak texture, specular reflections,
liquids, bleeding, and illumination changes
can disrupt the correspondences on which SfM relies.
Limited annotations restrict the geometric priors learned across endoscopic domains.
\textbf{Second, geometry supervision does not explicitly constrain how tissue
points move over time.} Endoscopic scenes combine camera motion, coherent tissue motion
such as breathing, and local deformation from tool interaction. These processes can
produce similar image changes, creating ambiguity between camera movement and scene
deformation. A model can therefore predict camera motion from deforming tissue
even with a stationary camera, misaligning observations and distorting reconstructed geometry.

To expand reliable geometric supervision, we develop a data curation and annotation
pipeline that combines vision-language screening, tracking-assisted camera
reconstruction, and camera-conditioned dense annotation. We assess annotation
reliability through geometric quality checks and comparisons between
independent correspondence sources.
With this pipeline, we construct \scopedata{}, a collection of approximately 5,000
endoscopic clips covering real and synthetic gastrointestinal (GI) endoscopy and laparoscopy.
The collection also includes two newly collected evaluation sets:
\scopephantom{} provides externally tracked camera references in a physical colon
phantom, while \scopereal{} supports a blinded user study of reconstruction
quality in clinical videos.

We present \scopename{}: Structured Camera and Organ Motion Prediction for Endoscopy,
an endoscopic 4D geometry foundation model.
To our knowledge, it is the first such model to jointly predict camera parameters,
dense geometry, and 3D tissue trajectories from monocular RGB video in a single
forward pass. The model combines supervised fine-tuning (SFT)
for geometry with Common--Residual Motion (\crm).
SFT on \scopedata{} learns endoscopic geometric priors, improving depth and
camera-pose estimation. To address the second challenge, CRM explicitly constrains
tissue motion by composing a common SE(3)
transform with point-dependent residual motion. With trajectory supervision,
CRM further improves camera and depth estimation over geometric fine-tuning alone
while enabling dense 3D tissue tracking.

Our main contributions are:
\begingroup
\setlength{\topsep}{3pt}
\setlength{\partopsep}{0pt}
\setlength{\itemsep}{2pt}
\setlength{\parsep}{0pt}
\setlength{\parskip}{0pt}
\setlength{\leftmargini}{1.3em}
\begin{itemize}
  \item \textbf{Data:} A curation and annotation pipeline constructs \scopedata{},
  approximately 5,000 endoscopic clips for learning and evaluation,
  including phantom and real-colonoscopy benchmarks.
  \item \textbf{Model:} \scopename{} combines geometric SFT with \crm{} and
  trajectory supervision to improve camera and depth estimation while enabling
  dense 3D tissue tracking.
  \item \textbf{Evaluation:} We demonstrate leading geometry performance across
  in-domain and held-out endoscopic benchmarks, improved 3D tissue tracking,
  and more accurate long-sequence camera reconstruction.
\end{itemize}
\endgroup

%% file: sections/02_related_work.tex
\section{Related Work}
\label{sec:related}

\textbf{General geometry and 4D models.}
Structure-from-Motion (SfM) and Multi-View Stereo (MVS) recover geometry
through image correspondence and costly per-scene optimization
\citep{schonberger2016sfm,schonberger2016mvs}.
Geometry foundation models learn feed-forward reconstruction through large-scale pretraining
\citep{wang2024dust3r,tang2025mvdust3r,yang2025fast3r,wang2025cut3r,
wang2025vggt,wang2026vggtomega,lin2025da3,keetha2026mapanything}.
DUSt3R predicts paired pointmaps without calibration; VGGT jointly predicts
cameras and geometry through within-image and cross-image attention;
VGGT-$\Omega$ improves attention efficiency to scale training.
Depth Anything 3 represents single-image and multi-view geometry through
depth and ray maps, while MapAnything incorporates available camera and
depth information. Applying these advances to endoscopy requires broader
data coverage and reliable annotations.

General 4D models extend reconstruction to dynamic scenes through joint geometry
and motion prediction~\citep{zhang2026d4rt,karhade2025any4d,xiao2025spatialtrackerv2}.
4RC~\citep{luo2026fourrc} separates base geometry from relative motion.
SM4RT~\citep{lin2026sm4rt} introduces Structure-of-Motion (SoM) to represent
point motion through shared motion bases with point-dependent assignments.
In endoscopy, a tissue surface can move coherently with breathing while deforming
locally under tool interaction. \scopename{} separates these motions through
common--residual composition, constraining local deformation while preserving
shared movement.

\textbf{Endoscopic geometry models.}
Neural fields and Gaussian representations model deforming tissue,
but require optimizing a new scene for each
sequence~\citep{wang2022endonerf,zha2023endosurf,liu2024endogaussian,hayoz2024online}.
Methods that also estimate camera motion must distinguish it from tissue deformation;
ambiguity between the two can lead to tracking drift~\citep{shan2026nrgsslam}.
Feed-forward models instead learn across sequences and avoid per-scene fitting.
Early approaches estimate depth and camera motion through self-supervised image reconstruction
\citep{ozyoruk2021endoslam,shao2022afsfm}.
AF-SfMLearner~\citep{shao2022afsfm} compensates for appearance changes, but
photometric supervision remains ambiguous in weakly textured or deforming tissue.
More recent methods adapt pretrained general-domain models to endoscopic reconstruction.
EndoDAC~\citep{cui2024endodac} uses parameter-efficient depth adaptation, while
Endo3DAC~\citep{cui2025endo3dac} jointly estimates depth, pose, and intrinsics.
Endo-FASt3R~\citep{zeinoddin2025endofast3r} combines pretrained depth and pose models.
Endo3R~\citep{guo2025endo3r} supports online pointmap reconstruction through
short-term and long-term memories.
These models benefit from general-domain pretraining, but limited endoscopic
supervision restricts their ability to handle anatomical and visual variation.
Geometry supervision alone also leaves tissue motion insufficiently constrained.
\scopename{} combines geometric SFT on \scopedata{} with CRM to improve
geometry estimation and constrain tissue deformation.

\textbf{Endoscopic datasets.}
Endoscopic geometry learning is constrained by limited data scale and a shortage
of dense, accurate geometric annotations
\citep{allan2021scared,bobrow2023c3vd,azagra2023endomapper,rau2024simcol3d,hayoz2023pose}.
SCARED~\citep{allan2021scared} obtains depth through structured light in porcine
cadavers and propagates it to intermediate frames under a static-scene
assumption. C3VD~\citep{bobrow2023c3vd} obtains reference geometry by registering
colonoscopy images of physical phantoms to their 3D models. Both provide valuable
supervision under controlled conditions, with limited coverage of clinical
appearance and deformation. EndoMapper~\citep{azagra2023endomapper} broadens
clinical coverage through complete calibrated colonoscopy and gastroscopy procedures,
but lacks dense geometric
ground truth throughout its real videos. SimCol3D~\citep{rau2024simcol3d} provides
synthetic depth and camera poses, while StereoMIS~\citep{hayoz2023pose}
provides in-vivo stereo video and camera poses from robot kinematics.
Synthetic appearance and specialized acquisition setups limit the coverage of
routine clinical conditions. To broaden coverage, \scopedata{} combines existing
datasets with curated and annotated clinical video. Phantom and real-colonoscopy
benchmarks assess camera accuracy and reconstruction quality, respectively.

%% file: sections/03_data.tex
\section{SCOPE-5K: Endoscopic Geometry Data}
\label{sec:data}
\label{sec:data-overview}

\scopedata{} provides depth and camera supervision for learning endoscopic
geometric priors, together with benchmarks for evaluating reconstruction quality.
\scopedata{} contains 4,855 in-domain clips across seven sources and 55
out-of-domain clips, spanning real and synthetic GI endoscopy and laparoscopy
(Figure~\ref{fig:teaser}a). Our pipeline adds camera and dense geometry annotations
to EndoMapper-Real and dense depth to StereoMIS. We also introduce
\scopephantom{} and \scopereal{} for controlled camera evaluation and clinical
reconstruction assessment. Appendix~\ref{app:splits} details source composition,
annotation coverage, and splits.

\begin{figure*}[t]
  \centering
  \includegraphics[width=\linewidth]{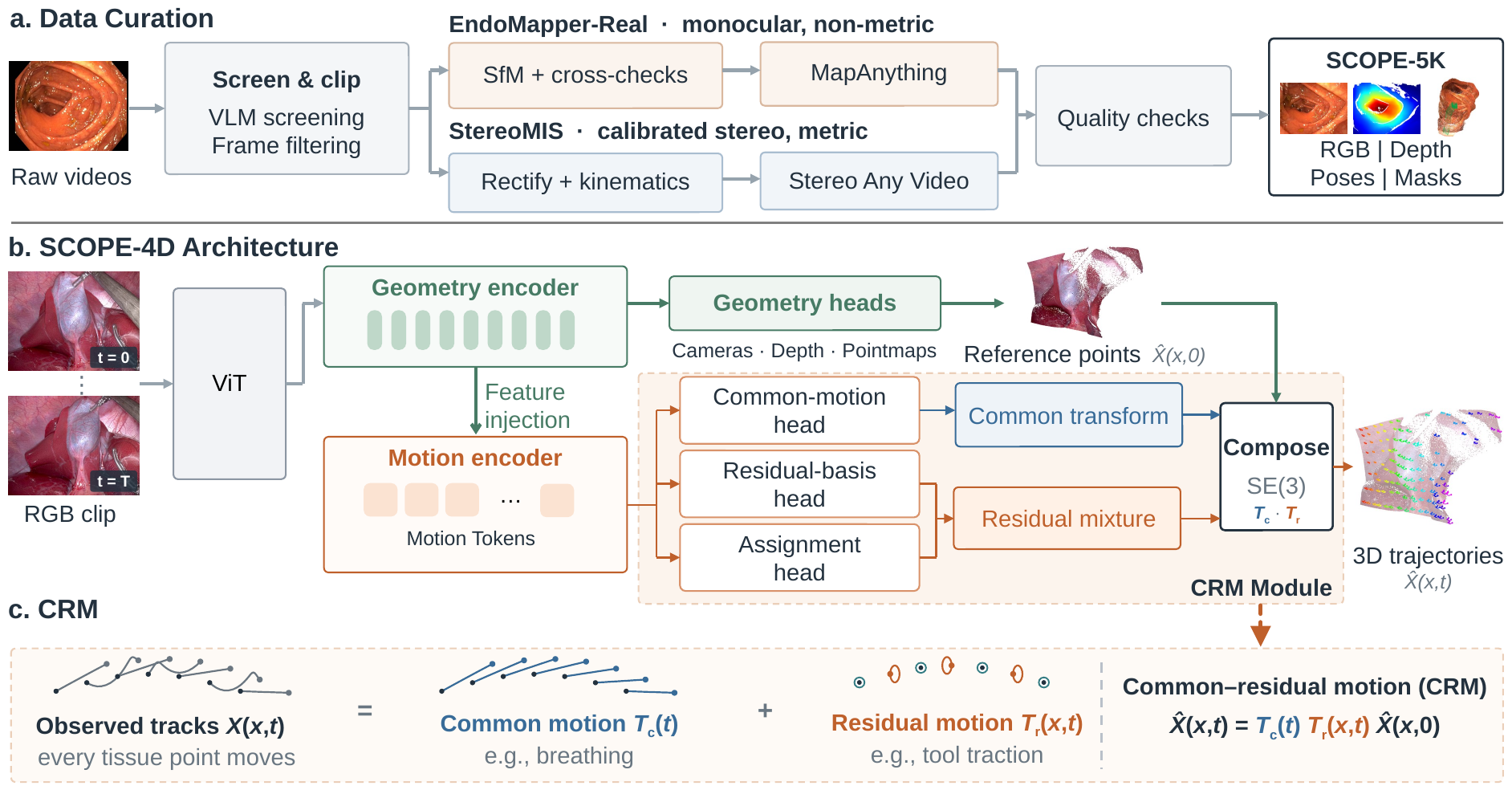}
  \caption{\textbf{Data annotation and \scopename{}.}
  (a) \scopedata{} curation and annotation.
  (b) Joint camera, geometry, and 3D tissue-track prediction.
  (c) Common--Residual Motion composes common tissue movement with local deformation.}
  \label{fig:method}
\vspace{-1em}
\end{figure*}

\subsection{Geometry Annotation}
\label{sec:geometry-annotation}

Our annotation pipeline combines clip selection, geometry reconstruction, and
quality control, with reconstruction tailored to the available camera information
(Figure~\ref{fig:method}a).

\textbf{Monocular geometry annotation.}
Unlike datasets with dedicated capture guidelines such as DL3DV-10K~\citep{ling2024dl3dv},
routine clinical videos contain obstructions, poor visibility, and abrupt camera motion.
For EndoMapper-Real, a VLM selects segments with visible anatomy and sufficient camera
movement, excluding obstructed or irrelevant content. Image-based filters reject frames
with blur, exposure failures, specular reflections, weak texture, or abrupt illumination
changes. We split the filtered videos at temporal discontinuities into reconstructable clips.

Weak texture, deformation, and specular reflections make local-feature matches
unreliable for endoscopic SfM~\citep{barbed2022superpointendoscopy}.
We instead convert MFT~\citep{neoral2024mft} temporal tracks into pairwise matches
for incremental SfM.
LoMa~\citep{nordstrom2026loma} and SIFT with LightGlue~\citep{lowe2004sift,lindenberger2023lightglue}
provide independent reconstructions to cross-check the recovered camera trajectory
after similarity alignment.

Conventional MVS estimates depth by photometric matching~\citep{schonberger2016mvs},
which is unreliable under endoscopic illumination changes, reflections, and weak
texture~\citep{liu2020endoscopydepth}. We instead condition
MapAnything~\citep{keetha2026mapanything} on the recovered poses and intrinsics
to generate dense depth and point clouds using its learned geometric priors.

Quality control checks multi-view consistency of dense predictions and agreement
between independent camera reconstructions. Pixel masks exclude
depth-edge artifacts and regions without multi-view support.
Across 93 recordings, screening produces 3,034 clean clips; geometric QC and
held-out filtering retain 2,379.
Appendix~\ref{app:pipeline} details the checks, thresholds, and verification results.

\textbf{Stereo geometry annotation.}
StereoMIS provides calibrated stereo video and camera poses from robot kinematics,
allowing us to annotate depth without reconstructing cameras. We use Stereo Any Video
(SAV)~\citep{jing2025stereoanyvideo} to estimate disparity from rectified stereo pairs
and convert it to depth using the known focal length and baseline.
Clip selection and quality control follow the same principles as the monocular
pipeline, with checks tailored to stereo geometry.
Appendices~\ref{app:curation}--\ref{app:pipeline} detail the VLM prompt,
annotation procedures, and QC criteria.

\subsection{SCOPE-colon Benchmark}
\label{sec:data-evaluation}

We collect \scopephantom{} and \scopereal{} to evaluate colonoscopy reconstruction.
For \scopephantom{}, we synchronize video captured in a physical colon phantom
with a five-degree-of-freedom electromagnetic tracker. The tracker provides
camera position and viewing-direction references for quantitative evaluation.
\scopereal{} uses clinical colonoscopy videos for a blinded assessment of
reconstruction quality.
Together, they support controlled camera evaluation and clinical validation
(Section~\ref{sec:experiments}); Appendix~\ref{app:inhouse} details their acquisition
and annotations.

\FloatBarrier

%% file: sections/04_method.tex
\section{SCOPE-4D: Endoscopic 4D Geometry Foundation Model}
\label{sec:method}

\scopename{} combines geometric supervised fine-tuning on \scopedata{} with
Common--Residual Motion (\crm) to jointly predict cameras, dense geometry, and
3D tissue tracks from monocular RGB video (Figure~\ref{fig:method}b--c).

\subsection{Geometry Supervised Fine-Tuning}
\label{sec:sft}

To learn geometric priors suited to endoscopy, we apply supervised fine-tuning
(SFT) to a pretrained geometry model using camera and depth annotations from
\scopedata{}.
Given an uncalibrated monocular RGB video $\mathcal I=(I_0,\ldots,I_{T-1})$,
where $I_t\in\mathbb R^{H\times W\times3}$, we use the first frame, $t=0$,
as the reference frame. An image encoder extracts frame
features, which the backbone aggregates across views into geometry features
$\{F_t\}_{t=0}^{T-1}$. Camera and depth heads decode these features into
intrinsics $\mathbf K_t\in\mathbb R^{3\times3}$, camera poses
$C_t\in\mathrm{SE}(3)$, and depth maps $D_t\in\mathbb R^{H\times W}$.
Here $C_t$ maps camera coordinates into a shared reference frame.
Back-projecting each depth map with $\mathbf K_t$ and applying $C_t$ produces
a dense pointmap $P_t\in\mathbb R^{H\times W\times3}$ in this frame.
The resulting mapping is
\begin{equation}
  f:\mathcal I\mapsto
  \{\mathbf K_t,C_t,D_t,P_t\}_{t=0}^{T-1}.
  \label{eq:geometry-mapping}
\end{equation}
We denote this geometry-only model as \scopesftname{} and the full geometry--motion
model as \scopename{}. Tracking tissue points requires explicit motion modeling.

\subsection{Common--Residual Motion}
\label{sec:crm}

Common--Residual Motion (\crm) constrains tissue movement beyond camera and depth
supervision, reducing ambiguity between camera motion and tissue deformation.

\textbf{CRM representation.}
Structure-of-Motion (SoM) in SM4RT~\citep{lin2026sm4rt} mixes shared motion bases
using point-dependent assignments fixed over time. In endoscopy, breathing induces
coherent tissue movement while tool interaction causes local deformation.
A single mixture can spread local deformation into surrounding tissue, yet
suppressing it can weaken shared movement. CRM instead composes a shared rigid
transform with a point-dependent residual transform (Figure~\ref{fig:method}c),
constraining local deformation relative to common motion.
Each transform is parameterized by a six-dimensional
twist $\xi=(v,\omega)$, where $v,\omega\in\mathbb R^3$ are its translational and
angular components. CRM uses a common twist $\xi_c(t)$ and $K$ residual
bases $\{\Delta\xi_k(t)\}_{k=0}^{K-1}$. The first basis is a fixed zero-residual anchor,
$\Delta\xi_0(t)\equiv0$, which introduces no local deformation.
The model predicts the common twist and the remaining $K-1$ residual twists
per frame. All twists are zero at $t=0$, so each track begins at its reference
point. For a reference-frame pixel $x$,
logits $\ell_k(x)$ define time-invariant weights $q_k(x)$ that mix the residual bases:
\begin{equation}
  q_k(x)=\frac{\exp\ell_k(x)}{\sum_{j=0}^{K-1}\exp\ell_j(x)},\qquad
  \Delta\xi(x,t)=\sum_{k=1}^{K-1}q_k(x)\Delta\xi_k(t).
  \label{eq:assignment}
\end{equation}
We convert twist coordinates into SE(3) transforms using the exponential
map~\citep{sola2018micro}, defined in Appendix~\ref{app:crm-proof}.
Composing the common and residual transforms gives
\begin{equation}
  \widehat X(x,t)=T_c(t)T_r(x,t)X(x,0),\qquad
  T_c(t)=\operatorname{Exp}(\xi_c(t)),\quad
  T_r(x,t)=\operatorname{Exp}(\Delta\xi(x,t)).
  \label{eq:crm}
\end{equation}
Here $X(x,0)$ is the reference 3D point, and its predicted positions
$\widehat X(x,t)$ form the 3D tissue track.

For a reference point $p=X(x,0)$, CRM bounds displacement relative to common motion:
\begin{equation}
 \|\widehat X(x,t)-T_c(t)p\|_2\leq(1-q_0(x))B(p,t),
 \label{eq:crm-bound-main}
\end{equation}
where $B(p,t)$ depends only on the residual bases and the reference point
(Appendix~\ref{app:crm-proof}). For fixed bases, increasing $q_0(x)$ tightens
this bound. As $q_0(x)\to1$, the point follows $T_c(t)p$. Thus, the
anchor suppresses local deformation without freezing the tissue, whereas a zero
mixed twist in SoM suppresses all point motion.
Appendix~\ref{app:crm-proof} provides the proof.

\subsection{Joint Geometry and Motion Prediction}
\label{sec:motion-learning}

\scopename{} decodes geometry and motion from shared features (Figure~\ref{fig:method}b).
Tissue tracks are represented in the reference camera frame, separately from
camera poses $C_t$. Joint trajectory supervision gives the shared features an
explicit representation of tissue movement, helping camera and depth prediction
account for deformation.
Linear projections map four intermediate geometry-backbone layers into a motion
transformer that combines spatial and temporal information. Common-motion,
residual-motion, and dense assignment heads predict $\xi_c(t)$,
$\Delta\xi_k(t)$, and reference-frame logits $\ell_k(x)$, respectively.
Applying CRM to the reference pointmap $P_0$ produces dense tracks
$\widehat X_t\in\mathbb R^{H\times W\times3}$, where
$\widehat X_t(x)=\widehat X(x,t)$ is the position at time $t$ of reference point $x$.
The complete mapping is
\begin{equation}
 f_{\mathrm{4D}}:\mathcal I\mapsto
 \{\mathbf K_t,C_t,D_t,P_t,\widehat X_t\}_{t=0}^{T-1}.
 \label{eq:scope4d-mapping}
\end{equation}
\subsection{Training}
\label{sec:training-supervision}

We jointly supervise scene geometry and tissue motion:
\begin{equation}
 \mathcal L = \mathcal L_{\mathrm{geo}}+\mathcal L_{\mathrm{tracking}},
 \qquad \mathcal L_{\mathrm{tracking}} =\lambda_{\mathrm{tr}}\mathcal L_{\mathrm{tr}}+\lambda_c\mathcal L_c+\lambda_q\mathcal L_q,
\label{eq:tracking-loss}
\end{equation}
where the $\lambda$ terms balance the motion objectives. Following the
MapAnything training framework~\citep{keetha2026mapanything}, geometry supervision
uses $\mathcal L_{\mathrm{geo}}=\mathcal L_{\mathrm{pts}}+
\mathcal L_{\mathrm{cam}}+\mathcal L_{\mathrm{depth}}$ to constrain predicted
pointmaps, camera parameters, and depth maps against their annotations.
Motion supervision also updates the shared backbone, shaping the features used
for camera and depth prediction.

Tracking labels combine trajectories from the synthetic Sano
dataset~\citep{openh2026}, derived from rendered flow and depth,
with MFT correspondences lifted into 3D using StereoMIS depth and camera poses.
StereoMIS tracks exclude occlusions and instruments.
Trajectory supervision $\mathcal L_{\mathrm{tr}}$ compares predicted and annotated
3D displacements with a weighted squared loss, balancing points with larger and
smaller residual motion. For the common-motion target, we robustly select
inlier correspondences $\mathcal J_t$ from annotated tracks $X_i(t)$ and fit a
rigid transform:
\begin{equation}
 T_c^\star(t)=\underset{G\in\mathrm{SE}(3)}{\arg\min}
 \sum_{i\in\mathcal J_t}\|GX_i(0)-X_i(t)\|_2^2.
 \label{eq:common-target}
\end{equation}
The common-motion loss $\mathcal L_c$ compares the point positions produced by
predicted and fitted transforms. Assignment supervision $\mathcal L_q$ uses binary
KL divergence between predicted anchor probabilities and targets derived from
track residuals relative to this transform. Small residuals receive high anchor
targets, encouraging points to follow common motion without unnecessary local
deformation. Appendix~\ref{app:motion-targets} gives the target construction and
loss definitions.

%% file: sections/05_experiments.tex
\section{Experiments}
\label{sec:experiments}

\textbf{Implementation details.}
Geometric SFT and the full model are independently initialized from the same
pretrained VGGT-$\Omega$ geometry backbone~\citep{wang2026vggtomega} and trained on
\scopedata{} for 29,000 steps. The full model initializes its motion modules from
SM4RT~\citep{lin2026sm4rt}.
Appendices~\ref{app:optimization} and~\ref{app:evaluation} detail training and evaluation.

\subsection{3D Geometry Reconstruction and Tracking}

\input{tables/main_geometry}

\begin{figure*}[t]
  \centering
  \includegraphics[width=\linewidth]{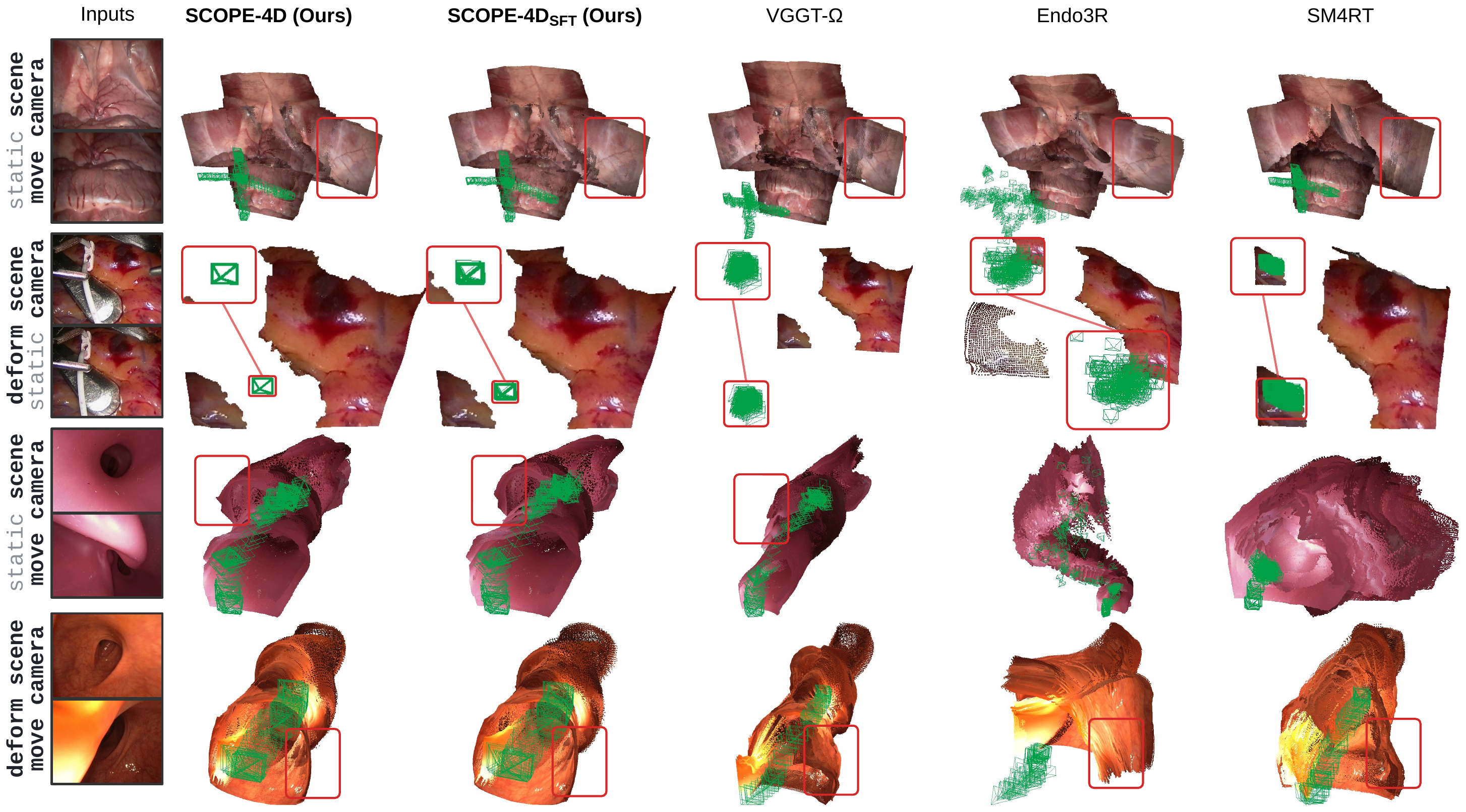}
  \vspace{-2em}
  \caption{\textbf{Qualitative comparison of 3D geometry reconstruction.}
  Instrument regions are masked out.
  Highlighted regions show surface consistency
  and suppression of spurious camera motion.}
  \label{fig:geometry-qualitative}
\vspace{-1em}
\end{figure*}

\textbf{3D geometry reconstruction.}
Geometric SFT on \scopedata{} substantially improves camera and depth estimation
(Table~\ref{tab:main-geometry}). Relative to pretrained VGGT-$\Omega$, in-domain
(ID) pose AUC@$5^\circ$ rises from 7.81 to 29.44, while depth AbsRel falls from
0.093 to 0.042. Geometric SFT outperforms all compared endoscopy-specific methods
on every metric reported in Table~\ref{tab:main-geometry}. Full \scopename{}
further improves ID pose AUC and depth accuracy.
On out-of-domain (OOD) benchmarks, without dataset-specific fine-tuning,
the full model outperforms competing methods
in camera-stability ATE, ARE, and $\delta_{1.25}$ on the held-out laparoscopic
Hamlyn benchmark~\citep{recasens2021endodm} and achieves the lowest
ATE and viewing-direction error on \scopephantom{}.
Figure~\ref{fig:geometry-qualitative} shows more coherent surfaces, fewer duplicated
layers, and better-preserved colon structure. In its stationary-camera example,
the full model yields more tightly clustered camera predictions than geometric
SFT despite tissue deformation, while VGGT-$\Omega$, Endo3R, and SM4RT exhibit
more pronounced spurious camera motion.
Appendix~\ref{app:results} provides per-dataset results and further comparisons.

\input{tables/tracking}

\begin{figure*}[t]
  \centering
  \includegraphics[width=\linewidth]{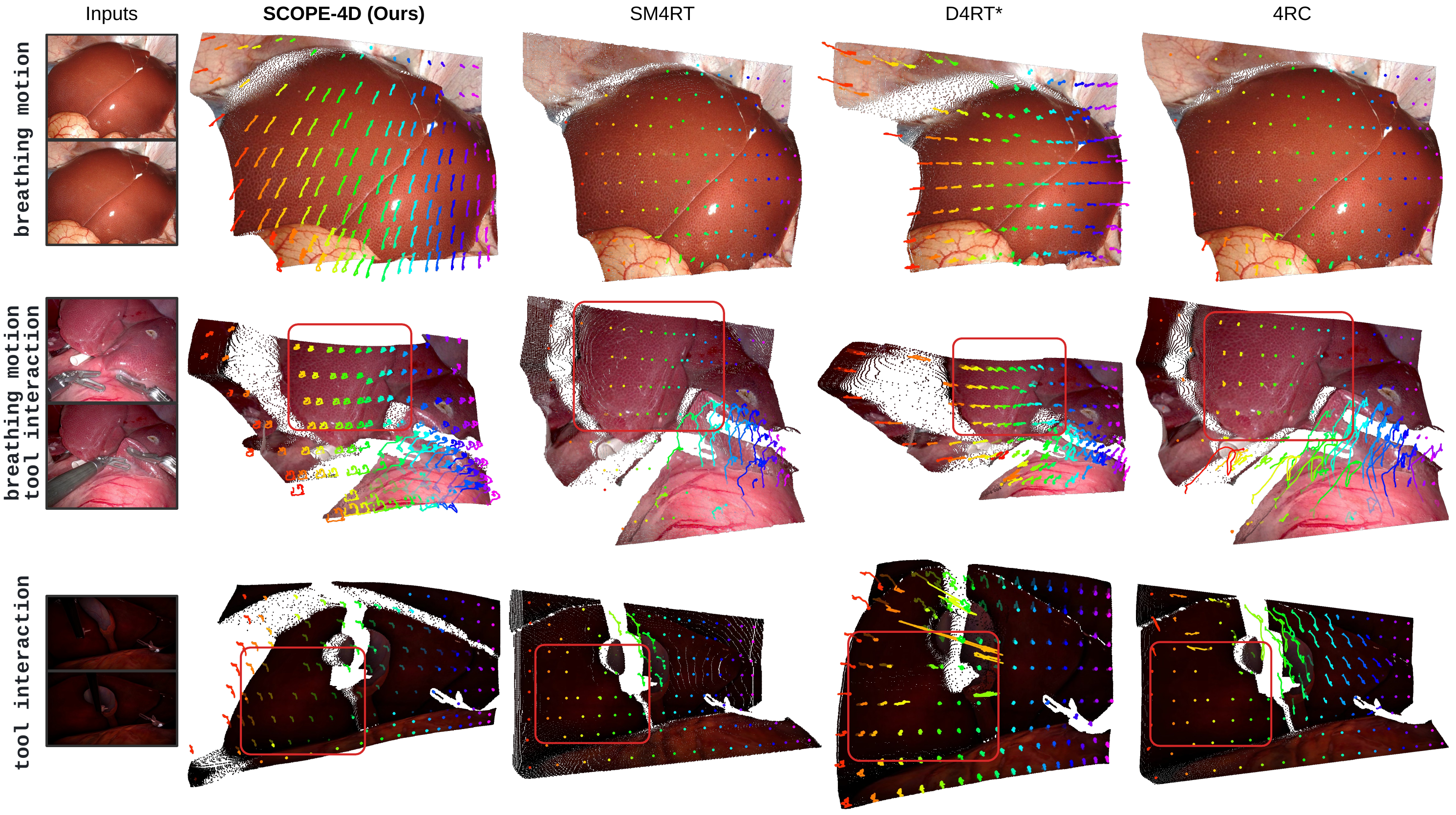}
  \caption{\textbf{3D tissue tracking under different motion conditions.}
  Instrument regions are masked out. Red boxes highlight the suppression of spurious motion.}
  \label{fig:tracking-qualitative}
\vspace{-1em}
\end{figure*}

\textbf{3D tracking.}
\scopename{} achieves the highest APD on both Sano and StereoMIS and the lowest
EPE on StereoMIS (Table~\ref{tab:tracking}).
On StereoMIS, it improves EPE by 19.8\% over the best competing method.
It also outperforms the competing tracking methods in camera and depth metrics
on both datasets.
Figure~\ref{fig:tracking-qualitative} shows coherent tissue trajectories with
larger local displacements concentrated near instrument interactions.
SM4RT and 4RC underestimate shared tissue movement in these examples, whereas
\scopename{} preserves it while reducing spurious motion around local interactions.
APD at individual distance thresholds is reported in Appendix~\ref{app:tracking}.

\subsection{Ablation Studies}
\label{sec:ablation}

\input{tables/ablation}

\textbf{Data composition.}
To assess our annotation pipeline's value, we add StereoMIS (SM), EndoMapper-Real
(EM-R), or both to geometry-only SFT on five datasets: SCARED (SC), SimCol3D (SIM),
C3VD, EndoMapper-Sim (EM-S), and Sano (SA) (Table~\ref{tab:ablation}(a)).
ID averages cover these five datasets.
Both added sources improve depth accuracy on unseen Hamlyn data and provide
complementary gains in phantom camera estimation. Combining them yields the
strongest overall depth results among the variants. These results demonstrate
the value of \scopedata{}'s diverse geometric supervision for generalization
across endoscopic domains.

\textbf{Motion design.}
On StereoMIS (Table~\ref{tab:ablation}(b)), full CRM improves pose AUC@$5^\circ$
by 22.8\% and tracking APD by 13.0\% over SoM, with comparable depth accuracy.
Adding a zero-motion anchor and assignment supervision
improves EPE and APD; common--residual composition and its supervision further
improve pose AUC and APD. These results support motion modeling for both geometry
estimation and tissue tracking. Table~\ref{tab:ablation}(c)
shows reduced spurious motion in surrounding tissue while retaining deformation
near the interaction. Appendix~\ref{app:motion-results}
reports APD at all thresholds.

\subsection{Downstream Tasks and User Study}
\label{sec:long}

\input{tables/downstream_user_study}

\textbf{Long-sequence reconstruction.}
We evaluate on RealSynCol~\citep{lena2026realsyncol} using a framework similar to
VGGT-Long~\citep{deng2025vggtlong}. All models reconstruct overlapping chunks
and align them sequentially with Sim(3) transforms, without loop closure or global alignment.
Both \scopename{} variants outperform the baselines (Table~\ref{tab:long}).
Gain compares the full model with the best competing method for each metric.
The full model achieves lower ATE than geometric SFT, supporting joint geometry
and motion modeling for long-sequence camera estimation (Appendix Figure~\ref{fig:long}).

\phantomsection
\textbf{User study.}
\label{sec:user-study}
Six clinical and biomedical experts evaluate ten \scopereal{} clips with method identities hidden,
ranking four reconstructions separately for camera trajectories, point clouds,
and depth maps (1 best, 4 worst). Table~\ref{tab:user-study} reports mean ranks
with 95\% bootstrap confidence intervals in brackets.
\mbox{\scopename{}} achieves the best mean rank and ranks first in 70\% of
evaluations in each dimension.
In paired comparisons with pretrained VGGT-$\Omega$, participants prefer
\scopename{} in 75.0\%, 75.0\%, and 76.7\% of evaluations, respectively.
Appendix~\ref{app:user-study} details the protocol and analysis.

%% file: tables/main_geometry.tex
\begin{table*}[t]
\caption{\textbf{Camera pose and depth across domains.}
Hamlyn pose measures camera stability; phantom ARE$^*$ excludes roll.
Gain compares our best variant per metric with the best competing method.}
\label{tab:main-geometry}
\centering
\footnotesize
\setlength{\tabcolsep}{3pt}
\resizebox{\textwidth}{!}{%
\begin{tabular}{@{}lccccc|cccc|*{3}{>{\centering\arraybackslash}p{4.5em}}@{}}
\toprule
& \multicolumn{5}{c}{\textbf{In-domain average}}
& \multicolumn{4}{c}{\textbf{Hamlyn (OOD)}}
& \multicolumn{3}{c}{\textbf{\scopephantom{} (OOD)}} \\
\cmidrule(lr){2-6}\cmidrule(lr){7-10}\cmidrule(lr){11-13}
& \multicolumn{3}{c}{Pose} & \multicolumn{2}{c}{Depth}
& \multicolumn{2}{c}{Pose} & \multicolumn{2}{c}{Depth}
& \multicolumn{3}{c}{Pose} \\
\cmidrule(lr){2-4}\cmidrule(lr){5-6}\cmidrule(lr){7-8}\cmidrule(lr){9-10}\cmidrule(lr){11-13}
Method & AUC@$5^\circ\uparrow$ & AUC@$30^\circ\uparrow$ & ATE$\downarrow$
& AbsRel$\downarrow$ & $\delta_{1.25}\uparrow$
& ATE$\downarrow$
& ARE$\downarrow$
& AbsRel$\downarrow$ & $\delta_{1.25}\uparrow$
& ATE$\downarrow$ & RTE$\downarrow$ & ARE$^{*}\downarrow$ \\
\midrule
\multicolumn{13}{l}{\textit{General-domain}} \\
VGGT & 2.94 & 29.11 & 0.128 & 0.160 & 74.61 & 4.610 & 1.964 & 0.190 & 69.19 & 16.53 & 16.78 & 12.14 \\
VGGT-$\Omega$ & 7.81 & 48.07 & 0.049 & 0.093 & 91.07 & 4.496 & 1.746 & 0.135 & 80.82 & 8.54 & \underline{10.79} & 8.22 \\
MapAnything & 2.63 & 26.22 & 0.122 & 0.141 & 85.00 & 4.984 & 1.960 & 0.166 & 74.15 & 15.98 & 17.70 & 26.98 \\
CUT3R & 0.27 & 15.55 & 0.131 & 0.300 & 54.06 & 4.571 & 1.693 & 0.185 & 70.27 & 13.64 & 16.56 & 22.48 \\
SpatialTrackerV2 & 6.01 & 44.85 & 0.058 & 0.113 & 86.98 & 5.499 & 2.241 & 0.142 & 77.42 & 10.92 & 13.80 & 11.23 \\
4RC & 6.12 & 41.85 & 0.068 & 0.138 & 79.71 & 3.639 & 1.438 & \textbf{0.120} & \underline{84.26} & 13.98 & 15.33 & 9.13 \\
Any4D & 0.29 & 11.18 & 0.160 & 0.215 & 65.25 & 4.755 & 1.114 & 0.167 & 72.23 & 16.94 & 17.57 & 12.15 \\
SM4RT & 4.86 & 38.72 & 0.087 & 0.147 & 77.78 & 3.517 & 1.350 & \textbf{0.120} & 83.94 & 14.46 & 15.50 & 9.17 \\
\midrule
\multicolumn{13}{l}{\textit{Endoscopy-specific}} \\
Endo3R & 0.26 & 11.14 & 0.179 & 0.185 & 70.06 & 20.982 & 9.492 & 0.223 & 62.77 & 16.42 & 17.05 & 16.13 \\
AF-SfMLearner & 1.45 & 20.76 & 0.134 & 0.253 & 56.94 & 15.562 & 5.979 & 0.190 & 68.20 & 10.35 & 13.88 & 10.31 \\
EndoDAC & 1.41 & 19.08 & 0.130 & 0.201 & 66.22 & 18.057 & 8.968 & 0.191 & 68.87 & 11.51 & 15.03 & 10.54 \\
Endo-FASt3R & 4.13 & 35.02 & 0.062 & 0.230 & 60.39 & 9.152 & 3.794 & 0.211 & 64.75 & 8.70 & 11.39 & 9.22 \\
\midrule
\textbf{\boldmath\scopesft{}} & \underline{29.44} & \underline{67.33} & \textbf{0.016} & \underline{0.042} & \underline{98.52} & \textbf{2.927} & \underline{1.014} & 0.129 & 83.77 & \underline{8.41} & \textbf{10.71} & \underline{7.37} \\
\textbf{\ourmodel{}} & \textbf{29.86} & \textbf{67.41} & \underline{0.017} & \textbf{0.040} & \textbf{98.68} & \underline{2.976} & \textbf{1.013} & \underline{0.124} & \textbf{85.23} & \textbf{8.29} & 10.82 & \textbf{7.21} \\
\midrule
Gain (\%) & +282.2\% & +40.2\% & +67.2\% & +56.8\% & +8.4\% & +16.8\% & +9.0\% & -3.6\% & +1.2\% & +2.9\% & +0.7\% & +12.3\% \\
\bottomrule
\end{tabular}}
\vspace{-1em}
\end{table*}

%% file: tables/tracking.tex
\begin{table}[t]
\caption{\textbf{3D tracking and geometry on Sano and StereoMIS.}
D4RT$^*$: open-source reproduction.}
\label{tab:tracking}
\centering
\footnotesize
\setlength{\tabcolsep}{3pt}
\resizebox{\textwidth}{!}{%
\begin{tabular}{@{}lccccccc|ccccccc@{}}
\toprule
& \multicolumn{7}{c}{\textbf{Sano}}
& \multicolumn{7}{c}{\textbf{StereoMIS}} \\
\cmidrule(lr){2-8}\cmidrule(lr){9-15}
& \multicolumn{2}{c}{Tracking} & \multicolumn{3}{c}{Pose} & \multicolumn{2}{c}{Depth}
& \multicolumn{2}{c}{Tracking} & \multicolumn{3}{c}{Pose} & \multicolumn{2}{c}{Depth} \\
\cmidrule(lr){2-3}\cmidrule(lr){4-6}\cmidrule(lr){7-8}\cmidrule(lr){9-10}\cmidrule(lr){11-13}\cmidrule(lr){14-15}
Method & EPE$\downarrow$ & APD$\uparrow$ & ATE$\downarrow$ & RTE$\downarrow$ & RRE$\downarrow$ & AbsRel$\downarrow$ & $\delta_{1.25}\uparrow$
& EPE$\downarrow$ & APD$\uparrow$ & ATE$\downarrow$ & RTE$\downarrow$ & RRE$\downarrow$ & AbsRel$\downarrow$ & $\delta_{1.25}\uparrow$ \\
\midrule
D4RT$^{*}$ & 2.035 & 72.94 & 0.014 & 0.239 & 0.215 & 0.193 & 65.76 & 7.540 & 34.12 & 0.043 & 0.940 & 0.344 & 0.147 & 87.20 \\
SpatialTrackerV2 & 0.933 & 90.58 & 0.017 & 0.223 & 0.125 & \underline{0.145} & \underline{78.76} & 6.870 & 38.98 & 0.049 & 0.677 & 0.195 & \underline{0.107} & \underline{95.24} \\
4RC & 0.523 & 94.72 & \underline{0.006} & \underline{0.100} & \underline{0.062} & 0.186 & 66.15 & \underline{6.266} & \underline{44.28} & 0.039 & 0.481 & \underline{0.146} & 0.114 & 90.42 \\
Any4D & 0.535 & 94.41 & 0.010 & 0.376 & 0.309 & 0.188 & 66.01 & 7.277 & 42.45 & 0.061 & 2.203 & 0.765 & 0.144 & 82.72 \\
SM4RT & \textbf{0.411} & \underline{95.64} & 0.007 & 0.104 & 0.081 & 0.182 & 67.31 & 6.276 & 43.95 & \underline{0.025} & \underline{0.427} & 0.158 & 0.116 & 92.57 \\
\midrule
\textbf{\ourmodel{}} & \underline{0.442} & \textbf{96.11} & \textbf{0.001} & \textbf{0.054} & \textbf{0.051} & \textbf{0.033} & \textbf{99.75} & \textbf{5.022} & \textbf{53.08} & \textbf{0.010} & \textbf{0.211} & \textbf{0.114} & \textbf{0.050} & \textbf{98.80} \\
\midrule
Gain (\%) & -7.5\% & +0.5\% & +81.6\% & +45.9\% & +18.4\% & +77.1\% & +26.7\% & +19.8\% & +19.9\% & +59.4\% & +50.7\% & +21.9\% & +53.3\% & +3.7\% \\
\bottomrule
\end{tabular}}
\vspace{-1em}
\end{table}

%% file: tables/ablation.tex
\begin{table}[t]
\caption{\textbf{Data and motion ablations.}
(a) Geometry-only SFT by data source. (b) StereoMIS motion design.
(c) SoM vs.\ CRM. Gain in (a): full dataset versus pretrained VGGT-$\Omega$.}
\label{tab:ablation}
\centering
\begingroup
\setlength{\parskip}{0pt}
\newsavebox{\dataablationbox}
\newsavebox{\motionablationbox}
\sbox{\dataablationbox}{%
\begin{minipage}[b]{\linewidth}
\centering
\scriptsize
\setlength{\tabcolsep}{3.5pt}
\input{tables/data_ablation}
\end{minipage}}
\sbox{\motionablationbox}{%
\begin{minipage}[b]{0.47\linewidth}
\centering
\fontsize{8}{9.5}\selectfont
\setlength{\tabcolsep}{1.5pt}
\input{tables/motion_ablation_transposed}
\end{minipage}}
{\fontsize{8}{9.5}\selectfont
\makebox[\linewidth]{\strut\textbf{(a) Data composition}}\par}
\nointerlineskip\vspace{4pt}
\noindent\raisebox{\dp\dataablationbox}{\usebox{\dataablationbox}}\par
\vspace{6pt}
{\fontsize{8}{9.5}\selectfont
\makebox[0.47\linewidth]{\strut\textbf{(b) Motion design}}\hfill
\makebox[0.51\linewidth]{\strut\textbf{(c) Qualitative motion comparison}}\par}
\nointerlineskip\vspace{4pt}
\noindent\raisebox{\dp\motionablationbox}{\usebox{\motionablationbox}}\hfill
\makebox[0.51\linewidth]{%
\includegraphics[height=\dimexpr\ht\motionablationbox+\dp\motionablationbox\relax]{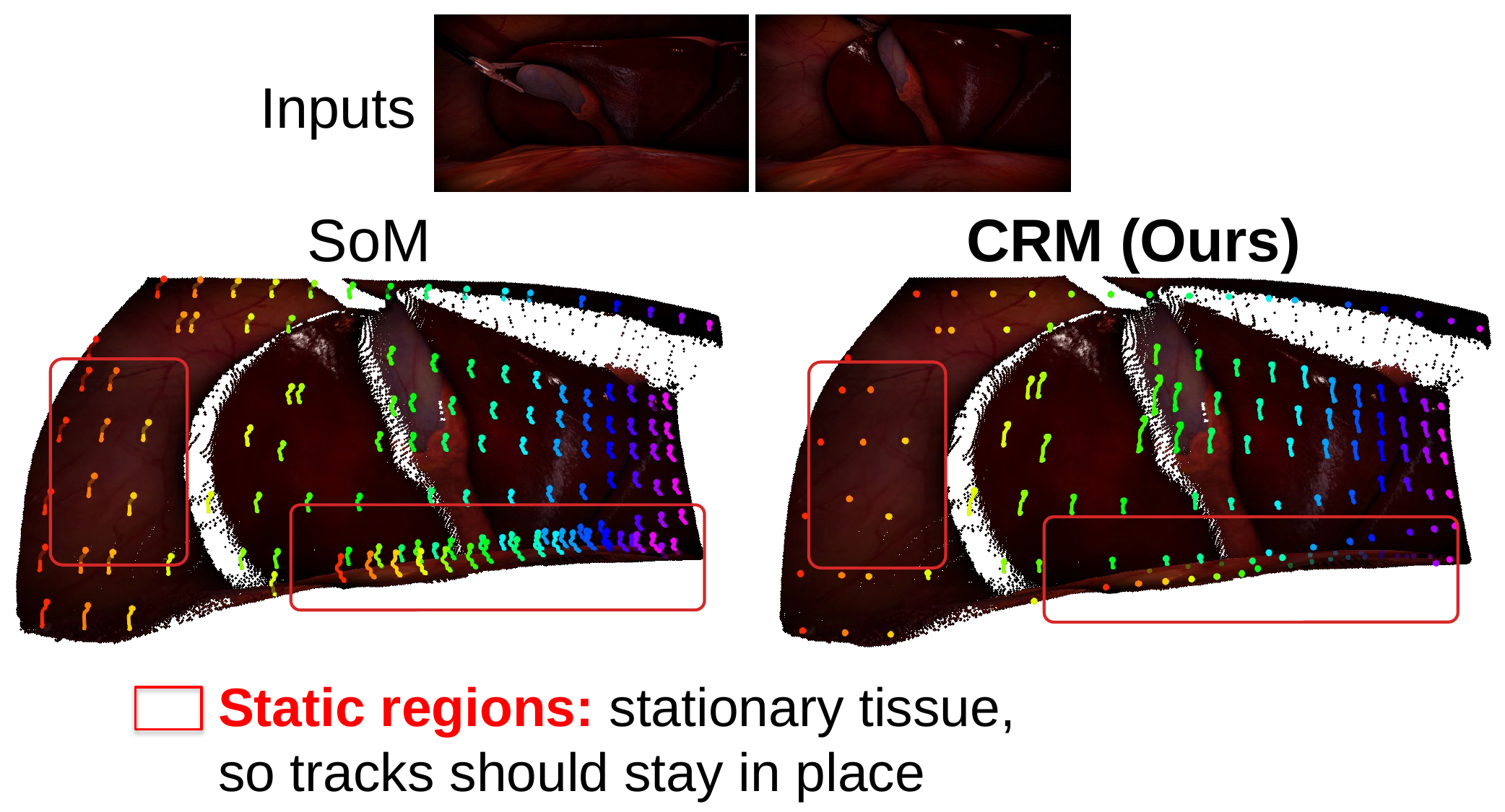}}\par
\endgroup
\vspace{-1em}
\end{table}

%% file: tables/data_ablation.tex
\resizebox{\linewidth}{!}{%
\begin{tabular}{@{}lcccc|cc|*{2}{>{\centering\arraybackslash}p{6em}}@{}}
\toprule
& \multicolumn{4}{c}{\textbf{In-domain average}}
& \multicolumn{2}{c}{\textbf{Hamlyn (OOD)}}
& \multicolumn{2}{c}{\textbf{\scopephantom{} (OOD)}} \\
\cmidrule(lr){2-5}\cmidrule(lr){6-7}\cmidrule(lr){8-9}
& \multicolumn{2}{c}{Pose} & \multicolumn{2}{c}{Depth}
& \multicolumn{2}{c}{Depth} & \multicolumn{2}{c}{Pose} \\
\cmidrule(lr){2-3}\cmidrule(lr){4-5}\cmidrule(lr){6-7}\cmidrule(lr){8-9}
& AUC@$5^\circ\uparrow$ & AUC@$30^\circ\uparrow$
& AbsRel$\downarrow$ & $\delta_{1.25}\uparrow$
& AbsRel$\downarrow$ & $\delta_{1.25}\uparrow$ & RTE$\downarrow$ & ARE$^{*}\downarrow$ \\
\midrule
No finetune (VGGT-$\Omega$ weights) & 9.55 & 53.03 & 0.077 & 94.03 & 0.135 & 80.82 & 10.79 & 8.22 \\
\midrule
w/ SC+SIM+C3VD+EM-S+SA & 35.99 & \textbf{74.08} & \underline{0.031} & 99.50 & 0.143 & 80.93 & 11.28 & \underline{7.12} \\
w/ SC+SIM+C3VD+EM-S+SA+SM & 33.10 & 70.02 & 0.032 & 99.53 & 0.142 & 81.40 & \underline{10.51} & \textbf{6.83} \\
w/ SC+SIM+C3VD+EM-S+SA+EM-R & \textbf{36.62} & \underline{74.07} & 0.032 & \underline{99.57} & \textbf{0.128} & \underline{83.04} & \textbf{10.37} & 7.37 \\
\textbf{Full dataset} & \underline{36.02} & 73.68 & \textbf{0.030} & \textbf{99.59} & \underline{0.129} & \textbf{83.77} & 10.71 & 7.37 \\
\midrule
Gain (\%) vs No finetune & +277.1\% & +38.9\% & +61.6\% & +5.9\% & +4.3\% & +3.7\% & +0.7\% & +10.4\% \\
\bottomrule
\end{tabular}}

%% file: tables/motion_ablation_transposed.tex
\begin{tabular*}{\linewidth}{@{\extracolsep{\fill}}lcccc@{}}
\toprule
Metric & \shortstack{SoM\\(baseline)} & + anchor & \shortstack{\textbf{Full model}\\+ CRM} & Gain (\%) \\
\midrule
EPE$\downarrow$ & 5.94 & \underline{5.30} & \textbf{5.27} & +11.2\% \\
APD$\uparrow$ & 46.58 & \underline{50.11} & \textbf{52.63} & +13.0\% \\
\midrule
AUC@$5^\circ\uparrow$ & 5.67 & \underline{6.09} & \textbf{6.96} & +22.8\% \\
AUC@$30^\circ\uparrow$ & 26.86 & \underline{27.17} & \textbf{27.66} & +3.0\% \\
\midrule
AbsRel$\downarrow$ & 0.050 & \textbf{0.048} & \underline{0.049} & +0.3\% \\
$\delta_{1.25}\uparrow$ & \underline{98.82} & \underline{98.82} & \textbf{98.85} & +0.0\% \\
\bottomrule
\end{tabular*}

%% file: tables/downstream_user_study.tex
\begin{table}[t]
\centering
\begingroup
\makeatletter
\long\def\@makecaption#1#2{%
  \vskip\abovecaptionskip
  {\raggedright #1: #2\par}%
  \vskip\belowcaptionskip}
\makeatother
\newsavebox{\scopeLongCaption}
\newsavebox{\scopeStudyCaption}
\newsavebox{\scopeLongBody}
\newsavebox{\scopeStudyBody}
\newlength{\scopePairCaptionHeight}
\newlength{\scopePairBodyHeight}
\newlength{\scopeLongRowExtra}
\newlength{\scopeStudyRowExtra}
\setlength{\scopeLongRowExtra}{0pt}
\setlength{\scopeStudyRowExtra}{0pt}
\newcommand{\scopeLongTable}{%
  \begin{tabular*}{\linewidth}{@{\extracolsep{\fill}}lccc@{}}
  \toprule
  Method & ATE$\downarrow$ & RTE$\downarrow$ & RRE$\downarrow$ \\[\scopeLongRowExtra]
  \midrule
  VGGT & 93.21 & 23.65 & 15.11 \\[\scopeLongRowExtra]
  VGGT-$\Omega$ & 85.92 & 21.67 & 31.21 \\[\scopeLongRowExtra]
  SM4RT & 88.95 & 21.29 & 16.98 \\[\scopeLongRowExtra]
  Endo3R & 77.87 & 14.75 & 22.92 \\[\scopeLongRowExtra]
  \midrule
  \textbf{\boldmath\scopesft{}} & \underline{73.52} & \textbf{10.38} & \textbf{13.38} \\[\scopeLongRowExtra]
  \textbf{\ourmodel{}} & \textbf{71.17} & \underline{10.94} & \underline{13.61} \\[\scopeLongRowExtra]
  \midrule
  Gain (\%) & +8.6\% & +25.9\% & +9.9\% \\[\scopeLongRowExtra]
  \bottomrule
  \end{tabular*}}
\newcommand{\scopeRankCell}[2]{%
  \begin{tabular}[t]{@{}c@{}}#1\\[-1pt]{\scriptsize[#2]}\end{tabular}}
\newcommand{\scopeStudyTable}{%
  \begin{tabular*}{\linewidth}{@{\extracolsep{\fill}}lccc@{}}
  \toprule
  Method & Camera$\downarrow$ & Point$\downarrow$ & Depth$\downarrow$ \\[\scopeStudyRowExtra]
  \midrule
  VGGT-$\Omega$ & \scopeRankCell{1.93}{1.65, 2.25} & \scopeRankCell{1.95}{1.65, 2.28} & \scopeRankCell{1.98}{1.67, 2.32} \\[\scopeStudyRowExtra]
  SM4RT & \scopeRankCell{2.90}{2.60, 3.08} & \scopeRankCell{3.22}{2.80, 3.57} & \scopeRankCell{3.33}{2.87, 3.73} \\[\scopeStudyRowExtra]
  Endo3R & \scopeRankCell{3.83}{3.53, 4.00} & \scopeRankCell{3.52}{3.23, 3.78} & \scopeRankCell{3.37}{3.00, 3.70} \\[\scopeStudyRowExtra]
  \midrule
  \textbf{\ourmodel{}} & \scopeRankCell{\textbf{1.33}}{1.12, 1.62} & \scopeRankCell{\textbf{1.32}}{1.13, 1.55} & \scopeRankCell{\textbf{1.32}}{1.13, 1.53} \\[\scopeStudyRowExtra]
  \bottomrule
  \end{tabular*}}
\sbox{\scopeLongCaption}{\begin{minipage}[t]{0.465\linewidth}
  \caption{\textbf{Long-sequence reconstruction.}}
  \label{tab:long}
\end{minipage}}
\sbox{\scopeStudyCaption}{\begin{minipage}[t]{0.51\linewidth}
  \caption{\textbf{Clinical expert assessment.}}
  \label{tab:user-study}
\end{minipage}}
\setlength{\scopePairCaptionHeight}{\dimexpr\ht\scopeLongCaption+\dp\scopeLongCaption\relax}
\ifdim\dimexpr\ht\scopeStudyCaption+\dp\scopeStudyCaption\relax>\scopePairCaptionHeight
  \setlength{\scopePairCaptionHeight}{\dimexpr\ht\scopeStudyCaption+\dp\scopeStudyCaption\relax}
\fi
\setlength{\tabcolsep}{2pt}
\sbox{\scopeLongBody}{\begin{minipage}{0.465\linewidth}\footnotesize\scopeLongTable\end{minipage}}
\sbox{\scopeStudyBody}{\begin{minipage}{0.51\linewidth}\footnotesize\scopeStudyTable\end{minipage}}
\setlength{\scopePairBodyHeight}{\dimexpr\ht\scopeStudyBody+\dp\scopeStudyBody-\ht\scopeLongBody-\dp\scopeLongBody\relax}
\ifdim\scopePairBodyHeight>0pt
  \setlength{\scopeLongRowExtra}{\dimexpr\scopePairBodyHeight/8\relax}
\else
  \setlength{\scopeStudyRowExtra}{\dimexpr-\scopePairBodyHeight/5\relax}
\fi
\begin{minipage}[t]{0.465\linewidth}
  \vspace{0pt}%
  \begin{minipage}[t][\scopePairCaptionHeight][t]{\linewidth}
    \usebox{\scopeLongCaption}
  \end{minipage}\par
  \footnotesize\scopeLongTable
\end{minipage}\hfill
\begin{minipage}[t]{0.51\linewidth}
  \vspace{0pt}%
  \begin{minipage}[t][\scopePairCaptionHeight][t]{\linewidth}
    \usebox{\scopeStudyCaption}
  \end{minipage}\par
  \footnotesize\scopeStudyTable
\end{minipage}
\endgroup
\vspace{-1em}
\end{table}

%% file: sections/06_conclusion.tex
\section{Conclusion}
We present \scopename{}, an endoscopic 4D geometry foundation model that jointly
predicts camera parameters, dense geometry, and 3D tissue tracks from monocular
RGB video. Our curation and annotation pipeline constructs \scopedata{} to
expand geometric supervision and evaluation coverage. Geometric SFT learns
endoscopic priors from this data, while CRM separates common tissue movement
from local deformation. Together with trajectory supervision, this representation
further improves geometry over geometric SFT alone while enabling dense tissue
tracking. Experiments demonstrate accurate geometry across ID and OOD benchmarks,
improved 3D tracking, and more stable long-sequence reconstruction. A blinded
study with six clinical and biomedical experts also favors \scopename{} on real clinical video.
The phantom and clinical benchmarks assess tracked camera accuracy and perceived
reconstruction quality, respectively.
These findings support combining broader endoscopic supervision with explicit
tissue-motion modeling for feed-forward 4D reconstruction.

%% file: sections/appendix.tex
\raggedbottom
\makeatletter
\setlength{\@fptop}{0pt}
\setlength{\@fpsep}{12pt}
\setlength{\@fpbot}{0pt plus 1fil}
\makeatother
\suppressfloats[t]
\section*{Appendix}
In the appendix, we provide the following:
\begin{itemize}
  \setlength{\itemsep}{0pt}
  \setlength{\parsep}{0pt}
  \item Section~\ref{app:data}: Dataset construction.
  \item Section~\ref{app:implementation}: Implementation details.
  \item Section~\ref{app:crm-proof}: Proof of the CRM residual bound.
  \item Section~\ref{app:evaluation}: Evaluation protocols.
  \item Section~\ref{app:results}: Additional results.
\end{itemize}

\input{sections/appendix_data}

\FloatBarrier
\input{sections/appendix_implementation}

\FloatBarrier
\input{sections/appendix_proof}

\FloatBarrier
\input{sections/appendix_evaluation}

\FloatBarrier
\input{sections/appendix_results}

%% file: sections/appendix_data.tex
\section{Dataset Construction}
\label{app:data}

\subsection{Data Composition and Splits}
\label{app:splits}
\scopedata{} covers real and synthetic GI endoscopy and laparoscopy.
Table~\ref{tab:scope5k-overview} lists sources and annotation coverage;
Table~\ref{tab:datasets} gives collection sizes, disjoint training,
validation, and test splits, and sampling proportions.
SCARED and EndoMapper-Real hold out source recordings for testing;
C3VD holds out the descending region.
SimCol3D and EndoMapper-Sim split by synthetic sequence, StereoMIS by
non-overlapping temporal segment, and Sano by simulation attempt.

\input{tables/scope5k_overview}
\input{tables/dataset_comparison}

\textbf{Annotation sources.}
SCARED provides structured-light-derived 3D points, camera poses, and
calibration~\citep{allan2021scared}. C3VD provides depth, camera poses, and
calibration registered to phantom geometry~\citep{bobrow2023c3vd,golhar2026c3vdv2}.
SimCol3D and EndoMapper-Sim provide rendered depth, camera poses, and
intrinsics~\citep{rau2024simcol3d,azagra2023endomapper}.
Sano denotes the SANO\_VR\_Cholecystectomy subset of
Open-H-Embodiment~\citep{openh2026}, which provides rendered depth, camera
parameters, and segmentation labels. The contributor's dataset card provides
the source description and access link~\citep{SANO_VR_Cholecystectomy_2025}.
We use these annotations to supervise depth, cameras, and pointmaps.

EndoMapper-Real provides calibrated clinical video without dense geometric ground
truth~\citep{azagra2023endomapper}, while StereoMIS provides calibrated stereo
video and camera poses from robot kinematics~\citep{hayoz2023pose}.
Our pipeline adds cameras and dense geometry to EndoMapper-Real and dense depth to
StereoMIS to supervise depth, cameras, and pointmaps.

For evaluation, Hamlyn provides depth and calibration but no camera-pose
ground truth for the selected clips~\citep{recasens2021endodm}; RealSynCol provides
rendered depth, camera trajectories, and intrinsics~\citep{lena2026realsyncol}.
\scopephantom{} provides tracked camera positions and viewing directions;
\scopereal{} supports specialist reconstruction assessment.

\subsection{EndoMapper-Real Curation and Annotation}
\label{app:curation}
We convert raw clinical video into reconstructable clips with camera and dense
geometry annotations (Algorithm~\ref{alg:annotation-pipeline}). Gemini 3.5 Flash
acts as a VLM judge, assessing tissue visibility, camera travel, obstruction, and
image quality with the prompt below.

\input{figures/appendix_annotation_workflow}

\textbf{Clip selection.}
Frame-level filters refine the VLM-selected segments by removing blur, exposure
failures, strong highlights, and frames with insufficient texture or tissue
visibility. We then split the retained frames at abrupt viewpoint changes and
temporal gaps to form coherent 50--300-frame clips. Source frame indices keep the clips aligned with their annotations.

\textbf{Geometric annotation.}
For camera reconstruction in low-texture tissue, MFT tracks provide the primary
SfM correspondences~\citep{neoral2024mft}. Independent reconstructions from LoMa
and SIFT with LightGlue allow us to check camera agreement across correspondence
sources~\citep{nordstrom2026loma,lowe2004sift,lindenberger2023lightglue}.
MapAnything then uses the recovered cameras and undistorted images to predict
dense depth and point clouds~\citep{keetha2026mapanything}. Pixel masks exclude
unreliable predictions, depth-edge artifacts, and regions without multi-view
support. Clip-level geometric checks then determine which annotations to retain.

\clearpage
\input{sections/appendix_vlm_prompt}
\clearpage

\subsection{StereoMIS Curation and Annotation}
\label{app:stereomis-curation}
For calibrated StereoMIS videos, we use the provided camera poses and estimate
depth from stereo pairs. Clip selection follows the same semantic screening and
frame filtering as EndoMapper-Real, without requiring camera travel.
Non-white-light segments are excluded, and source frame indices keep stereo
images, poses, and masks aligned.

We rectify each stereo pair and estimate disparity with Stereo Any Video
(SAV)~\citep{jing2025stereoanyvideo}. Depth follows
$d=f_xb/\mathrm{disparity}$, where $f_x$ is the horizontal focal length and $b$
is the stereo baseline. Camera poses from robot kinematics are expressed relative
to the first camera in each clip; instrument masks use the rectified image grid.
This yields clips with aligned depth and camera annotations.

\subsection{Quality Control}
\label{app:pipeline}
Table~\ref{tab:qc} lists the EndoMapper-Real QC checks and thresholds.
A hard-gate failure yields Reject; a soft flag without a hard failure yields
Borderline; passing all checks yields Accept. Training uses Accept and
Borderline clips, while held-out evaluation uses Accept clips.

\begin{table}[H]
\centering\small
\caption{\textbf{Clip-level geometry checks.}}
\label{tab:qc}
\begin{tabularx}{\linewidth}{@{}lXl@{}}
\toprule
Check & Criterion & Role \\
\midrule
Registration & One significant reconstruction; at least 90\% of frames registered & Hard gate \\
Integrity & Readable depth, consistent frame counts, finite cameras & Hard gate \\
Connectivity & Largest point-cloud component at least 90\% & Hard gate \\
Coverage & Valid depth over at least 20\% of clip pixels & Hard gate \\
Parallax & Median maximum sparse-point parallax at least $1.5^\circ$ & Soft flag \\
Camera adoption & Dense-model camera change at most 1.5\% translation and $1^\circ$ rotation & Soft flag \\
Sparse agreement & Median scale-aligned relative depth discrepancy at most 0.20; scale drift at most 0.25 & Soft flag \\
Outliers & At most 10\% of points exceed the mean eight-neighbor distance by two standard deviations & Soft flag \\
Near-zero baseline & Flag when at least 40\% of frames have a translation-induced parallax proxy below $0.004^\circ$ & Soft flag \\
Cross-source poses & Reject if at least two alternatives are usable and all have aligned trajectory residuals above 10\% & Hard gate \\
 & Flag if only one alternative is usable and its aligned trajectory residual exceeds 10\% & Soft flag \\
\bottomrule
\end{tabularx}
\end{table}

Sparse--dense discrepancy is the median absolute relative depth difference
following median scale alignment; scale drift is the interquartile range of
per-view scale ratios divided by their median. Connectivity uses 26-neighbor
voxels of width $1/64$ of the cloud extent between coordinate-wise 2nd and 98th
percentiles. Cross-source residuals measure mean camera-center distance after
Sim(3) alignment, normalized by the alternative trajectory's bounding-box
diagonal, using at least eight shared frames. Unavailable comparisons do not
affect the QC label.

Table~\ref{tab:curation-yield} reports clip and frame counts. Excluding 11
Borderline clips from held-out recordings leaves 2,379 retained clips.
Human verification assigned a Reject label to 28 of the 2,379 clips (1.18\%):
2 previously labeled Accept and 26 Borderline.
Figure~\ref{fig:qc-examples} shows Borderline cases with single-source camera
disagreement or sparse--dense depth inconsistency, and Reject cases with
disagreement across both alternatives or insufficient camera registration.

\begin{table}[!ht]
\centering\small
\caption{\textbf{EndoMapper-Real curation counts.}
Frame counts precede training conversion.}
\label{tab:curation-yield}
\begin{tabular}{@{}lrr@{}}
\toprule
Stage & Clips & Frames \\
\midrule
Raw videos (93 recordings) & -- & 3,973,969 \\
Clean clips (Stage 1) & 3,034 & 319,727 \\
\quad Accept & 1,733 & 177,358 \\
\quad Borderline & 657 & 70,501 \\
\quad Reject & 644 & 71,868 \\
\midrule
Retained collection & 2,379 & 246,981 \\
\bottomrule
\end{tabular}
\end{table}
\input{figures/appendix_qc_examples}
\FloatBarrier

\subsection{SCOPE-colon Benchmark}
\label{app:inhouse}
\scopephantom{} contains 14 physical colon phantom clips with 128 evaluated
views each. Checkerboard calibration provides intrinsics; a five-degree-of-freedom
electromagnetic tracker records camera positions and viewing directions,
but not rotation about the viewing axis (roll).

\scopereal{} contains ten clinical colonoscopy clips,
with 128 views per clip, for specialist assessment of reconstruction quality.
Appendix~\ref{app:user-study} describes the study interface and ranking protocol.

\textbf{Ethical considerations.}
All data were obtained under institutional review board (IRB) approval.
Clinical recordings serve only to evaluate reconstruction quality retrospectively;
the study does not alter clinical procedures or patient care.

%% file: tables/scope5k_overview.tex
\begin{table}[!ht]
\centering
\small
\setlength{\tabcolsep}{3pt}
\caption{\textbf{Data composition and annotation coverage of \scopedata{}.}
Annotations reflect those available in \scopedata{}.
$\dagger$ denotes annotations generated by our pipeline.}
\label{tab:scope5k-overview}
\begin{tabularx}{\linewidth}{@{}Xccccccr@{}}
\toprule
\multirow{2}{*}{Data source} & \multicolumn{2}{c}{Domain} &
\multicolumn{2}{c}{Data type} & \multicolumn{2}{c}{Annotations} & \multirow{2}{*}{Clips} \\
\cmidrule(lr){2-3}\cmidrule(lr){4-5}\cmidrule(lr){6-7}
 & GI & Laparo & Synthetic & Real & Depth & Camera & \\
\midrule
EndoMapper-Real~\citep{azagra2023endomapper} & $\checkmark$ & -- & -- & $\checkmark$ & $\checkmark^{\dagger}$ & $\checkmark^{\dagger}$ & 2,379 \\
C3VD~\citep{bobrow2023c3vd} & $\checkmark$ & -- & -- & $\checkmark$ & $\checkmark$ & $\checkmark$ & 112 \\
SimCol3D~\citep{rau2024simcol3d} & $\checkmark$ & -- & $\checkmark$ & -- & $\checkmark$ & $\checkmark$ & 33 \\
EndoMapper-Sim~\citep{azagra2023endomapper} & $\checkmark$ & -- & $\checkmark$ & -- & $\checkmark$ & $\checkmark$ & 5 \\
Sano~\citep{openh2026} & -- & $\checkmark$ & $\checkmark$ & -- & $\checkmark$ & $\checkmark$ & 1,938 \\
StereoMIS~\citep{hayoz2023pose} & -- & $\checkmark$ & -- & $\checkmark$ & $\checkmark^{\dagger}$ & $\checkmark$ & 354 \\
SCARED~\citep{allan2021scared} & -- & $\checkmark$ & -- & $\checkmark$ & $\checkmark$ & $\checkmark$ & 34 \\
Hamlyn~\citep{hamlyn_dataset} & -- & $\checkmark$ & -- & $\checkmark$ & $\checkmark$ & -- & 22 \\
RealSynCol~\citep{lena2026realsyncol} & $\checkmark$ & -- & $\checkmark$ & -- & $\checkmark$ & $\checkmark$ & 9 \\
\scopephantom{} & $\checkmark$ & -- & -- & $\checkmark$ & -- & $\checkmark$ & 14 \\
\scopereal{} & $\checkmark$ & -- & -- & $\checkmark$ & -- & -- & 10 \\
\midrule
\textbf{\scopedata{} (Ours)} & $\checkmark$ & $\checkmark$ & $\checkmark$ & $\checkmark$ & $\checkmark$ & $\checkmark$ & \textbf{4,910} \\
\bottomrule
\end{tabularx}
\end{table}

%% file: tables/dataset_comparison.tex
\begin{table}[!ht]
\setlength{\abovecaptionskip}{3pt}
\setlength{\belowcaptionskip}{2pt}
\caption{\textbf{Dataset sizes, splits, and sampling proportions.}}
\label{tab:datasets}
\centering
\small
\setlength{\tabcolsep}{3pt}
\begin{tabular*}{\linewidth}{@{\extracolsep{\fill}}lrrrrrr@{}}
\toprule
\multirow{2}{*}{Dataset} & \multicolumn{2}{c}{Collected} & \multicolumn{3}{c}{Experimental split} & \multirow{2}{*}{Sampling (\%)} \\
\cmidrule(lr){2-3}\cmidrule(lr){4-6}
 & Clips & Views & Train & Val. & Test & \\
\midrule
\multicolumn{7}{@{}l}{\textit{In-domain}} \\
SCARED & 34 & 24,046 & 24 & 2 & 8 & 13.86 \\
SimCol3D & 33 & 37,833 & 22 & 2 & 9 & 13.25 \\
C3VD & 112 & 43,164 & 99 & 2 & 11 & 18.07 \\
EndoMapper-Sim & 5 & 1,610 & 3 & 1 & 1 & 3.01 \\
EndoMapper-Real & 2,379 & 184,796 & 2,347 & 4 & 28 & 24.10 \\
StereoMIS & 354 & 54,770 & 346 & 2 & 6 & 13.86 \\
Sano & 1,938 & 223,761 & 519 & 63 & 64 & 13.86 \\
\midrule
\multicolumn{7}{@{}l}{\textit{Out-of-domain}} \\
Hamlyn & 22 & 3,300 & 0 & 0 & 22 & 0 \\
\textbf{\scopephantom{}} & 14 & 1,792 & 0 & 0 & 14 & 0 \\
RealSynCol & 9 & 2,680 & 0 & 0 & 9 & 0 \\
\textbf{\scopereal{}} & 10 & 1,280 & 0 & 0 & 10 & 0 \\
\midrule
\textbf{Total} & \textbf{4,910} & \textbf{579,032} & \textbf{3,360} & \textbf{76} & \textbf{182} & \textbf{100.00} \\
\bottomrule
\end{tabular*}
\end{table}

%% file: figures/appendix_annotation_workflow.tex
\begin{algorithm}[t]
\caption{EndoMapper-Real curation and annotation}
\label{alg:annotation-pipeline}
\begin{lstlisting}[language=Python,basicstyle=\ttfamily\footnotesize,
  keywordstyle=\bfseries,commentstyle=\normalfont\itshape,
  numbers=left,numberstyle=\tiny,numbersep=7pt,xleftmargin=1.8em,
  columns=fullflexible,keepspaces=true,showstringspaces=false,
  aboveskip=4pt,belowskip=4pt]
def annotate(videos, held_out):
    dataset = []
    for video in videos:
        segments = vlm_judge(video)
        for clip in filter_and_split(segments):
            cameras = sfm(clip, matcher=MFT)
            if not valid(cameras):
                continue
            alternatives = [sfm(clip, matcher=m)
                            for m in (LoMa, SIFT_LightGlue)]
            rgb = undistort(clip, cameras)
            depth, points = mapanything(rgb, cameras)
            mask = valid_mask(depth, points, cameras)
            status = qc(cameras, alternatives, depth, points, mask)
            if status == REJECT:
                continue
            if video in held_out and status == BORDERLINE:
                continue
            dataset.append((rgb, cameras, depth, points, mask))
    return dataset
\end{lstlisting}
\end{algorithm}

%% file: sections/appendix_vlm_prompt.tex

\textbf{VLM prompt.}
We use the following prompt to define the screening criteria and output format.
\par\medskip
\begingroup
\color[rgb]{0.18,0.55,0.27}
\hrule height 0.4pt
\vspace{1pt}
\hrule height 0.4pt
\nobreak\smallskip
\footnotesize\ttfamily\raggedright
\setlength{\parindent}{0pt}
\setlength{\parskip}{1pt}

You are segmenting a FULL colonoscopy video into coarse SEMANTIC CHAPTERS and SCORING each chapter for whether it can be used to 3D-RECONSTRUCT the colon. Watch the entire video, then return ONLY a JSON array (no prose). Each element:

\{"start":\allowbreak{}"M:\allowbreak{}SS",\allowbreak{}"end":\allowbreak{}"M:\allowbreak{}SS",\allowbreak{}"state":\allowbreak{}"\textless{}one of:\allowbreak{} outside\_body,\allowbreak{} insertion,\allowbreak{} withdrawal\_inspection,\allowbreak{} wall\_redout,\allowbreak{} stationary,\allowbreak{} cleaning\_irrigation,\allowbreak{} therapy\_tool,\allowbreak{} unclear\textgreater{}",\allowbreak{} "scores":\allowbreak{}\{"structure\_visibility":\allowbreak{}0-3,\allowbreak{}"camera\_travel":\allowbreak{}0-3,\allowbreak{}"unobstructed":\allowbreak{}0-3,\allowbreak{}"image\_quality":\allowbreak{}0-3\},\allowbreak{} "reasons":\allowbreak{}\{"structure\_visibility":\allowbreak{}"\textless{}=8 words",\allowbreak{}"camera\_travel":\allowbreak{}"\textless{}=8 words",\allowbreak{}"unobstructed":\allowbreak{}"\textless{}=8 words",\allowbreak{}"image\_quality":\allowbreak{}"\textless{}=8 words"\}\}

PRINCIPLES for 3D reconstruction: a segment is reconstructable only if textured colon structure is in view AND the scope TRANSLATES (producing parallax). Brightness is NOT geometry -- specular highlights and the wet non-Lambertian sheen MOVE WITH the scope and are noise, not structure; judge structure by the lumen recess and haustral-fold ridges, NOT by how bright/wet a frame looks. Pure rotation (no translation) gives zero baseline = unreconstructable. A featureless red wall has no trackable texture even under motion. The colon deforms (peristalsis / CO2 insufflation), so shorter contiguous segments are preferred.

Rules: MERGE adjacent same-activity moments into ONE chapter (\textasciitilde{}5-20, temporal order, cover the whole timeline). Score each chapter 0-3 on FOUR INDEPENDENT axes (do NOT fuse them; do NOT output keep/discard):

  structure\_visibility -- is the colon LUMEN / haustral folds / textured mucosa VISIBLE? Judge by GEOMETRY (a DARK lumen recess / forward tunnel, concentric haustral-fold ridges, submucosal vessels), NOT by brightness, and INDEPENDENT of travel direction (insertion or withdrawal). A bright, wet, sharp frame full of mucosa can still have NO lumen (scope pressed on the wall) = 0.

      0 = 'red-out' / wall-apposition: scope pressed on / pointed straight at the mucosa, uniform red-orange, no lumen recess

      1 = lumen only glimpsed occasionally; mostly wall

      2 = lumen / haustral folds in view for MOST of the segment

      3 = clear lumen tunnel with concentric folds in view throughout

  camera\_travel -- does the scope TRANSLATE along the colon (advancing OR withdrawing), giving a baseline / parallax?

      0 = no net travel: static, dwelling on one spot, OR retroflexion (tip angulated \textasciitilde{}180 deg to view backward, scope shaft often visible) = rotation in place -- score 0 EVEN IF the lumen is clearly visible

      1 = small wobble, little net advance or withdrawal

      2 = clear advance OR withdrawal through the lumen

      3 = sweeps a long extent of colon (much new tissue continuously enters/leaves view)

  unobstructed -- free of endoscopic artifacts? Obstructions: simethicone bubbles/foam, bile (yellow-green), residual stool/debris, active water jet/submersion, blood, instruments (forceps/snare). NOT obstructions: thin clear mucus, a light water film, a few sparse bubbles.

      0 = fluid / foam / stool / tool covers most of the field

      1 = frequent heavy obstruction

      2 = minor moisture / a few bubbles / mild bile; tissue still clearly visible

      3 = clean, unobstructed

  image\_quality -- sharpness and exposure? NOTE: a DARK deep-lumen recess is EXPECTED and is the signal -- do NOT penalize darkness from a far lumen; penalize only true black / washed-white frames and motion blur.

      0 = mostly severely blurred or badly over/under-exposed

      1 = frequent motion blur or exposure problems

      2 = generally sharp and well-exposed, occasional blur

      3 = consistently sharp and well-exposed

- Timestamps in M:SS from the start of the video.
\par\nobreak\smallskip
\hrule height 0.4pt
\vspace{1pt}
\hrule height 0.4pt
\endgroup
\medskip

%% file: figures/appendix_qc_examples.tex
\begin{figure}[!ht]
\centering
\includegraphics[width=\linewidth]{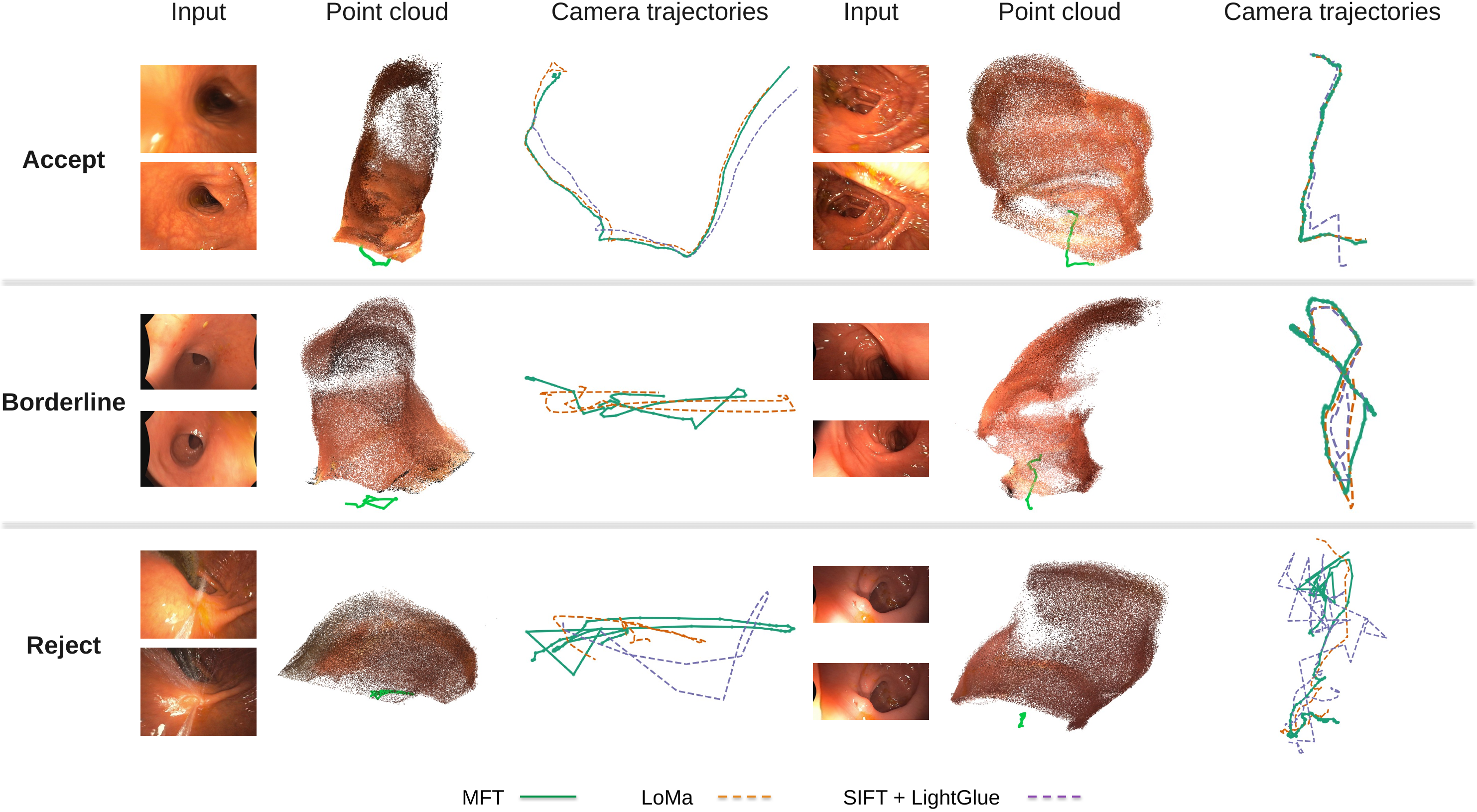}
\caption{\textbf{Quality-control examples.}}
\label{fig:qc-examples}
\end{figure}

%% file: sections/appendix_implementation.tex
\section{Implementation Details}
\label{app:implementation}

\subsection{Optimization and Trainable Modules}
\label{app:optimization}
We initialize the geometry backbone from pretrained VGGT-$\Omega$~\citep{wang2026vggtomega}
and the motion decoder, assignment head, and residual-twist decoder from
SM4RT~\citep{lin2026sm4rt}. The common-motion head has a zero-initialized
output layer.
We train both geometric SFT and the full model for 29,000 optimizer steps on
eight B200 GPUs. We use AdamW with weight decay 0.05, betas $(0.9,0.95)$,
500 warmup steps, and cosine learning-rate decay.
The full model uses a $2\times10^{-5}$ motion-branch learning rate and
$10^{-5}$ for the image encoder and attention backbone, decaying to zero; its geometry
heads remain fixed. Geometric SFT updates all parameters, using $10^{-5}$ for
the model and $5\times10^{-7}$ for the patch encoder, with minimum rates of
$10^{-7}$ and $5\times10^{-9}$, respectively.
Training uses landscape aspect ratios at a nominal 512-pixel resolution and
a budget of 24 images per GPU per step.

The full model uses $K=10$ residual bases, including the zero-residual anchor.
The tracking loss weights are $\lambda_{\mathrm{tr}}=2000$,
$\lambda_c=1$, and $\lambda_q=0.5$.

Dataset sampling proportions follow Table~\ref{tab:datasets}.
Geometric SFT uses 3,360 training clips and 76 validation clips; the full model
uses 3,349 training clips and 75 validation clips. The full-model training pool excludes nine Sano
clips without motion labels and two StereoMIS clips too short for view sampling.
Training samples contain 2--24 views, and validation uses 24 views.
Geometric SFT uses geometry/covisibility sampling. For Sano and StereoMIS,
\scopename{} samples ordered views with a target temporal span of 64 source frames;
other sources use geometry sampling.

\textbf{Ablation settings.}
Data-composition variants use the geometric SFT recipe, each initialized from
pretrained VGGT-$\Omega$ and trained with the same budget on eight B200 GPUs. Sampling weights are renormalized over the retained datasets.
Motion-design variants are trained on StereoMIS for 8,000 steps on four B200 GPUs.
The SoM baseline uses sparsemax without an anchor. The anchor variant uses
softmax with a zero-motion anchor and static/moving assignment supervision. Full CRM further adds common--residual composition and
its corresponding supervision. In this full formulation, the anchor represents
zero residual motion: points assigned to it follow the common transform rather
than remaining stationary.

\subsection{Architecture and Supervision}
\label{app:motion-targets}

\textbf{Features and coordinates.}
The motion branch reads backbone layers 4, 11, 17, and 23 (zero-indexed) and
projects their 2,048-dimensional features to 1,536 dimensions through linear
adapters. A motion transformer predicts per-frame twists and temporally shared
reference-frame assignments. Subtracting the first-frame common and residual
twists from their respective sequences makes both initial transforms identity.
Trajectories use first-camera coordinates. Transforms act on annotated reference
points during training and predicted reference points at inference.
Training coordinates are normalized by the mean reference-track distance, with
common translations scaled consistently.

\textbf{Constructing 3D tracking supervision.}
For StereoMIS, MFT correspondences are lifted into 3D using stereo depth and
camera poses, excluding occlusions and instruments. Sano trajectories fuse rendered
optical flow and depth across views using simulator cameras.

We robustly fit a common SE(3) transform to reference-to-target correspondences,
using a $3\epsilon$ inlier threshold and rigid refitting;
$\epsilon$ is a source-specific annotation noise floor.

Let $X_i(t)$ denote an annotated trajectory and $T_c^\star(t)$ the shared
common-motion target. The residual amplitude and anchor target are
\begin{align}
 r_i&=\operatorname{percentile}_{90,t>0}
       \left\|X_i(t)-T_c^\star(t)X_i(0)\right\|_2,\\
 r_i'&=\max(r_i-\epsilon,0),\qquad
 q_{0,i}^\star=1-\frac{r_i'}{r_i'+2\epsilon}.
 \label{eq:anchor-target}
\end{align}
We compute the percentile over valid observations with fitted common-motion
targets. Residuals are measured in millimeters before training normalization, with
$\epsilon=0.03$\,mm for Sano and $1.32$\,mm for StereoMIS.
For trajectory weighting, $r_i'>\epsilon$ defines the larger-residual group;
the remaining points form the smaller-residual group.

\textbf{Geometry supervision.}
We follow the MapAnything training framework for geometry
supervision~\citep{keetha2026mapanything}. Points, depth, and camera translations
are normalized by the mean valid point distance. The losses are
\begin{equation}
\begin{aligned}
\mathcal L_{\mathrm{pts}}
 &=\sum_t\Bigl[
 \bigl\langle c_t\,\rho\!\left(\phi(P_t)-\phi(P_t^\star)\right)
       -0.2\log c_t\bigr\rangle_{M_t}\\
 &\hspace{3.8em}+\bigl\langle\rho\!\left(
 \phi(P_t^{\mathrm{cam}})-\phi(P_t^{\mathrm{cam},\star})\right)
 \bigr\rangle_{M_t}^{\mathrm{trim}}\Bigr],\\
\mathcal L_{\mathrm{cam}}
 &=\sum_t\Bigl[\rho(\tau_t-\tau_t^\star)
 +\min_{s\in\{-1,1\}}\rho(u_t-su_t^\star)
 +\bigl\langle\rho(\mathbf r_t-\mathbf r_t^\star)\bigr\rangle_{\Omega}\Bigr],\\
\mathcal L_{\mathrm{depth}}
 &=\sum_t\bigl\langle\rho\!\left(
 \log(1+D_t)-\log(1+D_t^\star)\right)\bigr\rangle_{M_t}^{\mathrm{trim}}.
\end{aligned}
\label{eq:geometry-supervision}
\end{equation}
Here $P_t$ and $P_t^{\mathrm{cam}}$ are pointmaps in the first-camera and local
camera frames; $\tau_t$, $u_t$, and $\mathbf r_t$ denote translation, unit
quaternion, and camera ray directions; $D_t$ is camera-axis depth and $c_t$ is
predicted confidence. Stars denote supervision targets. Minimizing over quaternion signs handles
the two equivalent representations of each rotation.

The point transform and robust penalty are
\begin{equation}
 \phi(p)=\frac{\log(1+\|p\|_2)}{\|p\|_2}p,\qquad
 \rho(e)=\frac{|\alpha-2|}{\alpha}
 \left[\left(1+\frac{\|e\|_2^2}{\sigma^2|\alpha-2|}\right)^{\alpha/2}-1\right],
\end{equation}
with $\phi(0)=0$, $\alpha=0.5$, and $\sigma=0.05$.
Angle brackets denote pixel means over the annotation validity mask $M_t$ or
image grid $\Omega$. The trimmed mean excludes the largest 5\% of valid losses
within each real image; synthetic images use the full valid set.

\textbf{Motion supervision.}
For predicted and target displacements
$\widehat d_i(t)=\widehat X_i(t)-\widehat X_i(0)$ and
$d_i(t)=X_i(t)-X_i(0)$, define
\[
 \ell(\mathcal A)=\frac{1}{3|\mathcal A|}
   \sum_{(i,t)\in\mathcal A}\|\widehat d_i(t)-d_i(t)\|_2^2.
\]
The set $\mathcal A$ contains observations valid at both the reference and target
frames. With the larger- and smaller-residual observation sets
$\mathcal A_f$ and $\mathcal A_b$,
\begin{equation}
 \mathcal L_{\mathrm{tr}}=\ell(\mathcal A_f)+2.5\ell(\mathcal A_b).
\end{equation}

The common-motion loss compares reference points under predicted and fitted
common transforms:
\begin{equation}
 \mathcal L_c=\frac{1}{|\mathcal C|\,|\mathcal P|}
 \sum_{t\in\mathcal C}\sum_{i\in\mathcal P}
 \left\|T_c(t)X_i(0)-T_c^\star(t)X_i(0)\right\|_2.
\end{equation}
Here $\mathcal C$ indexes frames with valid common-motion targets, and
$\mathcal P$ indexes sampled reference points.
For valid reference seeds $\mathcal Q$, let $a_i=q_0(x_i)$ and
$b_i=q_{0,i}^\star$. The assignment loss is
\begin{equation}
 \mathcal L_q=\frac{1}{|\mathcal Q|}\sum_{i\in\mathcal Q}
 \left[b_i\log\frac{b_i}{a_i}+(1-b_i)\log\frac{1-b_i}{1-a_i}\right],
\end{equation}
where $\mathcal Q$ includes only seeds with valid observations and finite targets.

Motion losses are averaged over the full batch, with unlabeled samples
contributing zero. The joint objective is scaled by $2/T$ for $T$ input views.

%% file: sections/appendix_proof.tex
\section{Proof of the CRM Residual Bound}
\label{app:crm-proof}
We bound displacement relative to common motion under the CRM representation.

\textbf{SE(3) exponential map.}
The hat operator maps a twist $\xi=(v,\omega)\in\mathbb R^6$ to its
Lie-algebra matrix:
\begin{equation}
 \xi^{\wedge}=
 \begin{bmatrix}
  [\omega]_\times & v\\
  \mathbf 0^\top & 0
 \end{bmatrix}\in\mathfrak{se}(3),\qquad
 \operatorname{Exp}(\xi):=\exp(\xi^{\wedge})\in\mathrm{SE}(3).
 \label{eq:se3-exp}
\end{equation}
Here $[\omega]_\times p=\omega\times p$, and $\exp$ is the matrix exponential.

\textbf{Claim.}
Fix a reference point $p=X(x,0)$ and time $t$. Let $T_c(t)\in\mathrm{SE}(3)$,
$q_k(x)\geq0$, $\sum_{k=0}^{K-1}q_k(x)=1$, and $\Delta\xi_0(t)=0$.
For finite residual bases $\Delta\xi_k(t)=(v_k(t),\omega_k(t))$, define
\[
 B(p,t)=\max_{1\leq k<K}
 \left(\|v_k(t)\|_2+\|\omega_k(t)\|_2\|p\|_2\right).
\]
With $\Delta\xi(x,t)=\sum_{k=1}^{K-1}q_k(x)\Delta\xi_k(t)$ and
$\widehat X(x,t)=T_c(t)\operatorname{Exp}(\Delta\xi(x,t))p$,
\[
 \|\widehat X(x,t)-T_c(t)p\|_2\leq(1-q_0(x))B(p,t).
\]

\textbf{Proof.}
We use the exact SE(3) exponential map defined above.
For a twist $\zeta=(v,\omega)$, let $A=[\omega]_\times$. Its action is
\[
 \operatorname{Exp}(\zeta)p=e^A p+\int_0^1 e^{sA}v\,ds.
\]
Since $e^A-I=\int_0^1e^{sA}A\,ds$ and $e^{sA}$ is a rotation,
\[
\begin{aligned}
 \|\operatorname{Exp}(\zeta)p-p\|_2
 &=\left\|\int_0^1e^{sA}(Ap+v)\,ds\right\|_2\\
 &\leq\|Ap\|_2+\|v\|_2
 \leq\|\omega\|_2\|p\|_2+\|v\|_2.
\end{aligned}
\]
The common transform $T_c(t)$ preserves distances. Applying the triangle
inequality and $\sum_{k>0}q_k(x)=1-q_0(x)$ gives
\begin{equation}
\begin{aligned}
 \|\widehat X(x,t)-T_c(t)p\|_2
 &=\|\operatorname{Exp}(\Delta\xi(x,t))p-p\|_2\\
 &\leq\sum_{k>0}q_k(x)
   \bigl(\|v_k(t)\|_2+\|\omega_k(t)\|_2\|p\|_2\bigr)\\
 &\leq(1-q_0(x))B(p,t).
\end{aligned}
\label{eq:residual-bound}
\end{equation}
For fixed residual bases, displacement relative to common motion vanishes
as $q_0(x)\to1$: the point follows $T_c(t)p$. A zero mixed twist in SoM
instead suppresses all point motion.

%% file: sections/appendix_evaluation.tex
\section{Evaluation Protocols}
\label{app:metrics}
\label{app:evaluation}

\subsection{Compared Methods}
\label{app:contracts}
\input{tables/baseline_contracts}

Table~\ref{tab:contracts} summarizes the compared methods. D4RT$^*$ denotes the
open-source reproduction. A single resolution value specifies the target long
edge before cropping; paired values specify width $\times$ height.
Test-time optimization refers to geometric fitting: CUT3R fits focal length;
Endo3R fits focal length and estimates camera poses with PnP; D4RT uses
closed-form camera fitting from predicted points.

\subsection{In-Domain Protocol}
\label{app:pose-metrics}

We follow \citet{wang2026vggtomega,guo2025endo3r} in evaluating depth with
absolute relative error (AbsRel) and $\delta_{1.25}$. After sequence-level median
scale alignment, AbsRel measures mean absolute depth error divided by reference
depth; $\delta_{1.25}$ reports the percentage of predictions within a factor of
$1.25$ of the reference.

For camera poses, we follow \citet{wang2025vggt,wang2026vggtomega} in using
pose AUC at $5^\circ$ and $30^\circ$. Pose AUC is the normalized area under the
recall curve for the larger of the rotation and translation-direction angular
errors of each image pair. Following \citet{keetha2026mapanything},
we report absolute trajectory error (ATE) as the mean camera-position error after
scale normalization and rigid alignment. Relative translation error (RTE) and
relative rotation error (RRE) assess motion between adjacent frames, as in
\citet{guo2025endo3r}. We report mean translation-vector error in millimeters and mean rotation-angle
error in degrees, respectively.

For 3D tracking, we follow \citet{lin2026sm4rt} in reporting endpoint error (EPE)
and average percent of points within delta (APD). After sequence-level scale
alignment, EPE measures the Euclidean distance between predicted and reference
3D positions. APD averages the percentages of predictions within $1$, $2$, $5$,
and $10$\,mm of their references. We average scores over sequences within each dataset and compute the ID geometry
summary over the seven datasets.

\subsection{Out-of-Domain Protocol}
\label{app:ood-protocol}

On Hamlyn, we follow \citet{guo2025endo3r} in evaluating depth with AbsRel and
$\delta_{1.25}$. Instrument masks are generated with SAM~2.1 Hiera-Large
\citep{ravi2024sam2}, using box and point prompts, bidirectional video propagation,
and visual review with prompt refinement. The same source masks are mapped to each
method's evaluation grid; sequence-level median scale alignment and depth metrics
use the intersection of valid reference depth and non-instrument pixels.
Because camera-pose ground truth is unavailable, we evaluate
camera stability on clips with a stationary camera. ATE and absolute rotation
error (ARE) measure translation and rotation variation relative to the first
predicted pose.

On \scopephantom{}, camera estimates are evaluated against electromagnetic
tracker measurements. ATE measures camera-position root mean square error (RMSE)
after Sim(3) alignment, and RTE measures one-second relative translation RMSE. ARE$^*$ measures mean viewing-direction error after global
direction alignment. The asterisk indicates that the 5-DoF tracker does not
record roll.

For long RealSynCol sequences, we follow \citet{deng2025vggtlong}
in reporting ATE as camera-position RMSE after Sim(3) alignment.
Adjacent-frame RTE and RRE are translation-vector RMSE in millimeters and
rotation-angle RMSE in degrees, respectively.

\phantomsection
\textbf{Real-colonoscopy user study.}
\label{app:user-study}
All four methods in Table~\ref{tab:user-study} receive the same
128 input frames from each of ten clips.
Case and method order are randomized per participant. Rankings run from 1 (best)
to 4 (worst), without ties. Participants assess camera motion consistent with the
video, coherent colon-wall surfaces, and temporally stable depth with clear fold
and lumen boundaries; path length, surface coverage, and absolute depth colors
are not criteria. Good/bad examples and an unscored phantom practice view precede
the study; examples remain accessible during ranking.

Instrument masks use the same SAM~2.1-assisted procedure as Hamlyn and are shared
across methods: RGB masks are overlaid, depth regions are grayed out, and
corresponding 3D points are removed before sampling.
Figure~\ref{fig:user-study-interface} shows the synchronized comparison interface.
The six clinical and biomedical experts completed 180 four-method ranking tasks in total
(720 individual ranks), without missing responses or exclusions. Method identities were hidden.

Table~\ref{tab:user-study} reports mean ranks with 95\% percentile bootstrap
confidence intervals from 50,000 resamples of participants and clips separately,
preserving method pairing. Mean per-clip Kendall's $W$ is 0.780, 0.728, and
0.698 for camera trajectories, point clouds, and depth, respectively.
Statistical analyses were specified after collection. Exploratory two-sided
exact sign tests on participant-level mean rank differences do not reach the
0.05 level after Holm correction across the nine method--dimension comparisons.
Given the small study sample, we interpret the results
as sample-specific visual preferences rather than population-level superiority.

\FloatBarrier
\input{figures/appendix_user_study_masked_interface}

%% file: tables/baseline_contracts.tex
\begin{table*}[t]
\caption{\textbf{Compared methods.}}
\label{tab:contracts}
\centering
\scriptsize
\setlength{\tabcolsep}{2pt}
\begin{tabularx}{\linewidth}{@{}>{\raggedright\arraybackslash}p{.25\linewidth}>{\raggedright\arraybackslash}p{.10\linewidth}>{\raggedright\arraybackslash}X>{\raggedright\arraybackslash}p{.14\linewidth}cc@{}}
\toprule
Method & Input & Output & Resolution & \shortstack{Supplied\\intrinsics} & \shortstack{Test-time\\optimization} \\
\midrule
\multicolumn{6}{@{}l}{\textit{General 3D}} \\
VGGT~\citep{wang2025vggt} & RGB video & Depth, cameras, points & 518 & $\times$ & $\times$ \\
VGGT-$\Omega$~\citep{wang2026vggtomega} & RGB video & Depth, cameras, points & 512 & $\times$ & $\times$ \\
MapAnything~\citep{keetha2026mapanything} & RGB video & Depth, cameras, points & 518 & $\times$ & $\times$ \\
CUT3R~\citep{wang2025cut3r} & RGB video & Depth, cameras, points & 512 & $\times$ & $\checkmark$ \\
\midrule
\multicolumn{6}{@{}l}{\textit{General 4D}} \\
SpatialTrackerV2~\citep{xiao2025spatialtrackerv2} & RGB video & Depth, cameras, 3D tracks & 518 & $\times$ & $\times$ \\
4RC~\citep{luo2026fourrc} & RGB video & Depth, cameras, 3D tracks & 512 & $\times$ & $\times$ \\
Any4D~\citep{karhade2025any4d} & RGB video & Depth, cameras, 3D tracks & 518 & $\times$ & $\times$ \\
SM4RT~\citep{lin2026sm4rt} & RGB video & Depth, cameras, 3D tracks & 518 & $\times$ & $\times$ \\
D4RT$^{*}$~\citep{zhang2026d4rt} & RGB video & Depth, cameras, 3D tracks & $256\times256$ & $\times$ & $\checkmark$ \\
\midrule
\multicolumn{6}{@{}l}{\textit{Endoscopy}} \\
AF-SfMLearner~\citep{shao2022afsfm} & RGB video & Depth, camera poses & $320\times256$ & $\checkmark$ & $\times$ \\
EndoDAC~\citep{cui2024endodac} & RGB video & Depth, camera poses & $320\times256$ & $\times$ & $\times$ \\
Endo-FASt3R~\citep{zeinoddin2025endofast3r} & RGB video & Depth, camera poses & $320\times256$ (depth)\newline 512 (pose) & $\checkmark$ & $\times$ \\
Endo3R~\citep{guo2025endo3r} & RGB video & Depth, cameras, points & 320 & $\times$ & $\checkmark$ \\
\bottomrule
\end{tabularx}
\end{table*}

%% file: figures/appendix_user_study_masked_interface.tex
\begin{figure}[htbp]
\centering\small
\includegraphics[width=\linewidth]{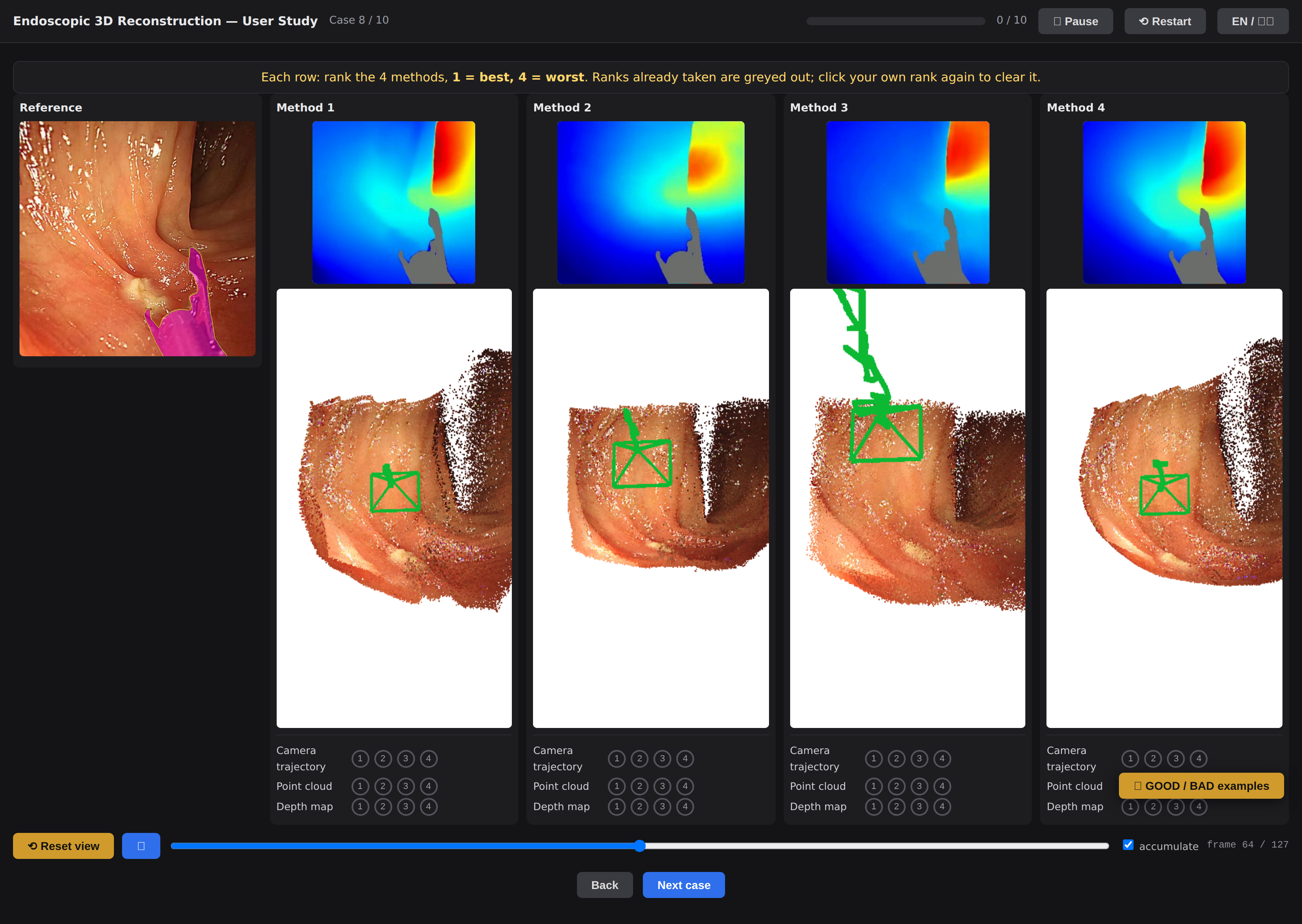}
\caption{\textbf{User-study interface.} Reference video, masked depth, interactive point clouds, and ranking controls for four anonymous methods. No rankings are selected.}
\label{fig:user-study-interface}
\end{figure}
\FloatBarrier

%% file: sections/appendix_results.tex
\section{Additional Results}
\label{app:results}

\subsection{Additional Quantitative Results}
\label{app:geometry-results}

\textbf{Geometry.}
We report per-dataset camera pose and depth results on SCARED,
SimCol3D, C3VD, EndoMapper-Sim, and EndoMapper-Real.
Sano and StereoMIS results are reported in Table~\ref{tab:tracking}.

\input{tables/geometry_per_domain}

\phantomsection
\textbf{3D tracking.}
\label{app:tracking}
Table~\ref{tab:tracking-details} reports APD at each distance threshold for Sano
and StereoMIS, supplementing the mean APD in Table~\ref{tab:tracking}.

\input{tables/tracking_details}

\phantomsection
\textbf{Motion ablation.}
\label{app:motion-results}
Table~\ref{tab:motion-ablation-details} expands Table~\ref{tab:ablation}(b)
with threshold-wise APD.

\input{tables/motion_ablation_details}
\FloatBarrier

\clearpage
\subsection{Additional Qualitative Results}
\label{app:visualization}

The following figures compare reconstructed geometry and depth across datasets.
Geometry panels use the same tissue-centered crop for all methods within a case;
distant camera frusta may lie outside the displayed region.

\input{figures/appendix_multicase}

\FloatBarrier

\clearpage
\input{figures/long_all_scenes}
\FloatBarrier

%% file: tables/geometry_per_domain.tex
\begin{table}[H]
\centering\footnotesize
\setlength{\tabcolsep}{3pt}
\caption{\textbf{Camera pose and depth estimation on SCARED.}}
\begin{tabular*}{\linewidth}{@{\extracolsep{\fill}}lccccc@{}}
\toprule
& \multicolumn{3}{c}{Pose} & \multicolumn{2}{c}{Depth} \\
\cmidrule(lr){2-4}\cmidrule(lr){5-6}
Method & AUC@$5^\circ\uparrow$ & AUC@$30^\circ\uparrow$ & ATE$\downarrow$ & AbsRel$\downarrow$ & $\delta_{1.25}\uparrow$ \\
\midrule
VGGT & $10.31$ & $64.78$ & $0.047$ & $0.047$ & $99.15$ \\
VGGT-$\Omega$ & $17.87$ & $76.51$ & $0.030$ & $0.038$ & $99.36$ \\
MapAnything & $6.10$ & $56.83$ & $0.050$ & $0.071$ & $95.00$ \\
CUT3R & $0.35$ & $11.80$ & $0.174$ & $0.135$ & $81.43$ \\
SpatialTrackerV2 & $8.65$ & $68.50$ & $0.049$ & $0.038$ & $99.35$ \\
4RC & $16.61$ & $77.25$ & $0.032$ & $0.035$ & $\underline{99.41}$ \\
Any4D & $0.49$ & $20.43$ & $0.084$ & $0.074$ & $97.45$ \\
SM4RT & $17.43$ & $76.49$ & $0.060$ & $\underline{0.034}$ & $\mathbf{99.47}$ \\
Endo3R & $0.17$ & $16.07$ & $0.161$ & $0.100$ & $89.38$ \\
AF-SfMLearner & $0.28$ & $16.18$ & $0.125$ & $0.119$ & $87.59$ \\
EndoDAC & $0.28$ & $17.22$ & $0.121$ & $0.121$ & $85.79$ \\
Endo-FASt3R & $14.82$ & $76.52$ & $0.047$ & $0.117$ & $88.16$ \\
\midrule
\textbf{\boldmath\scopesft{}} & $\mathbf{29.92}$ & $\mathbf{83.12}$ & $\underline{0.017}$ & $\underline{0.034}$ & $99.15$ \\
\textbf{\ourmodel{}} & $\underline{27.89}$ & $\underline{82.88}$ & $\mathbf{0.016}$ & $\mathbf{0.033}$ & $99.21$ \\
\bottomrule
\end{tabular*}
\end{table}

\begin{table}[H]
\centering\footnotesize
\setlength{\tabcolsep}{3pt}
\caption{\textbf{Camera pose and depth estimation on SimCol3D.}}
\begin{tabular*}{\linewidth}{@{\extracolsep{\fill}}lccccc@{}}
\toprule
& \multicolumn{3}{c}{Pose} & \multicolumn{2}{c}{Depth} \\
\cmidrule(lr){2-4}\cmidrule(lr){5-6}
Method & AUC@$5^\circ\uparrow$ & AUC@$30^\circ\uparrow$ & ATE$\downarrow$ & AbsRel$\downarrow$ & $\delta_{1.25}\uparrow$ \\
\midrule
VGGT & $0.44$ & $10.82$ & $0.288$ & $0.181$ & $68.03$ \\
VGGT-$\Omega$ & $6.69$ & $52.96$ & $0.072$ & $0.093$ & $92.77$ \\
MapAnything & $0.22$ & $7.39$ & $0.258$ & $0.172$ & $71.11$ \\
CUT3R & $0.09$ & $11.28$ & $0.223$ & $0.459$ & $32.57$ \\
SpatialTrackerV2 & $0.79$ & $35.24$ & $0.105$ & $0.155$ & $75.69$ \\
4RC & $1.19$ & $28.02$ & $0.163$ & $0.153$ & $76.02$ \\
Any4D & $0.06$ & $3.84$ & $0.227$ & $0.297$ & $43.55$ \\
SM4RT & $0.95$ & $22.27$ & $0.197$ & $0.165$ & $71.88$ \\
Endo3R & $0.08$ & $7.03$ & $0.202$ & $0.265$ & $54.12$ \\
AF-SfMLearner & $0.67$ & $14.14$ & $0.275$ & $0.306$ & $44.98$ \\
EndoDAC & $0.57$ & $14.55$ & $0.257$ & $0.242$ & $55.23$ \\
Endo-FASt3R & $0.79$ & $21.50$ & $0.130$ & $0.255$ & $53.13$ \\
\midrule
\textbf{\boldmath\scopesft{}} & $\underline{54.81}$ & $\underline{91.20}$ & $\mathbf{0.009}$ & $\underline{0.025}$ & $\underline{99.90}$ \\
\textbf{\ourmodel{}} & $\mathbf{59.70}$ & $\mathbf{91.98}$ & $\underline{0.010}$ & $\mathbf{0.023}$ & $\mathbf{99.91}$ \\
\bottomrule
\end{tabular*}
\end{table}

\begin{table}[H]
\centering\footnotesize
\setlength{\tabcolsep}{3pt}
\caption{\textbf{Camera pose and depth estimation on C3VD.}}
\begin{tabular*}{\linewidth}{@{\extracolsep{\fill}}lccccc@{}}
\toprule
& \multicolumn{3}{c}{Pose} & \multicolumn{2}{c}{Depth} \\
\cmidrule(lr){2-4}\cmidrule(lr){5-6}
Method & AUC@$5^\circ\uparrow$ & AUC@$30^\circ\uparrow$ & ATE$\downarrow$ & AbsRel$\downarrow$ & $\delta_{1.25}\uparrow$ \\
\midrule
VGGT & $7.37$ & $50.19$ & $0.097$ & $0.196$ & $65.00$ \\
VGGT-$\Omega$ & $18.54$ & $70.34$ & $0.042$ & $0.062$ & $97.05$ \\
MapAnything & $9.37$ & $60.34$ & $0.056$ & $0.144$ & $78.64$ \\
CUT3R & $0.51$ & $22.06$ & $0.138$ & $0.406$ & $38.45$ \\
SpatialTrackerV2 & $25.32$ & $77.26$ & $\underline{0.031}$ & $0.076$ & $94.10$ \\
4RC & $16.40$ & $66.05$ & $0.050$ & $0.117$ & $86.05$ \\
Any4D & $0.92$ & $27.01$ & $0.178$ & $0.270$ & $51.13$ \\
SM4RT & $7.68$ & $57.45$ & $0.103$ & $0.152$ & $78.62$ \\
Endo3R & $0.57$ & $19.60$ & $0.229$ & $0.153$ & $76.45$ \\
AF-SfMLearner & $8.06$ & $55.69$ & $0.130$ & $0.257$ & $51.23$ \\
EndoDAC & $7.71$ & $45.12$ & $0.138$ & $0.217$ & $61.30$ \\
Endo-FASt3R & $11.11$ & $58.56$ & $0.069$ & $0.246$ & $54.69$ \\
\midrule
\textbf{\boldmath\scopesft{}} & $\underline{66.31}$ & $\underline{90.66}$ & $\mathbf{0.009}$ & $\underline{0.031}$ & $\underline{99.35}$ \\
\textbf{\ourmodel{}} & $\mathbf{67.35}$ & $\mathbf{90.82}$ & $\mathbf{0.009}$ & $\mathbf{0.028}$ & $\mathbf{99.40}$ \\
\bottomrule
\end{tabular*}
\end{table}

\begin{table}[H]
\centering\footnotesize
\setlength{\tabcolsep}{3pt}
\caption{\textbf{Camera pose and depth estimation on EndoMapper-Sim.}}
\begin{tabular*}{\linewidth}{@{\extracolsep{\fill}}lccccc@{}}
\toprule
& \multicolumn{3}{c}{Pose} & \multicolumn{2}{c}{Depth} \\
\cmidrule(lr){2-4}\cmidrule(lr){5-6}
Method & AUC@$5^\circ\uparrow$ & AUC@$30^\circ\uparrow$ & ATE$\downarrow$ & AbsRel$\downarrow$ & $\delta_{1.25}\uparrow$ \\
\midrule
VGGT & $0.05$ & $19.17$ & $0.228$ & $0.153$ & $79.49$ \\
VGGT-$\Omega$ & $1.86$ & $47.59$ & $0.067$ & $\underline{0.062}$ & $99.51$ \\
MapAnything & $0.04$ & $8.53$ & $0.265$ & $0.088$ & $94.00$ \\
CUT3R & $0.03$ & $21.75$ & $0.120$ & $0.301$ & $49.62$ \\
SpatialTrackerV2 & $0.42$ & $43.41$ & $0.060$ & $0.077$ & $97.55$ \\
4RC & $0.54$ & $36.93$ & $0.080$ & $0.143$ & $77.74$ \\
Any4D & $0.03$ & $3.26$ & $0.301$ & $0.146$ & $76.16$ \\
SM4RT & $0.44$ & $32.78$ & $0.110$ & $0.157$ & $73.17$ \\
Endo3R & $0.01$ & $9.12$ & $0.240$ & $0.151$ & $76.05$ \\
AF-SfMLearner & $0.03$ & $11.03$ & $0.120$ & $0.305$ & $44.80$ \\
EndoDAC & $0.03$ & $11.31$ & $0.108$ & $0.182$ & $68.80$ \\
Endo-FASt3R & $0.04$ & $14.77$ & $0.051$ & $0.241$ & $54.13$ \\
\midrule
\textbf{\boldmath\scopesft{}} & $\mathbf{15.47}$ & $\mathbf{68.49}$ & $\mathbf{0.032}$ & $\mathbf{0.024}$ & $\underline{99.83}$ \\
\textbf{\ourmodel{}} & $\underline{14.06}$ & $\underline{65.30}$ & $\underline{0.037}$ & $\mathbf{0.024}$ & $\mathbf{99.84}$ \\
\bottomrule
\end{tabular*}
\end{table}

\begin{table}[H]
\centering\footnotesize
\setlength{\tabcolsep}{3pt}
\caption{\textbf{Camera pose and depth estimation on EndoMapper-Real.}}
\begin{tabular*}{\linewidth}{@{\extracolsep{\fill}}lccccc@{}}
\toprule
& \multicolumn{3}{c}{Pose} & \multicolumn{2}{c}{Depth} \\
\cmidrule(lr){2-4}\cmidrule(lr){5-6}
Method & AUC@$5^\circ\uparrow$ & AUC@$30^\circ\uparrow$ & ATE$\downarrow$ & AbsRel$\downarrow$ & $\delta_{1.25}\uparrow$ \\
\midrule
VGGT & $1.41$ & $31.43$ & $0.155$ & $0.214$ & $61.56$ \\
VGGT-$\Omega$ & $3.22$ & $49.08$ & $0.089$ & $0.157$ & $76.44$ \\
MapAnything & $1.49$ & $27.23$ & $0.156$ & $0.274$ & $81.11$ \\
CUT3R & $0.20$ & $17.54$ & $0.190$ & $0.408$ & $37.70$ \\
SpatialTrackerV2 & $3.03$ & $54.16$ & $0.097$ & $0.194$ & $68.16$ \\
4RC & $2.16$ & $44.34$ & $0.109$ & $0.219$ & $62.20$ \\
Any4D & $0.16$ & $12.38$ & $0.256$ & $0.388$ & $39.70$ \\
SM4RT & $1.58$ & $43.43$ & $0.106$ & $0.224$ & $61.41$ \\
Endo3R & $0.11$ & $9.92$ & $0.274$ & $0.247$ & $54.90$ \\
AF-SfMLearner & $0.94$ & $34.59$ & $0.204$ & $0.372$ & $37.38$ \\
EndoDAC & $1.19$ & $35.16$ & $0.200$ & $0.306$ & $45.84$ \\
Endo-FASt3R & $1.21$ & $39.23$ & $0.093$ & $0.358$ & $38.99$ \\
\midrule
\textbf{\boldmath\scopesft{}} & $\mathbf{23.59}$ & $\mathbf{77.61}$ & $\mathbf{0.032}$ & $\underline{0.095}$ & $\underline{92.94}$ \\
\textbf{\ourmodel{}} & $\underline{22.80}$ & $\underline{77.35}$ & $\underline{0.033}$ & $\mathbf{0.091}$ & $\mathbf{93.89}$ \\
\bottomrule
\end{tabular*}
\end{table}

%% file: tables/tracking_details.tex
\begin{table}[H]
\caption{\textbf{3D tracking APD at multiple thresholds.}}
\label{tab:tracking-details}
\centering\footnotesize
\setlength{\tabcolsep}{2pt}
\begin{tabular*}{\linewidth}{@{\extracolsep{\fill}}lcccc|cccc@{}}
\toprule
& \multicolumn{4}{c}{\textbf{Sano}} & \multicolumn{4}{c}{\textbf{StereoMIS}} \\
\cmidrule(lr){2-5}\cmidrule(lr){6-9}
Method & APD@1$\uparrow$ & APD@2$\uparrow$ & APD@5$\uparrow$ & APD@10$\uparrow$
& APD@1$\uparrow$ & APD@2$\uparrow$ & APD@5$\uparrow$ & APD@10$\uparrow$ \\
\midrule
D4RT$^{*}$ & 33.99 & 66.11 & 92.98 & 98.69 & 4.29 & 10.56 & 42.99 & 78.63 \\
SpatialTrackerV2 & 77.81 & 88.97 & 96.50 & 99.03 & 6.87 & 18.56 & 50.80 & 79.70 \\
4RC & 86.91 & 94.26 & 98.08 & \underline{99.63} & \underline{12.25} & \underline{26.19} & \underline{56.17} & \underline{82.50} \\
Any4D & 87.29 & 93.65 & 97.35 & 99.34 & 12.02 & 25.08 & 54.46 & 78.25 \\
SM4RT & \underline{89.48} & \underline{95.28} & \underline{98.17} & 99.62 & 12.20 & 25.26 & 56.14 & 82.21 \\
\midrule
\textbf{\ourmodel{}} & \textbf{90.51} & \textbf{95.81} & \textbf{98.42} & \textbf{99.72} & \textbf{19.09} & \textbf{39.53} & \textbf{66.25} & \textbf{87.44} \\
\bottomrule
\end{tabular*}
\end{table}

%% file: tables/motion_ablation_details.tex
\begin{table}[H]
\caption{\textbf{Motion ablation on StereoMIS.}}
\label{tab:motion-ablation-details}
\centering\footnotesize
\setlength{\tabcolsep}{3pt}
\begin{tabular*}{.85\linewidth}{@{\extracolsep{\fill}}lcccc@{}}
\toprule
Variant & APD@1$\uparrow$ & APD@2$\uparrow$ & APD@5$\uparrow$ & APD@10$\uparrow$ \\
\midrule
SoM (baseline) & 11.22 & 28.89 & 63.67 & 82.55 \\
w/ zero-motion anchor & \underline{12.73} & \underline{33.55} & \underline{66.33} & \textbf{87.83} \\
Full CRM & \textbf{19.02} & \textbf{37.30} & \textbf{66.89} & \underline{87.30} \\
\bottomrule
\end{tabular*}
\end{table}

%% file: figures/appendix_multicase.tex
\begin{figure*}[!ht]
\centering
\includegraphics[width=\linewidth,height=0.80\textheight,keepaspectratio]{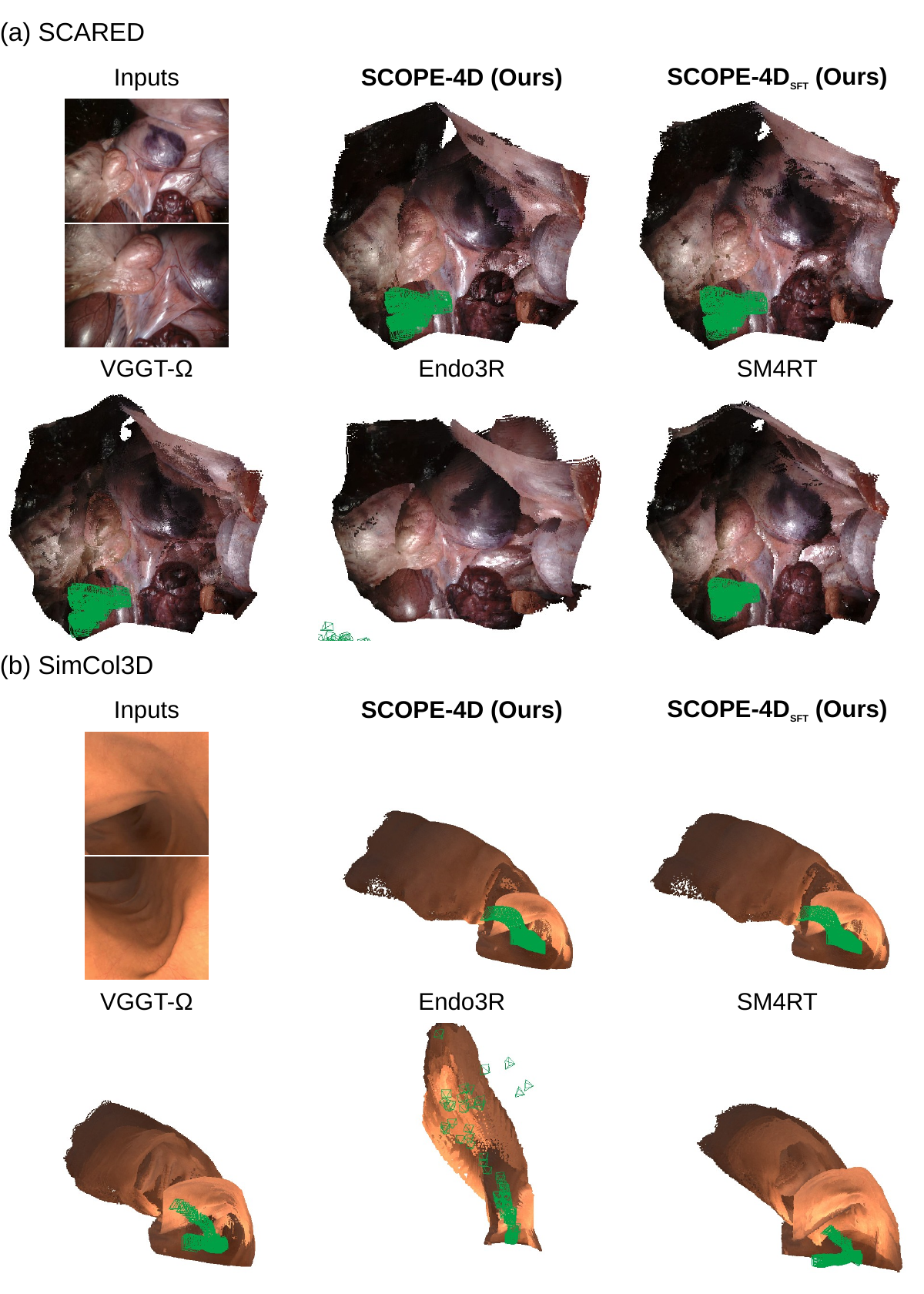}
\caption{\textbf{Additional 3D geometry comparisons.}}
\label{fig:app-geometry-1}
\end{figure*}

\begin{figure*}[p]
\centering
\includegraphics[width=\linewidth]{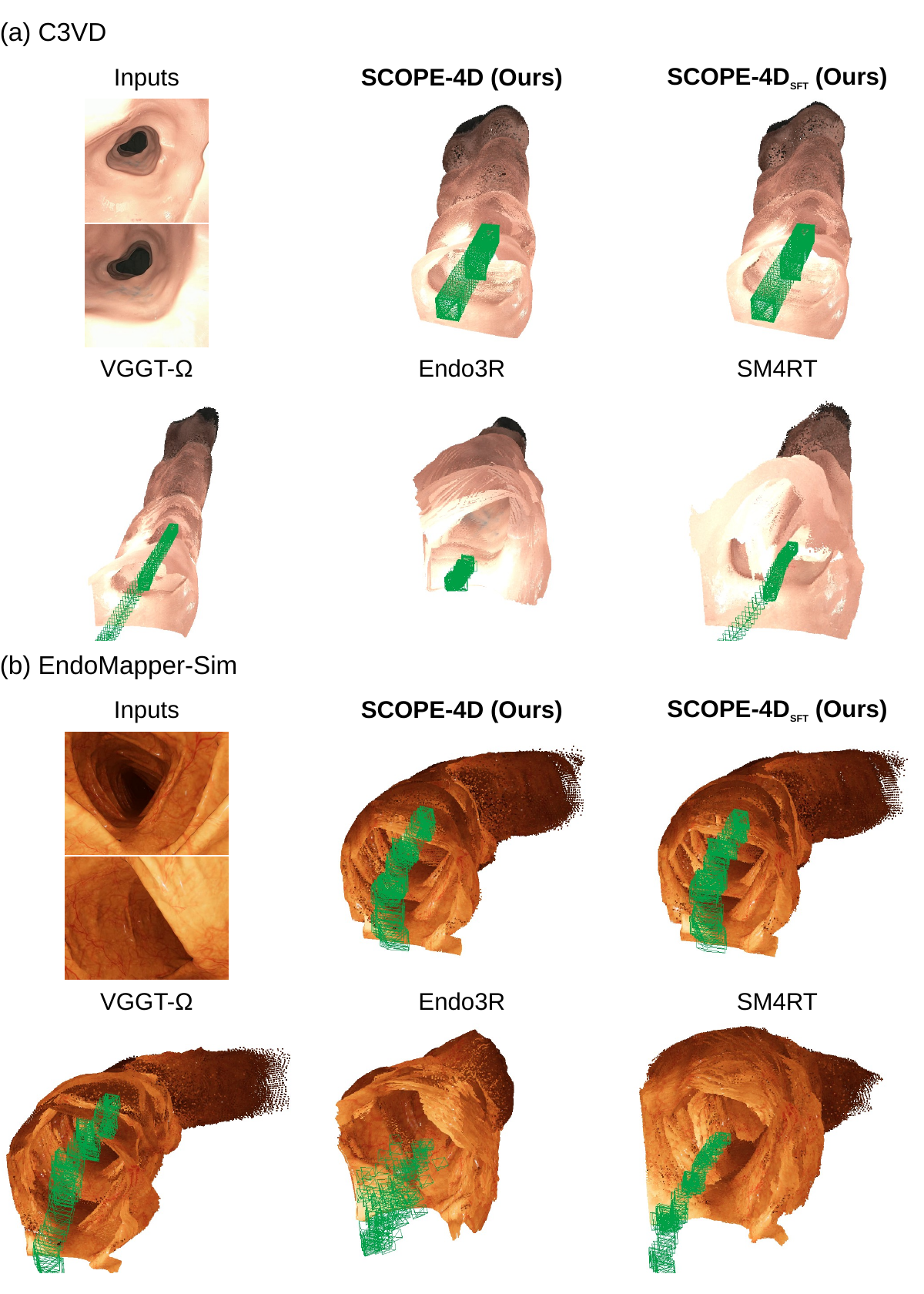}
\caption{\textbf{Additional 3D geometry comparisons.}}
\label{fig:app-geometry-2}
\end{figure*}

\begin{figure*}[p]
\centering
\includegraphics[width=\linewidth]{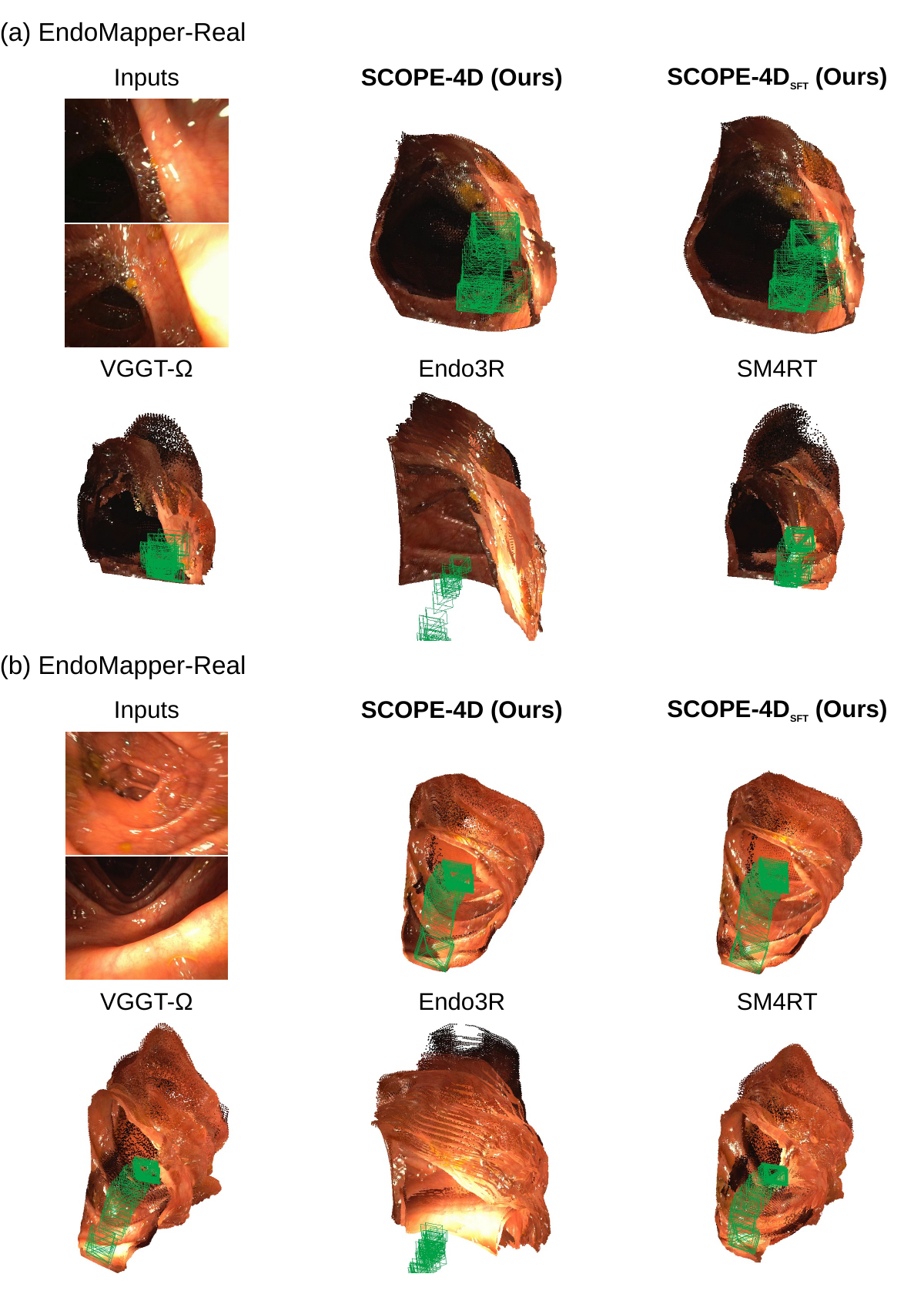}
\caption{\textbf{Additional 3D geometry comparisons.}}
\label{fig:app-geometry-3}
\end{figure*}

\begin{figure*}[p]
\centering
\includegraphics[width=\linewidth]{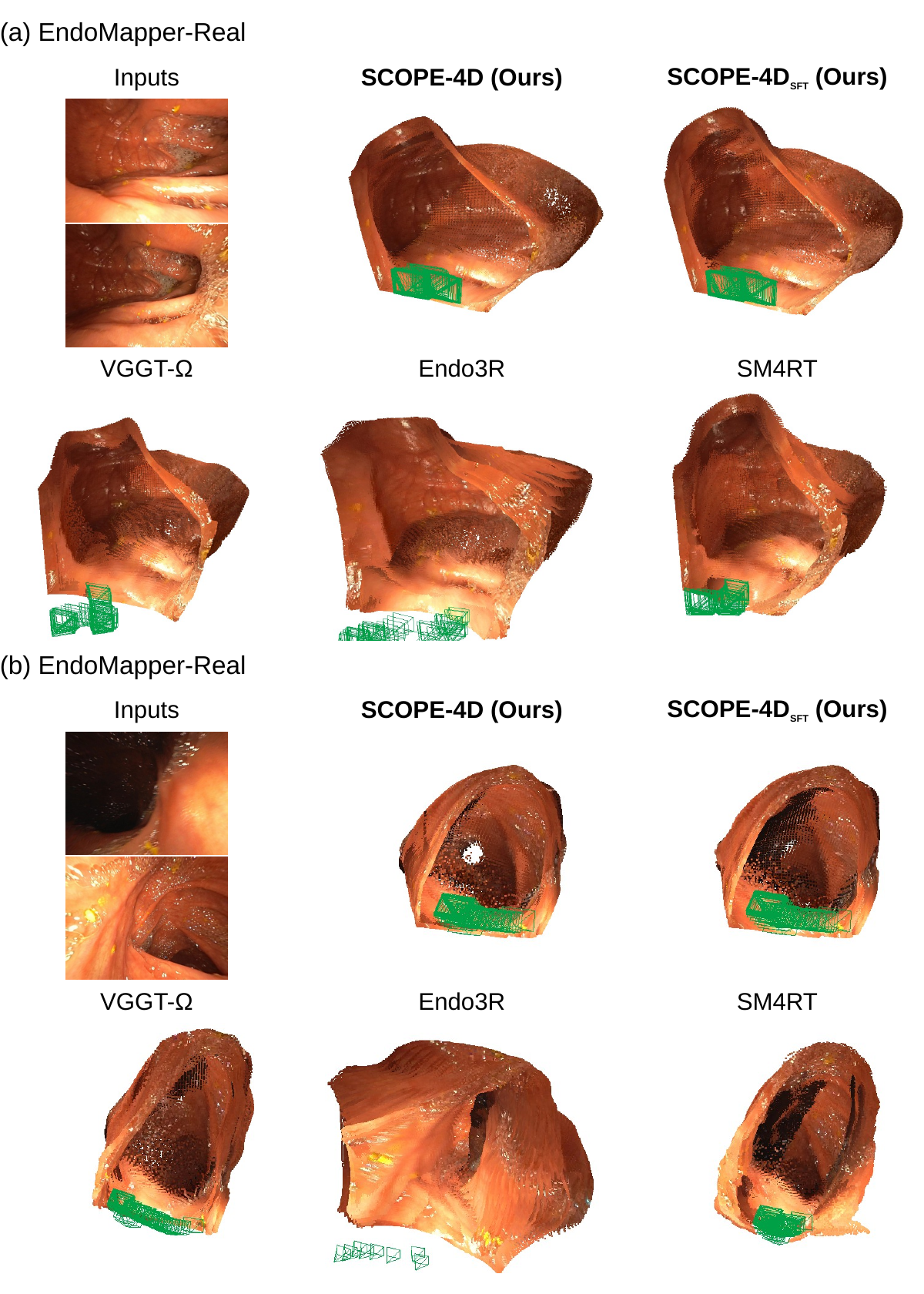}
\caption{\textbf{Additional 3D geometry comparisons.}}
\label{fig:app-geometry-4}
\end{figure*}

\begin{figure*}[p]
\centering
\includegraphics[width=\linewidth]{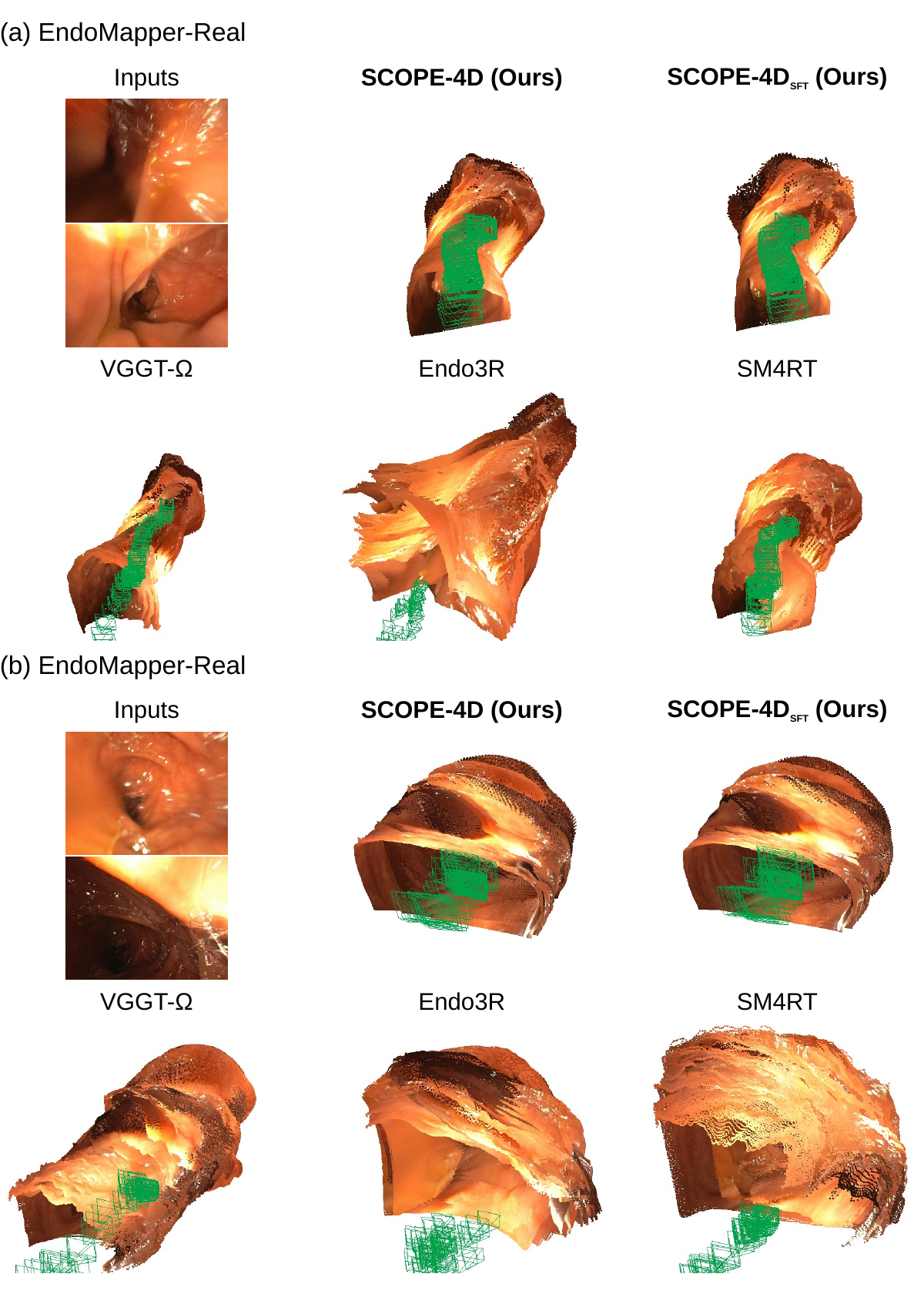}
\caption{\textbf{Additional 3D geometry comparisons.}}
\label{fig:app-geometry-5}
\end{figure*}

\begin{figure*}[p]
\centering
\includegraphics[width=\linewidth]{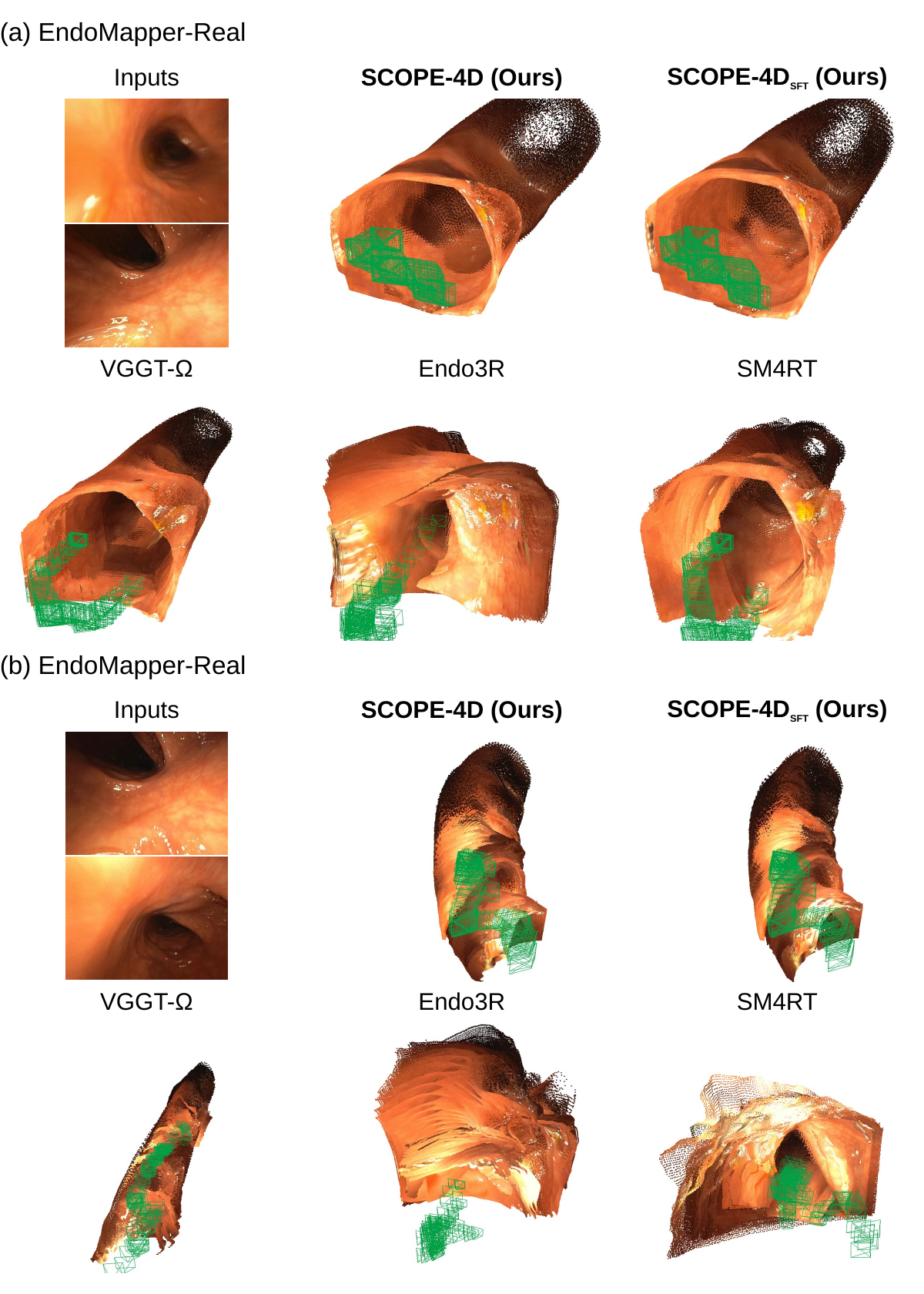}
\caption{\textbf{Additional 3D geometry comparisons.}}
\label{fig:app-geometry-6}
\end{figure*}

\begin{figure*}[p]
\centering
\includegraphics[width=\linewidth]{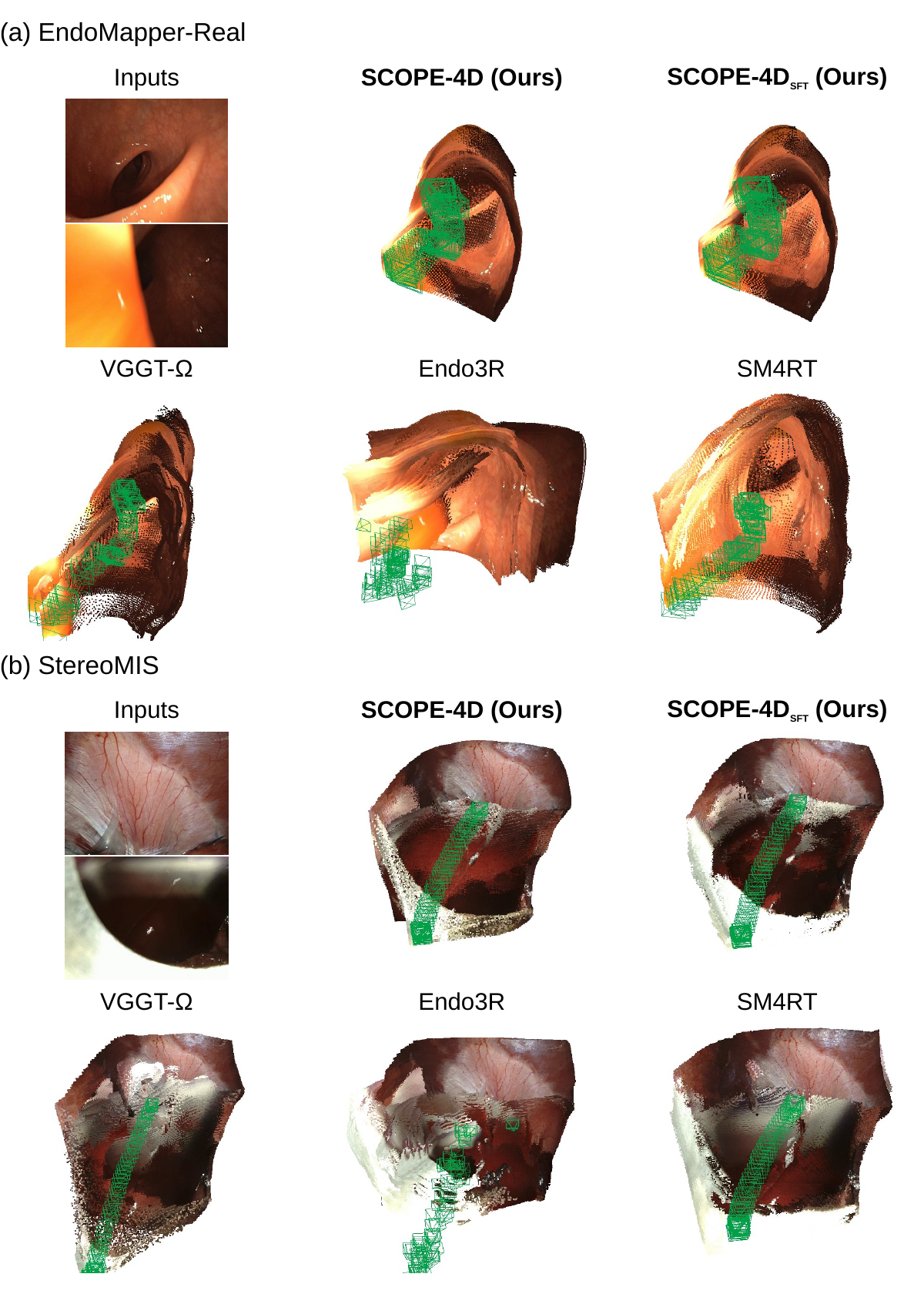}
\caption{\textbf{Additional 3D geometry comparisons.}}
\label{fig:app-geometry-7}
\end{figure*}

\begin{figure*}[p]
\centering
\includegraphics[width=\linewidth]{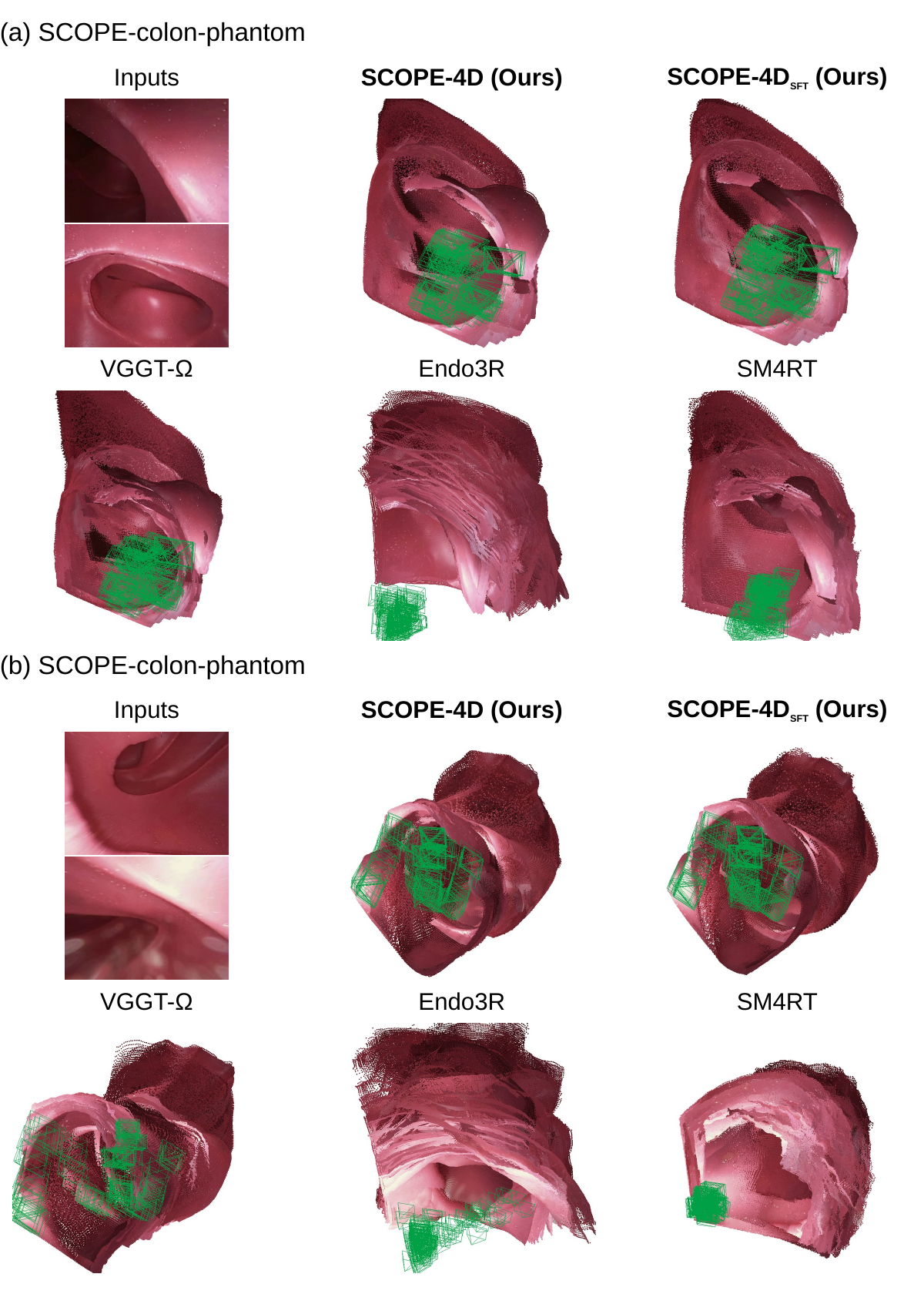}
\caption{\textbf{Additional 3D geometry comparisons.}}
\label{fig:app-geometry-8}
\end{figure*}

\begin{figure*}[p]
\centering
\includegraphics[width=\linewidth]{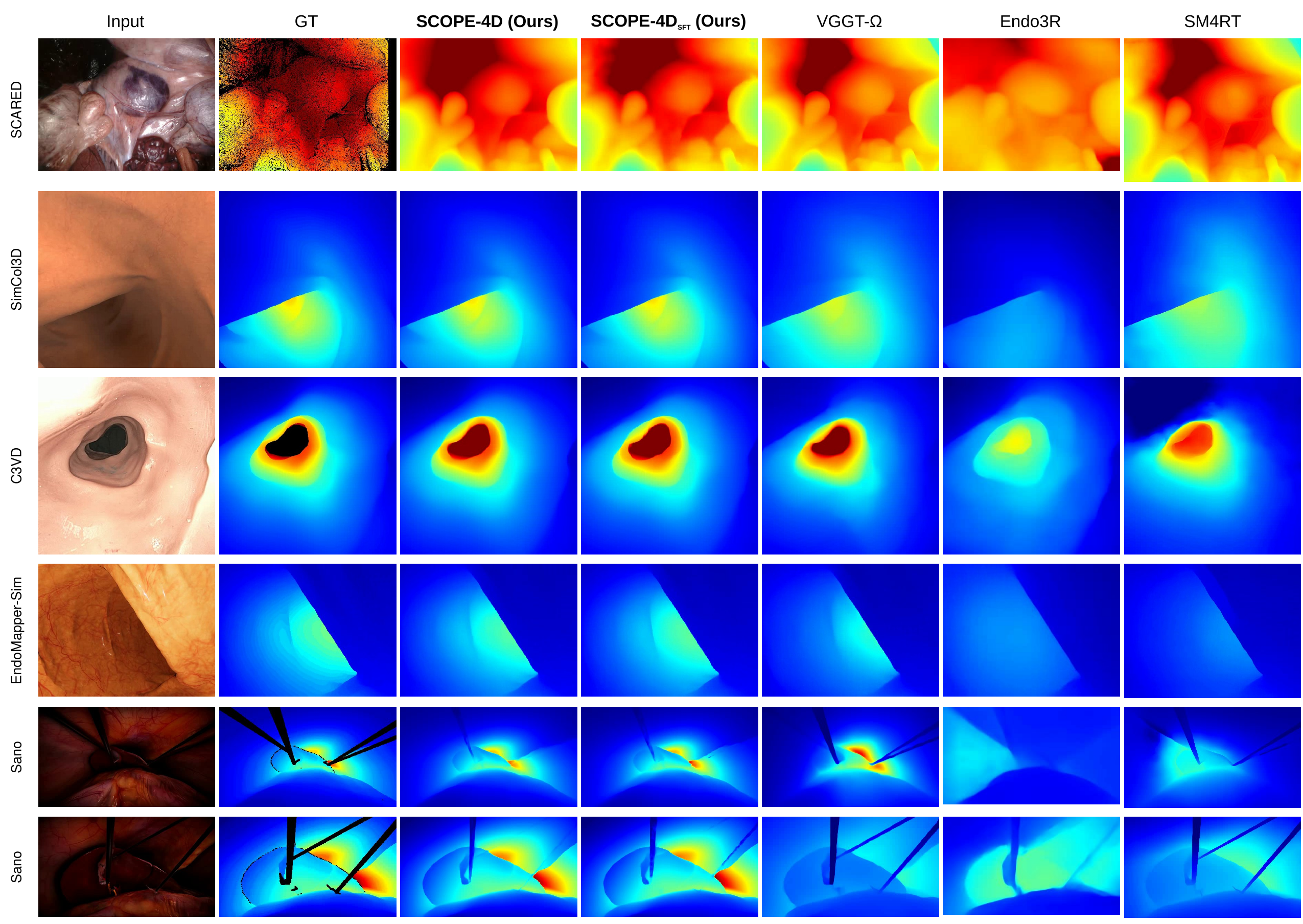}
\caption{\textbf{Additional depth comparisons.}}
\label{fig:app-depth-1}
\end{figure*}

%% file: figures/long_all_scenes.tex
\begin{figure*}[!ht]
\centering
\begingroup
\small
\setlength{\tabcolsep}{3pt}
\begin{tabular}{@{}ccc@{}}
SCOPE-4D (Ours) & \scopesft{} & VGGT-$\Omega$ \\[3pt]
\includegraphics[width=0.315\textwidth]{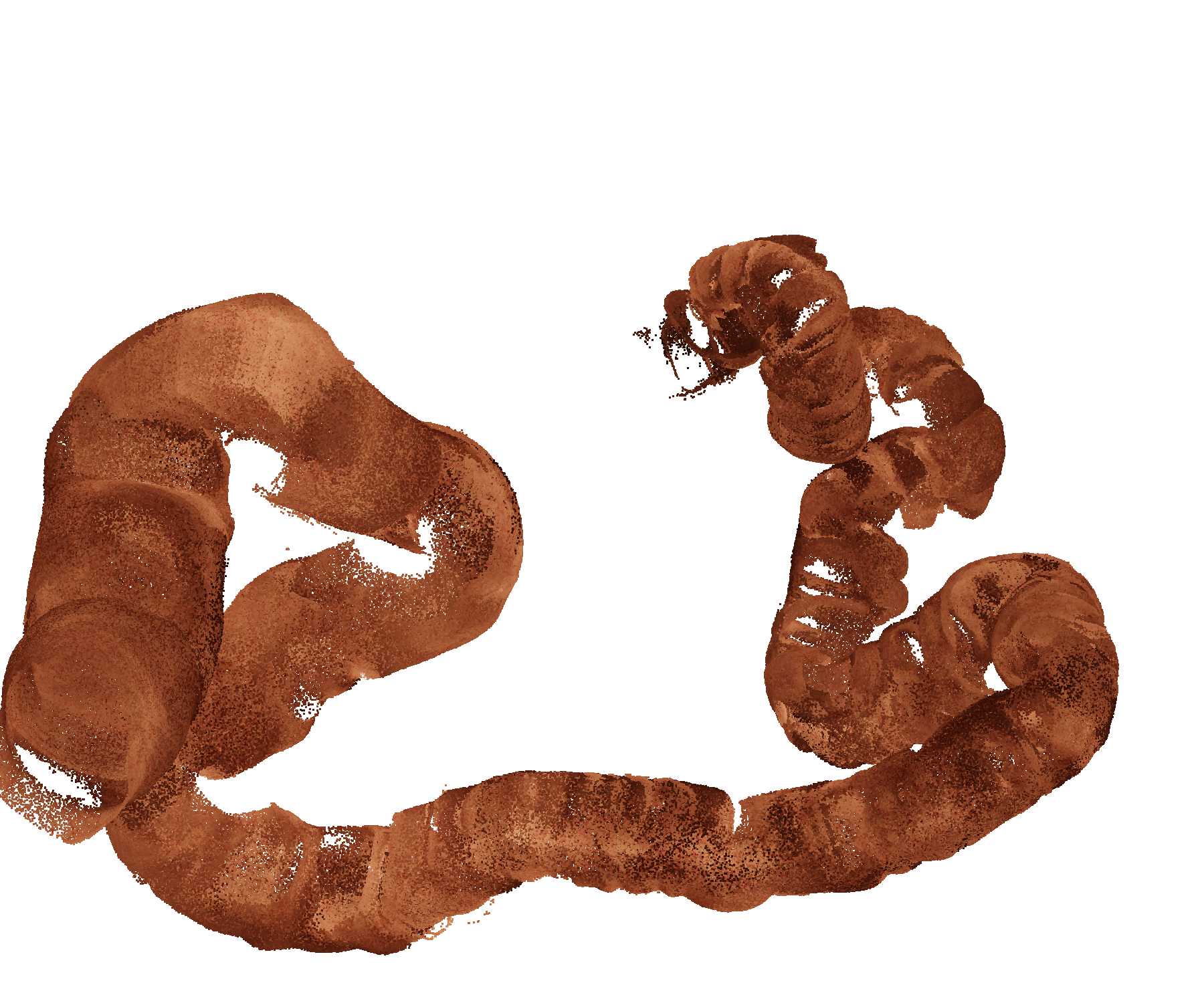} & \includegraphics[width=0.315\textwidth]{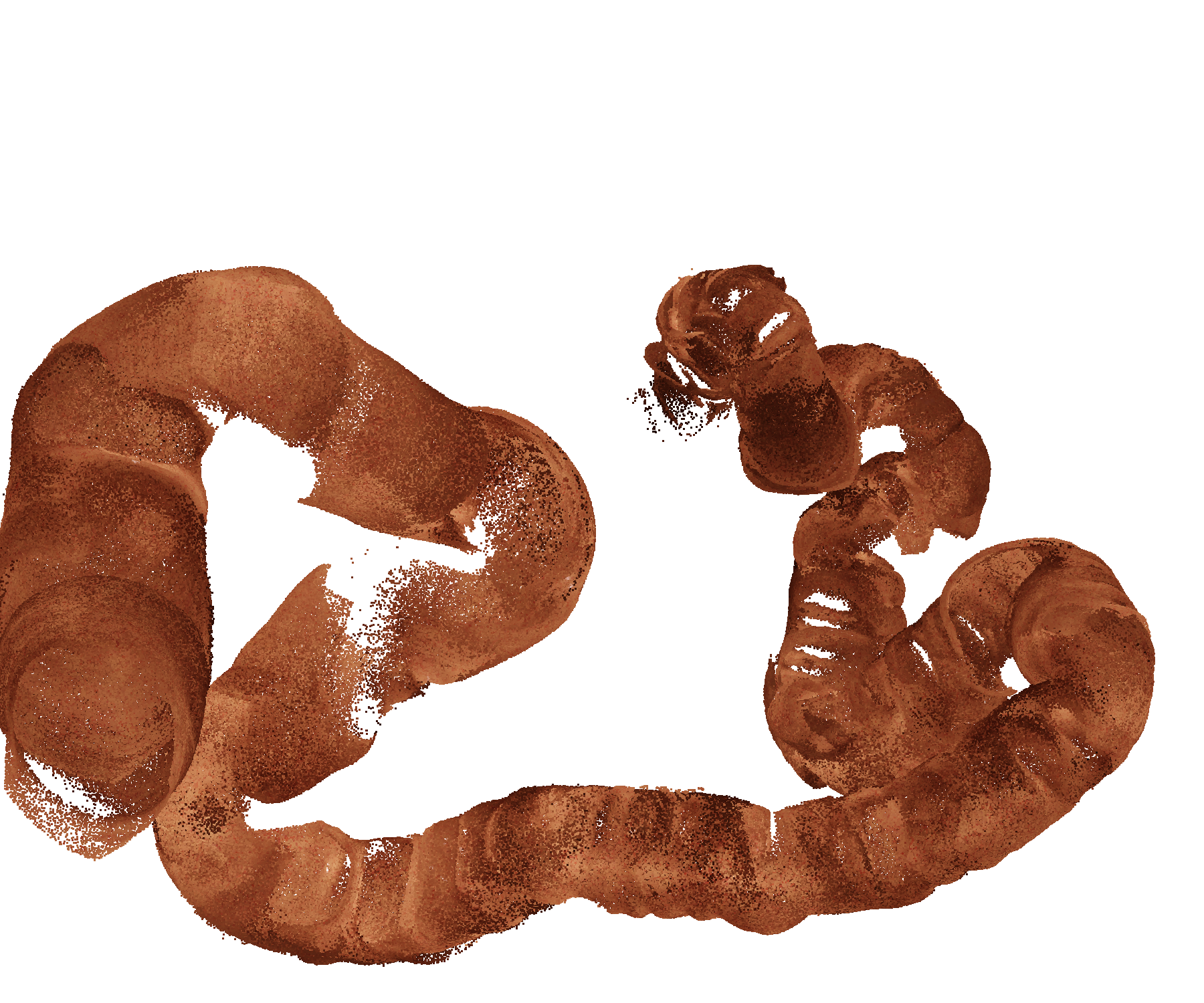} & \includegraphics[width=0.315\textwidth]{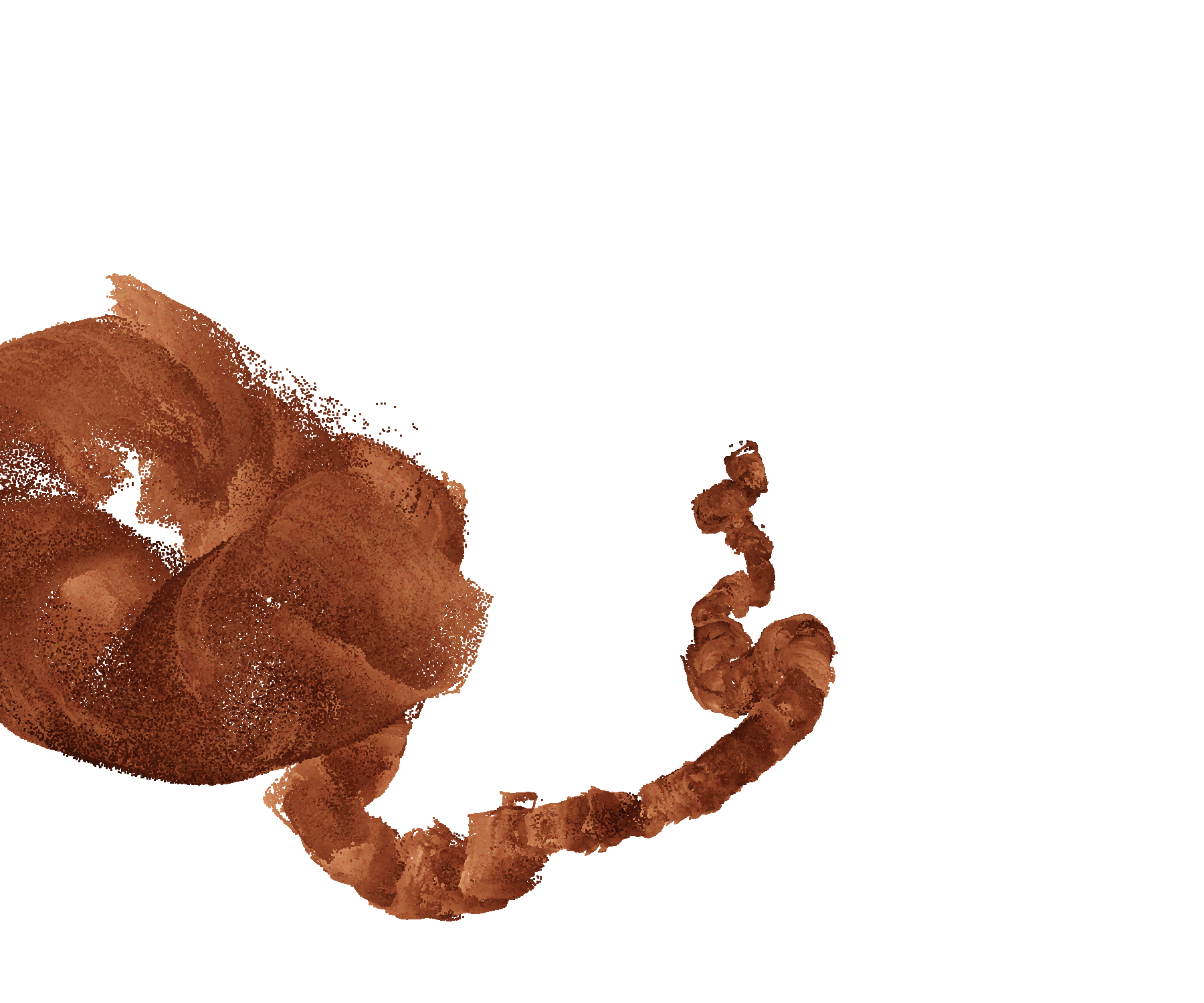} \\[3pt]
\includegraphics[width=0.315\textwidth]{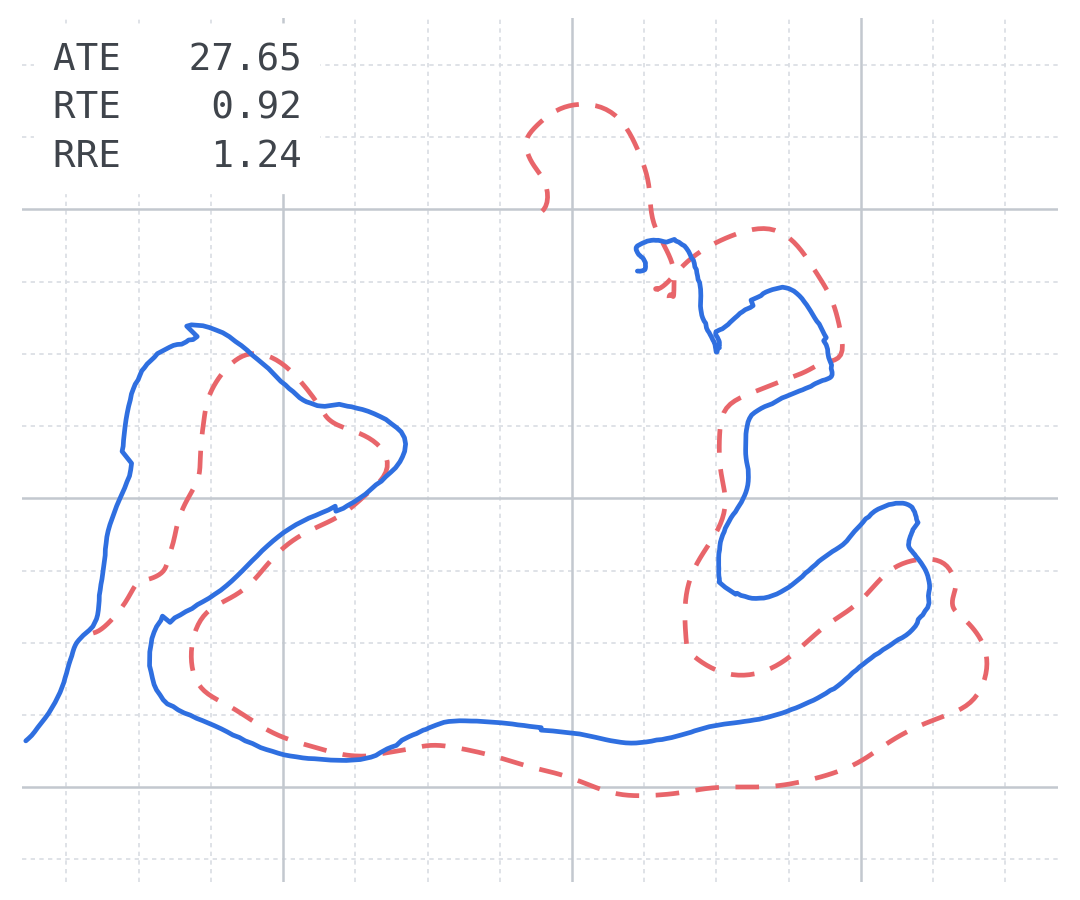} & \includegraphics[width=0.315\textwidth]{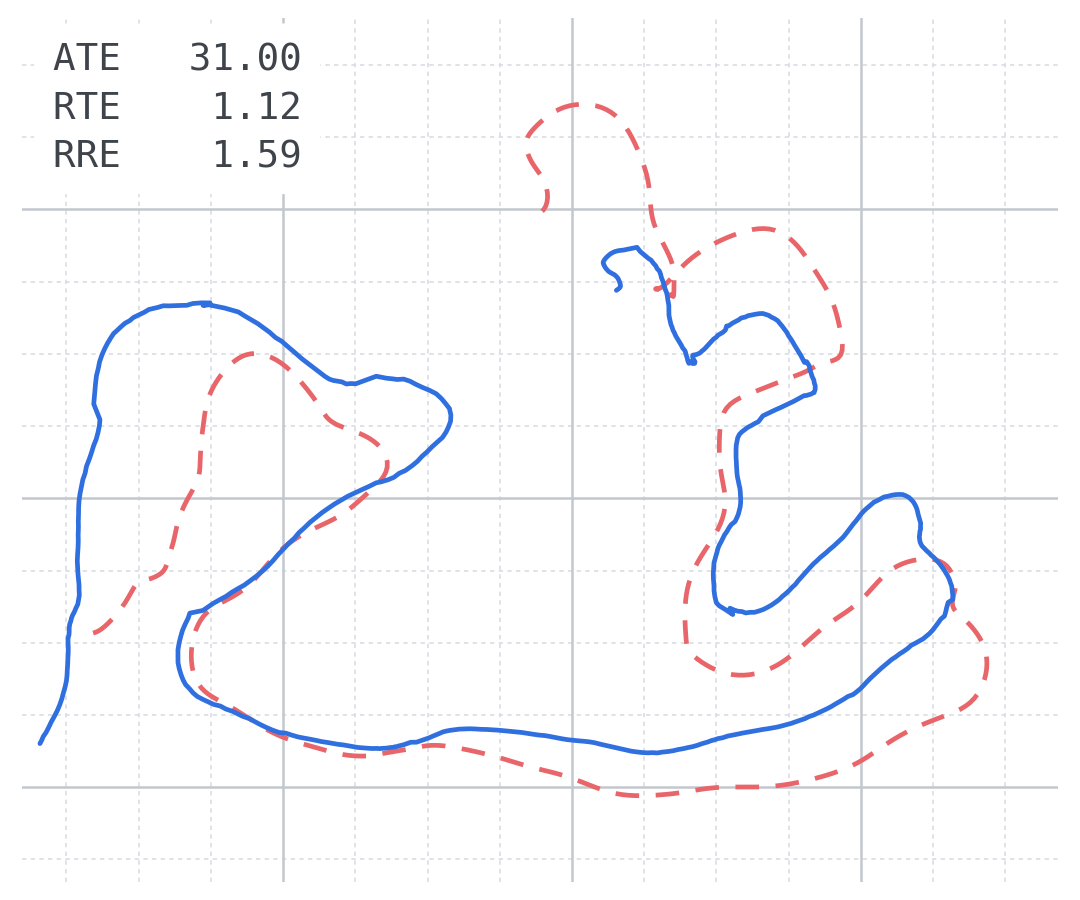} & \includegraphics[width=0.315\textwidth]{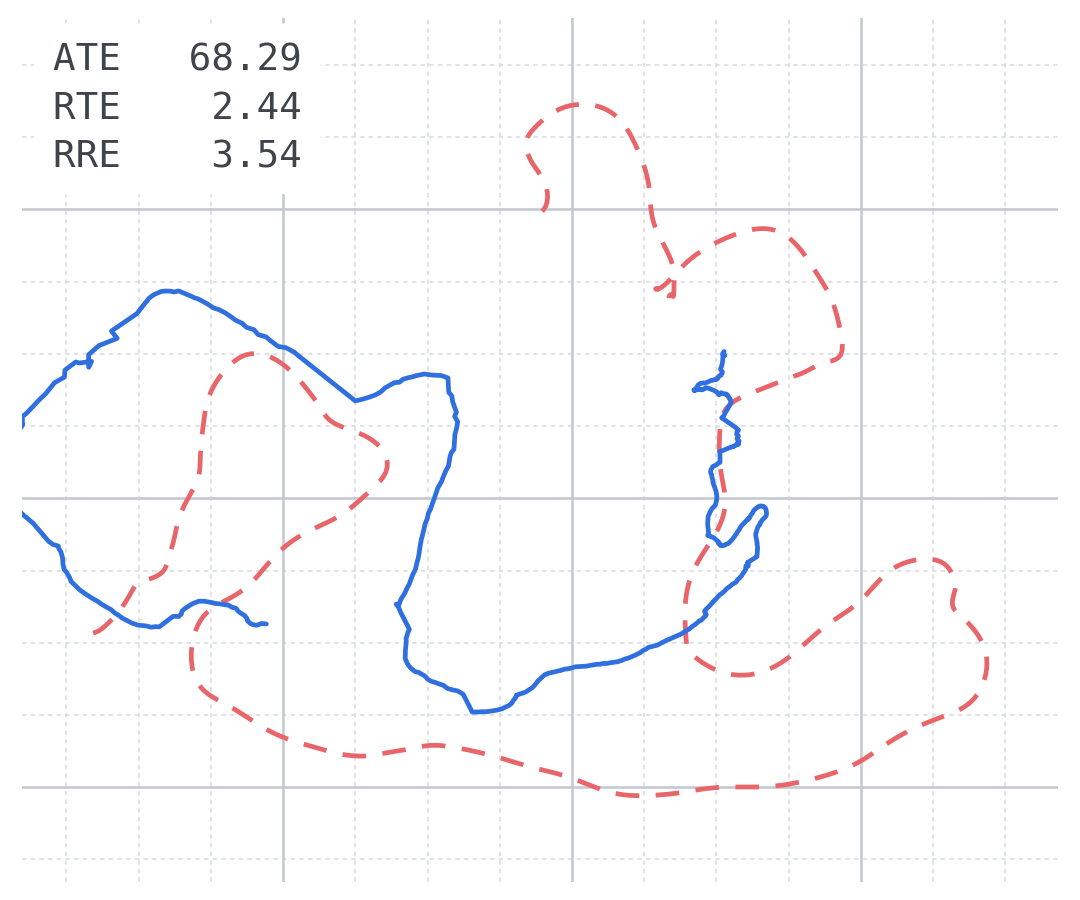} \\[3pt]
\noalign{\vskip 8pt}
VGGT & Endo3R & SM4RT \\[3pt]
\includegraphics[width=0.315\textwidth]{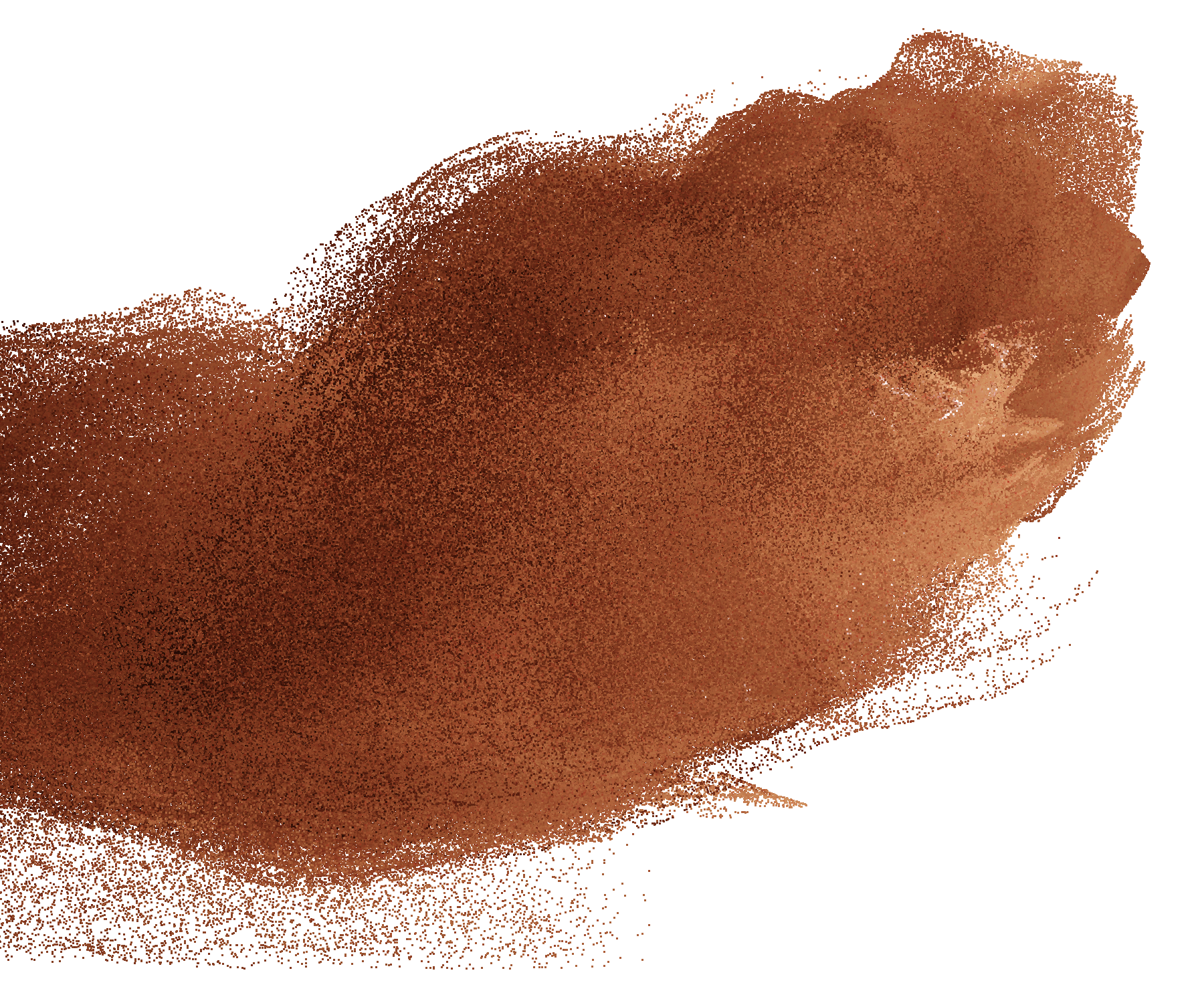} & \includegraphics[width=0.315\textwidth]{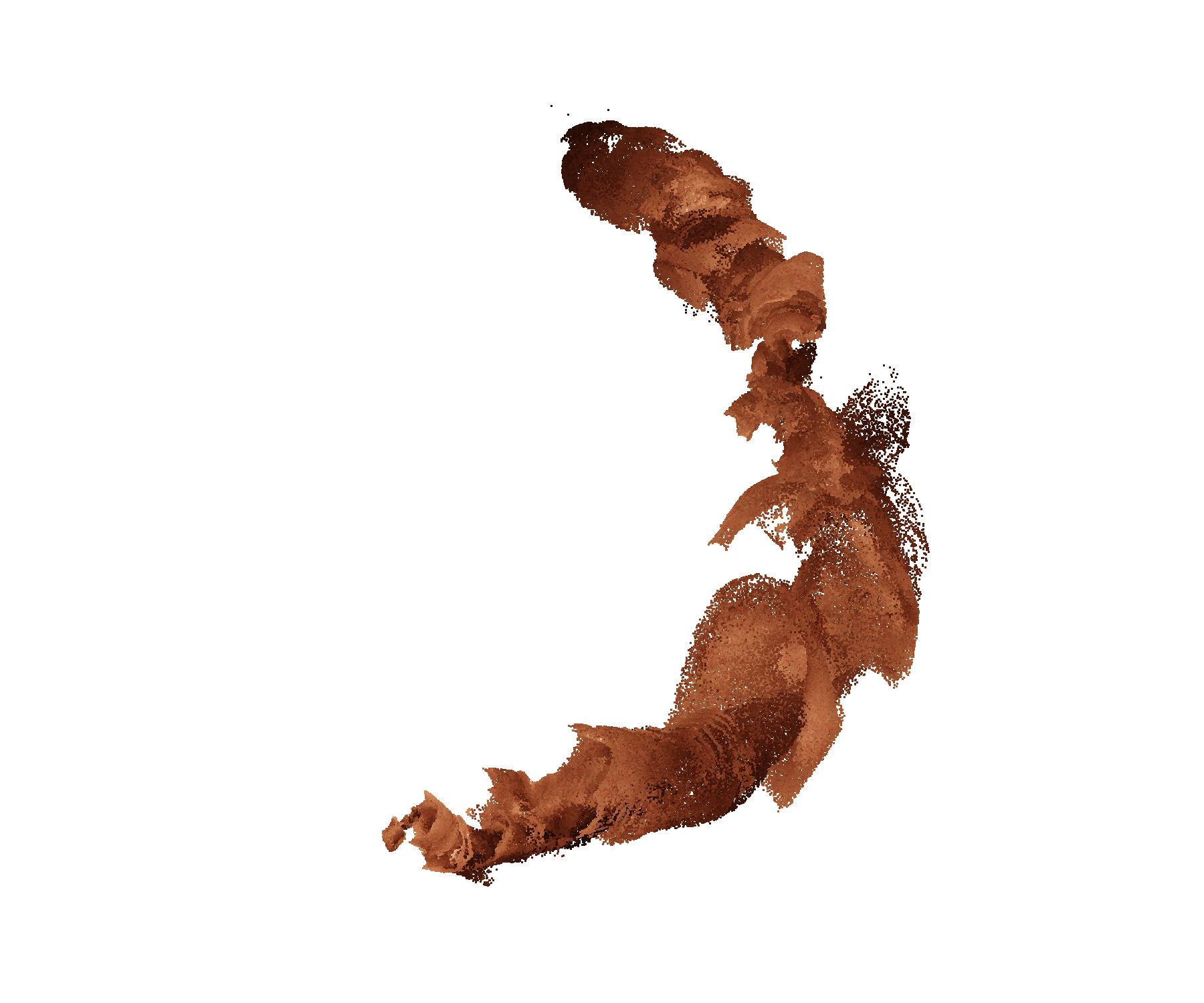} & \includegraphics[width=0.315\textwidth]{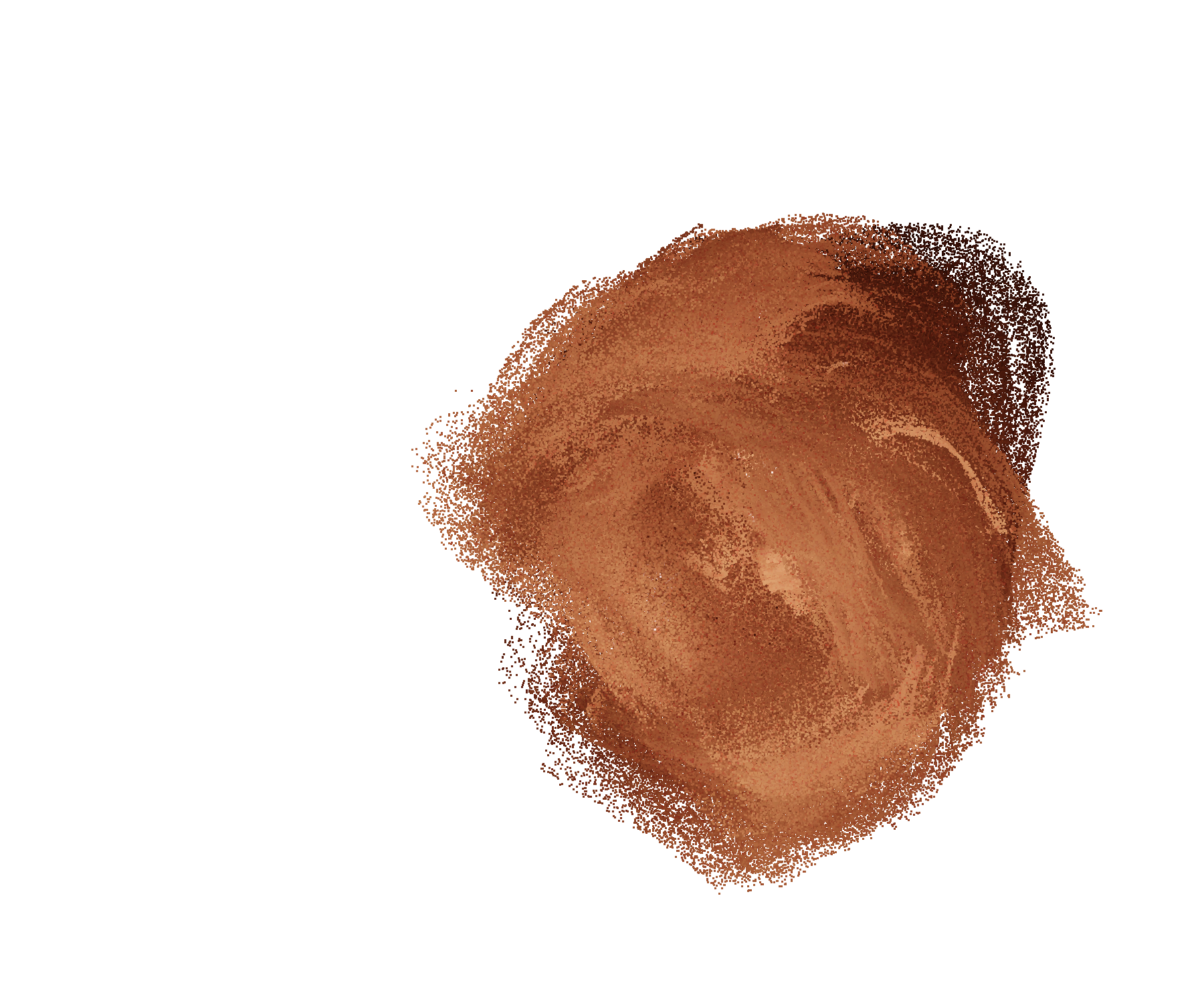} \\[3pt]
\includegraphics[width=0.315\textwidth]{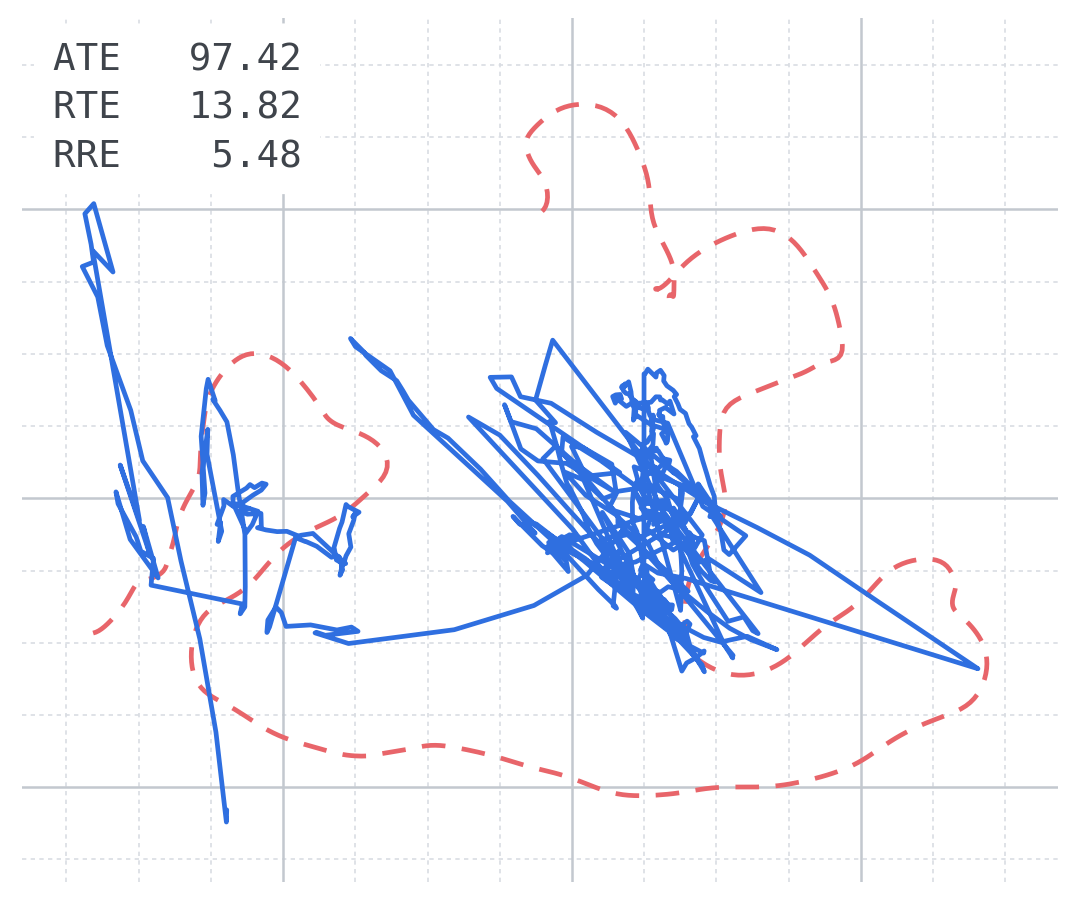} & \includegraphics[width=0.315\textwidth]{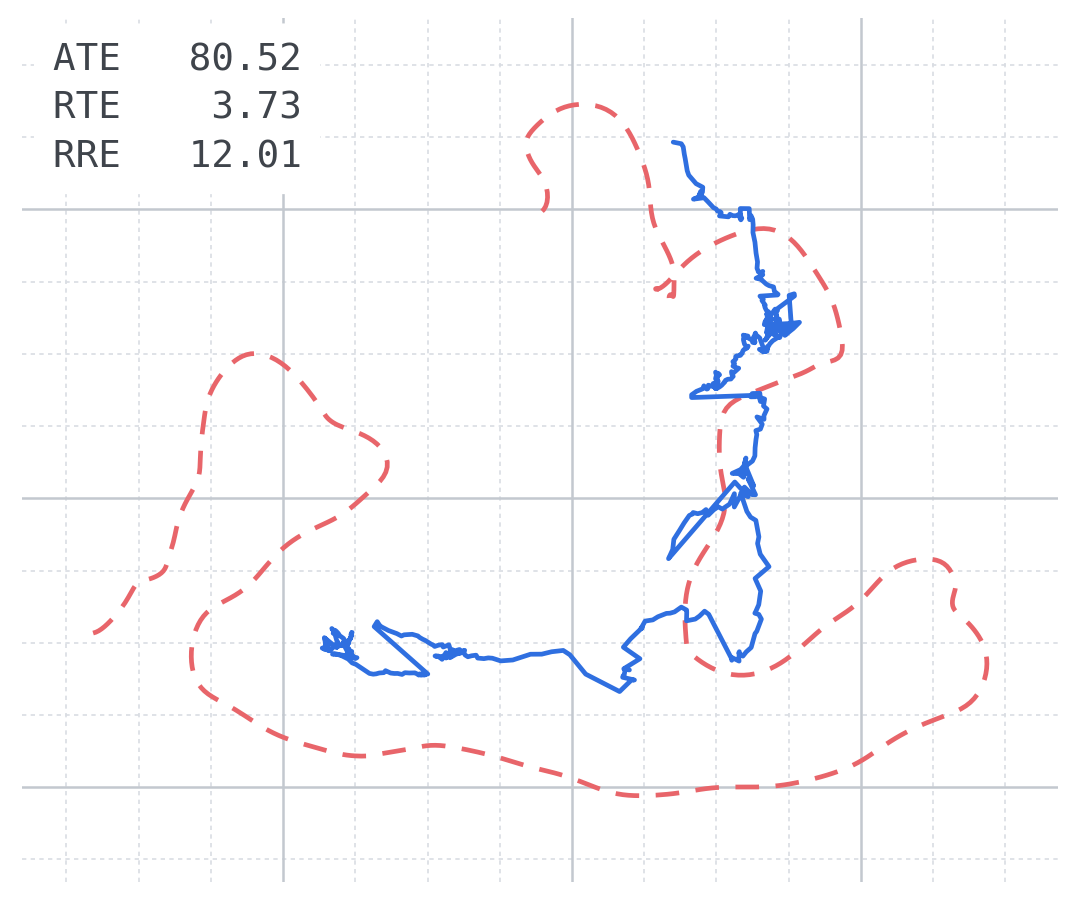} & \includegraphics[width=0.315\textwidth]{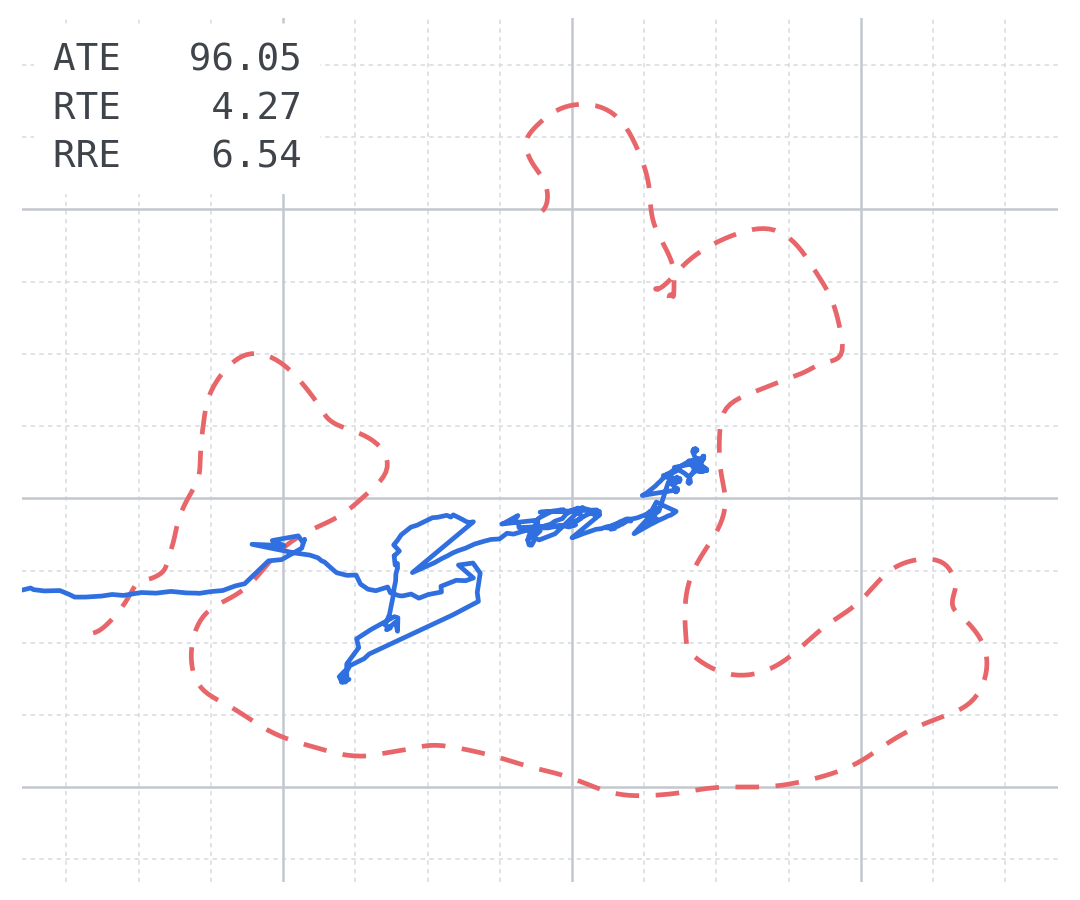} \\[3pt]
\end{tabular}
\par\vspace{3pt}
\textcolor{blue}{\rule[0.5ex]{12pt}{0.7pt}} Predicted trajectory\qquad\textcolor{red}{\textbf{-\,-\,-}} GT trajectory
\endgroup
\caption{\textbf{Long-sequence reconstruction on RealSynCol Colon 2.}}
\label{fig:long}
\vspace{-1em}
\end{figure*}